\documentclass{article}
\usepackage{microtype}
\usepackage{graphicx}
\usepackage{subcaption}
\usepackage{booktabs} 
\usepackage{colortbl}
\usepackage[table]{xcolor}
\usepackage{hyperref}
\usepackage{bm}
\usepackage{multirow}
\usepackage{adjustbox}
\usepackage[ruled,linesnumbered]{algorithm2e}

\SetKw{KwRequire}{\textbf{Require:}}
\SetKw{KwOutput}{\textbf{Output:}}

\usepackage[accepted]{icml2026}
\usepackage{amsmath}
\usepackage{amssymb}
\usepackage{mathtools}
\usepackage{amsthm}
\usepackage{pifont}
\usepackage{array}
\usepackage{tcolorbox}
\usepackage[capitalize,noabbrev]{cleveref}
\theoremstyle{plain}
\newtheorem{theorem}{Theorem}[section]

\newtheorem{lemma}[theorem]{Lemma}

\theoremstyle{definition}

\newtheorem{assumption}[theorem]{Assumption}
\theoremstyle{remark}

\usepackage[textsize=tiny]{todonotes}
\definecolor{bestgreen}{RGB}{220,235,225}
\definecolor{mygreen}{HTML}{238b21}
\definecolor{cite}{HTML}{001473}
\definecolor{bggray}{gray}{0.9}

\icmltitlerunning{Correcting Visual Blur Induced by Attention Distraction to Reduce Hallucinations: Algorithm and Theory}

\begin{document}

\twocolumn[
  \icmltitle{Correcting Visual Blur Induced by Attention Distraction to Reduce Hallucinations: Algorithm and Theory}

  \icmlsetsymbol{equal}{*}
  \begin{icmlauthorlist}
    \icmlauthor{Quanjiang Li}{equal,yyy}
\icmlauthor{Zhiming Liu}{equal,xxx}
    \icmlauthor{Wei Luo}{zzz}
    \icmlauthor{Tingjin Luo}{yyy}
    \icmlauthor{Chenping Hou}{yyy}

  \end{icmlauthorlist}

  \icmlaffiliation{yyy}{National University of Defense Technology, China}
  \icmlaffiliation{xxx}{Harbin Institute of Technology (Shenzhen), China}
\icmlaffiliation{zzz}{Xi'an Jiaotong University, China}

  \icmlcorrespondingauthor{Chenping Hou}{hcpnudt@hotmail.com}
\icmlcorrespondingauthor{Tingjin Luo}{tingjin-
luo@hotmail.com}


  \vskip 0.3in
]



\printAffiliationsAndNotice{\icmlEqualContribution}

\begin{abstract}
Multimodal large language models (MLLMs) frequently suffer from object hallucinations, yet the visual perceptual mechanism underlying this failure remains poorly understood. In this work, we reveal that hallucinations are strongly associated with a human-like attention distraction phenomenon, where humans under divided focus experience degraded visual clarity and produce inaccurate descriptions, while in models the same mechanism manifests as spatial inconsistency in multi-head attention and temporal fading of attention to image tokens during decoding.
We further provide theoretical insights  that attention dispersion increases model complexity and degrades classification generalization. Motivated by these findings, we propose an Attention-Focused Approach for Improved Image Perception (AFIP), which corrects   attention distraction  via cross-head  attention enrichment and reinforces visual grounding through dynamic historical attention enhancement. Extensive experiments on multiple  benchmarks and models validate the effectiveness  of AFIP without additional training. Code is available at: 
{\small \url{https://github.com/MIKUZ12/AFIP}}.
\end{abstract}
\begin{figure}
    \centering
    \includegraphics[width=1\linewidth]{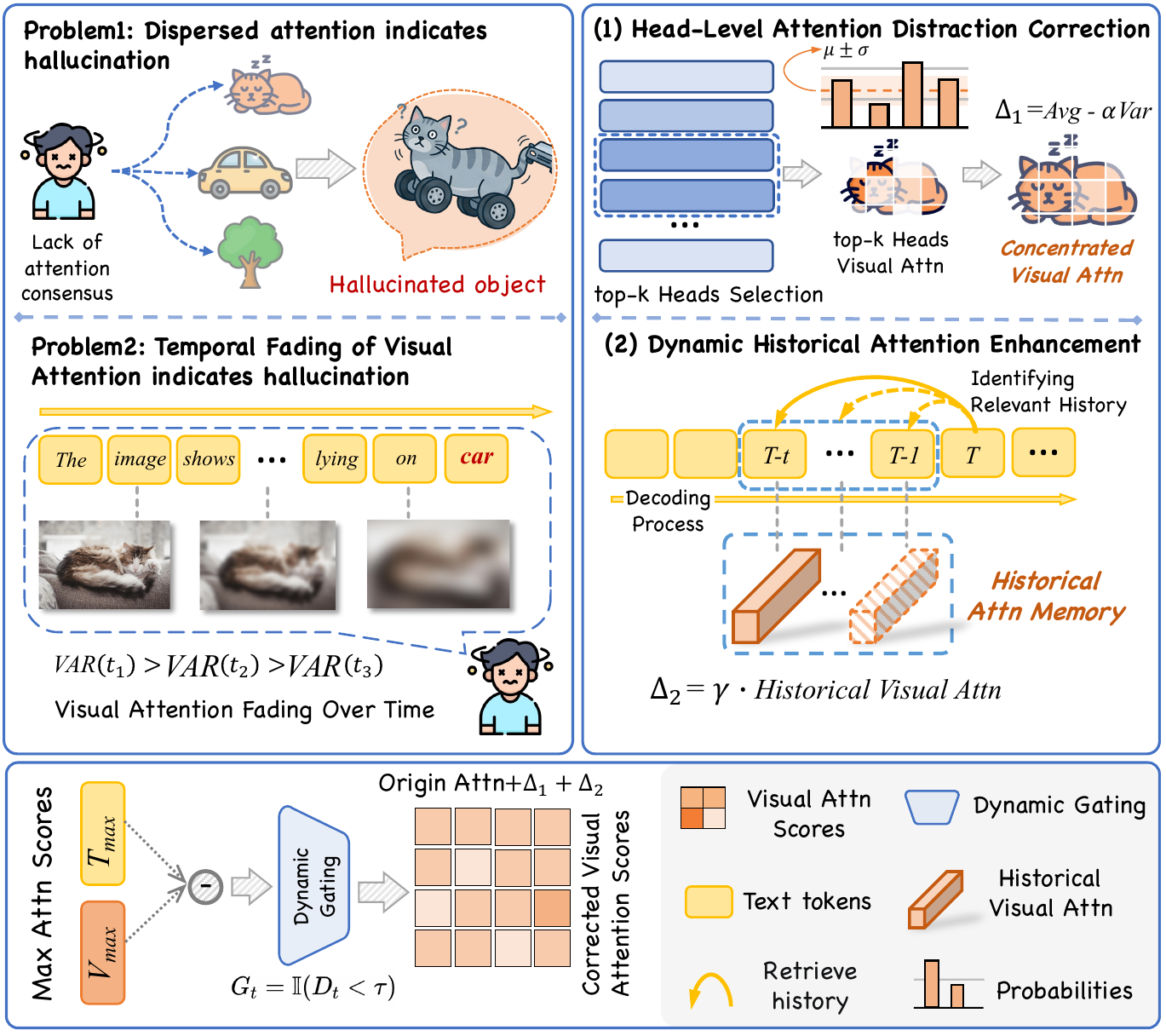}
    \caption{Motivation illustration and overview of AFIP.}
    \label{fig:motivation}
\end{figure}
\section{Introduction}
MLLMs have recently made substantial progress and have been applied to a wide range of tasks, including medical analysis \citep{lin2025healthgptmedicallargevisionlanguage,li2025theory}, creative content generation~\citep{wang2023docllmlayoutawaregenerativelanguage,huang2023languageneedaligningperception} and autonomous driving~\citep{sima2025drivelmdrivinggraphvisual, guo2024vlmautovlmbasedautonomousdriving}. Such progress is enabled by joint modeling of visual and textual information, which facilitates integrated multimodal understanding and reasoning across diverse real-world scenarios. Despite this impressive versatility \cite{li2026theory,li2025semi}, modern MLLMs still suffer from a critical limitation, as they may generate outputs that are insufficiently grounded in the actual visual input.  In particular, models can produce confident yet erroneous descriptions, such as attributing nonexistent objects or incorrectly specifying visual attributes (e.g., color, number, or spatial relations). This phenomenon is commonly referred to as hallucination and poses a substantial obstacle to the reliable deployment of MLLMs in risk-sensitive and high-precision applications.

To address the hallucination issue in MLLMs, recent researches have explored a diverse set of mitigation strategies that can be summarized into four paradigms. The first line of work exploits external knowledge sources through the incorporation of auxiliary databases or retrieval modules into the generation process, thereby supplying explicit factual or visual cues \cite{qu2024alleviating}. The second direction prioritizes model fine-tuning and aims to improve generation reliability via parameter updating driven by task-specific or consistency-oriented supervision \cite{yu2024rlhf}.  The third category concentrates on attention adjustment mechanisms, which modifies  attention distributions across different tokens during inference to reduce spurious correlations\cite{huang2024opera}.  The fourth stream builds upon contrastive learning principles, which compares alternative responses or representations so that outputs with stronger visual support are preferred \cite{leng2024mitigating}. The fifth paradigm focuses on hidden-state intervention, which modifies the internal representations of the model during inference so that the generated outputs better align with true visual perception ~\cite{zou2024look}.

Despite the proposal of a wide range of methods aimed at mitigating hallucinations, they still exhibit the following major limitations. (i) Existing methods often entail a reliance on supplementary resources or extended processing time. In particular, approaches based on external knowledge incorporation or  model fine-tuning require auxiliary components, such as  large-scale instruction-oriented datasets, to correct hallucinated outputs in a post-hoc manner \cite{yang2024rag, liu2023mitigating}. Although attention adjustment methods avoid dependence on external data, they often involve complex inference-time procedures, such as retrospection-based decoding or multi-step attention reallocation, which inevitably increase  memory consumption \cite{tu2026attention}. (ii)  The visual perceptual mechanism underlying hallucinations remains insufficiently explored. Compared with text, images carry denser information and exhibit richer structural patterns, which makes it difficult for MLLMs to faithfully encode, maintain, and exploit visual evidence \cite{liu2024paying}. Therefore, hallucinations should not be explained solely by language-side biases, since distortions introduced during visual interpretation can also play a crucial role. Although prior studies mainly associate hallucinations with language-related factors, such as text inertia \cite{liu2024paying}, the influence of visual perception has not been sufficiently examined. In practice, failures in visual information processing can directly weaken the grounding of generated tokens and  trigger hallucinated descriptions. (iii) Existing explanations of hallucination causes remain limited from both spatiotemporal and theoretical perspectives. The generation process of MLLMs involves spatial grounding over image regions as well as temporal reasoning during decoding. Errors arising from incorrect spatial attention may be progressively amplified in subsequent decoding steps. However, few studies have jointly examined hallucinations from the perspective of spatial attention dynamics and temporal degradation of visual grounding. Moreover, the analogy between model hallucination and human perceptual phenomena has rarely been investigated, despite its potential to provide intuitive insights into model behavior. From a theoretical standpoint, hallucinations are still insufficiently characterized. Since the final decoding stage can be viewed as a token classification process, the factors that impair token discrimination deserve further investigation.

To address these challenges, we begin by establishing a strong correspondence between hallucination phenomena in MLLMs and human physiological behavior under dispersed attention, supported by extensive statistical validation. As illustrated in Fig. \ref{fig:motivation}, when humans describe visual content under divided focus, frequent shifts in attention render the cognitive representation of the target image increasingly indistinct and less salient, leading to confabulated descriptions. A comparable pattern emerges in MLLMs, manifested as inconsistent multi-head attention distributions across spatial positions and a progressive decline of visual focus during autoregressive reasoning. These spatiotemporal misalignments impair the integration of visual information and compromise the fidelity of image perception. Besides, theoretical analyses of the multi-head attention architecture, together with the view that attention heads act as independent information sources whose fusion governs generalization, suggest that cross-head attention inconsistency increases model complexity and weakens generalization capability. Motivated by these insights, we propose the Attention Focused Approach for Improved Image Perception (AFIP), a training-free framework that imposes no additional inference overhead. AFIP calibrates high-confidence focused regions by aggregating multi-head information, while penalizing and eliminating irrelevant areas with a variance-induced regularization term. Furthermore, reliable visual focus derived from prior decoding steps is adaptively fused into the current state to mitigate temporal decay in perception. A dynamic gating mechanism, guided by the difference between peak visual and textual attention, prevents excessive adjustment and ensures response stability. Together, these designs enable robust visual grounding and substantially reduce hallucination. The primary contributions of this paper are summarized as follows:

\begin{itemize}
\item[$\bullet$]
For the first time, we reveal a strong analogy between the underlying causes of machine hallucinations in MLLMs and human visual blur resulting from attention distraction. This physiological phenomenon is further decomposed into two factors intrinsic to the MLLMs, i.e., spatial inconsistency of multi-head attention and temporal attenuation of visual attention.
\item[$\bullet$]
Through analyzing the architecture of the multi-head attention mechanism and the information aggregation capabilities it enables, we theoretically establish that discrepancies across attention heads increase model complexity and impair token-level discrimination, thereby reducing the overall generalization performance.
\item[$\bullet$]
We propose AFIP, a plug-and-play module that incurs no additional cost to mitigate object hallucinations. Extensive experiments demonstrate that AFIP surpasses mainstream baselines and attains SOTA status.
\end{itemize}
\begin{figure*}
    \centering
\includegraphics[width=1\linewidth]{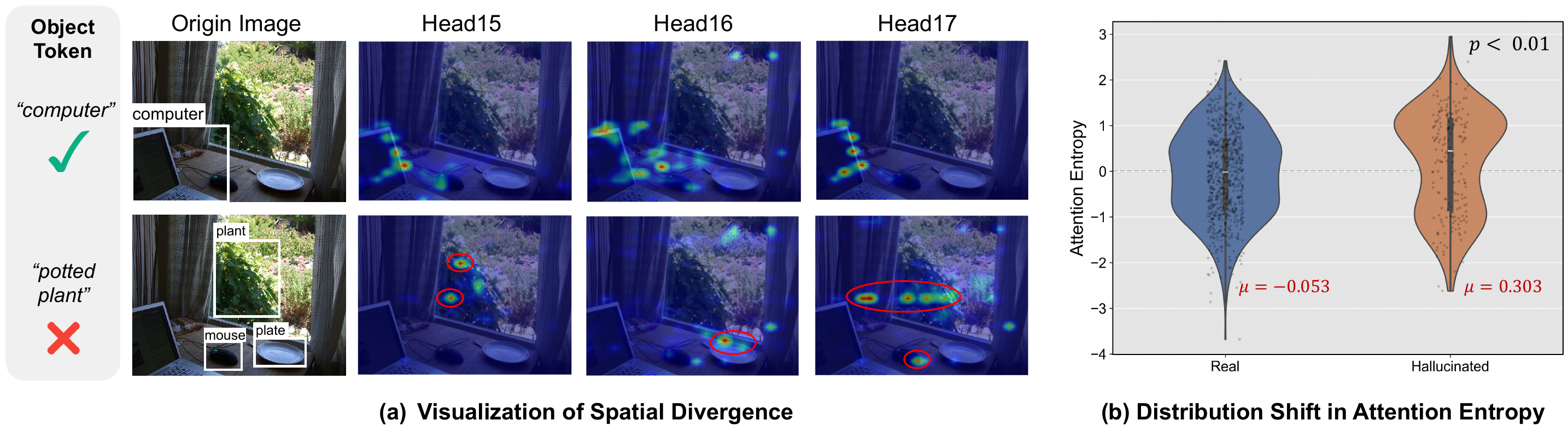}
\caption{Correspondence between spatial attention inconsistency and object hallucination. \textbf{(a)} Attention maps for the correct token computer'' and the hallucinated token potted plant''. Attention linked to the hallucinated token is broadly dispersed across the image, whereas attention for the correct token remains sharply concentrated on the target object. Additional examples are provided in Appendix~\ref{attn-incon}. \textbf{(b)} Distribution shift in attention entropy. Tokens are categorized as \emph{Real} or \emph{Hallucinated}. Real tokens predominantly exhibit low entropy, while hallucinated tokens correspond to substantially higher entropy values. The result $p<0.01$ indicates a statistically significant difference in mean entropy between the two groups.}
    \label{finding1}
\end{figure*}
\section{Related work}
\subsection{Vision Language models}
In recent years, vision-language modeling has undergone a significant paradigm shift. Early vision-language models primarily relied on BERT-style language decoders~\cite{devlin2019bert,koroteev2021bert} to integrate visual representations with text, enabling unified  representation learning but offering limited generative flexibility.~\cite{li2019visualbert,sun2019videobert, huangmises}. The emergence of large language models (LLMs)~\cite{brown2020language,achiam2023gpt,bai2023qwen, huang2026skill} significantly enhanced the expressive power of language backbones and spurred the development of MLLMs ~\cite{alayrac2022flamingo,wang2024qwen2,driess2023palme, liu2026all}. Leveraging end-to-end training and joint decoding over visual and textual tokens, MLLMs exhibit substantially improved adaptability. Contemporary MLLMs generally follow a unified processing pipeline in which images are first encoded by powerful vision backbones such as CLIP~\cite{radford2021learning} or EVA~\cite{fang2023eva}, aligned through lightweight projection modules, and decoded by pretrained LLMs. Nevertheless, the specific design of  projection modules differs substantially across models. LLaVA-1.5~\cite{liu2024improved} adopts a linear projection for modality alignment, while MiniGPT-4~\cite{zhu2023minigpt} utilizes a querying transformer to compress visual representations. More recent MLLMs, including Qwen-VL~\cite{bai2023qwenvlversatilevisionlanguagemodel}, InternVL~\cite{chen2024internvl}, and LLaVA-NeXT~\cite{li2024llava}, further improve multimodal reasoning through enhanced alignment and high-resolution visual instruction tuning.

\subsection{Object Hallucination in MLLMs}
Hallucination remains a persistent and consequential challenge in modern MLLMs. Such inaccuracies are particularly concerning in high-stakes domains, including medical imaging~\cite{kraljevic2021medgpt,zheng2025large,he2023geometric} and autonomous driving~\cite{cui2024survey}, where unreliable outputs may lead to severe downstream consequences.  In the vision-language setting, hallucination often manifests as object hallucination, where models generate plausible descriptions that mention objects missing from the image~\cite{li2023evaluating,rohrbach2018object}. 
A wide range of mitigation strategies has emerged in recent years. Early studies mainly relied on representation-level techniques, including contrastive learning, ROI-level feature fusion~\cite{biten2022let}, and data augmentation for reducing spurious cross-modal correlations~\cite{kim2023exposing}. Later efforts introduced visual instruction tuning~\cite{jiang2024hallucination,yu2024hallucidoctor,zhu2023debiased} and external expert modules for auxiliary verification and logical constraints during generation~\cite{chen2024halc,wu2024logical,zhao2024mitigating}. Meanwhile, substantial progress has also been made in hallucination evaluation and detection, with fine-grained benchmarks and diagnostic datasets proposed to assess object-level faithfulness and relational consistency~\cite{li2023evaluating,wang2023evaluation}. Building on these advances, an increasing number of training-free methods have been proposed to  mitigate hallucination in real-time deployments. Representative directions included decoding-time interventions for improving reasoning fidelity \cite{huang2024opera}, attention modulation \cite{tu2026attention} for restoring proper token-level correlations, and hidden-layer adjustments for enhancing visual encoding \cite{zou2024look}.
Although these approaches have shown promising results, the visual mechanisms underlying hallucination remain insufficiently understood. In particular, it is still unclear how MLLMs perceive visual content and where visual evidence becomes distorted during generation. This gap motivates us to investigate hallucination from the perspective of visual information processing and develop lightweight  interventions that improve visual grounding at the root level.

\section{Uncover The Core Reasons Behind Hallucinations}
In this section, we begin by delineating the fundamental operational mechanisms of MLLMs and introduce two pivotal metrics, i.e.,  Attention Entropy and Visual Attention Ratio, which are employed to probe internal signals and characterize model behavior. Utilizing these metrics, we conduct a comprehensive set of empirical analyses and identify two principal factors that drive hallucinations, including spatial inconsistency across multi-head attention and temporal attenuation of visual attention. Our case studies are based on a randomly sampled subset of the COCO validation split~\cite{lin2014microsoft}, primarily focusing on LLaVA-1.5-7B~\cite{liu2024improved}, with other models evaluated as well.
\subsection{Preliminary}

\noindent\textbf{Notation.}
We consider a multimodal large language model $\mathcal{F}_{\theta}$ parameterized by $\theta$, whose architecture is composed of a text embedding component, a vision encoder, a cross-modal alignment module, and a Transformer-based decoder. Given a sequence of input image tokens $\{{v}_{1}, \ldots,{v}_{n}\}$, together with prompt tokens $\{{t}_{1}, \ldots, {t}_{m}\}$, followed by the $t-1$ previously generated tokens $\{{y}_{1}, \ldots, {y}_{t-1}\}$, the model produces the current output token $y_t$ in an autoregressive manner. Define $L$ as the total number of Transformer layers, each comprising $H$ attention heads. At layer $l \in[L]$, the attention weights in head $h \in[H]$ are represented as $\bm{A}_{t}^{(l, h)} \in \mathbb{R}^{a_{t} \times a_{t}}$, where $a_{t}=n+m+t-1$. Besides, let $s$ and $e$ denote the start and end indices of the image tokens.

\noindent\textbf{Attention Entropy.}
At layer $l$, we obtain the head-averaged attention distribution over the $n$ visual patches for the currently generated token by ${\bm{A}^{l}_t}=\frac{1}{H}\sum_{h=1}^{H}\bm{A}^{(l,h)}_t$.
Then, we define the attention entropy as below:
\begin{equation}
\label{entropy}
\mathcal{D}^{l}_t \;=\; - \sum_{i=1}^{n}{\bm{A}}_{t}^{l}(a_t, i) \log({\bm{A}}_{t}^{l}(a_t, i) + \epsilon),
\end{equation}
where $\bm{A}_{t}^{l}(a_t,i)$ denotes the head-averaged attention weight from the current decoding position $a_t$ to the $i$-th visual token, and $\epsilon>0$ is a small constant for numerical stability. The entropy $\mathcal{D}^{l}_t$ captures the dispersion of attention across image patches. A smaller $\mathcal{D}^{l}_t$  corresponds to a concentrated attention distribution over a limited subset of visual patches, indicating a focused and selective visual association for the current token. Conversely, a larger entropy value reflects a spatially dispersed attention pattern, suggesting diffuse and uncertain visual grounding.

\noindent\textbf{Visual Attention Ratio.}
\label{VAR}
Regarding the $t$-th output token $y_t$ at layer $l$ and head $h$, we quantify its reliance on visual information by measuring the total attention mass assigned to the visual prefix:
\begin{equation}
\mathrm{VAR}_{t}^{(l,h)} \;\triangleq\; \sum_{i=1}^{n} \bm{A}_{t}^{(l,h)}(a_t, i).
\end{equation}
$\mathrm{VAR}_{t}^{(l,h)}$represents the proportion of attention devoted to visual tokens during the generation of $y_t$. A larger value indicates stronger dependence on image information, whereas a smaller value suggests that the token is generated predominantly from textual context.


\subsection{Finding 1: Spatial Inconsistency in Multi-head Attention indicates Hallucination} \label{finding1}
Inspired by humans’ stable visual fixation on salient regions, visually grounded tokens in MLLMs are expected to elicit concentrated attention with strong cross-head spatial consistency. In contrast, hallucinated tokens tend to be associated with diffuse visual attention and spatial misalignment across heads. To empirically examine this distinction, we extract the attention entropy    $\mathcal{D}^{l}_t$ at each decoding step. Following the CHAIR criteria~\cite{rohrbach2018object}, tokens are categorized into \textit{Real} and \textit{Hallucinated} groups, and their entropy distributions are compared through statistical significance testing. As illustrated in Fig.~\ref{finding1}, grounded tokens, such as \textit{computer}, exhibit sharply localized attention and strong spatial alignment across heads, whereas hallucinated tokens display dispersed attention patterns and pronounced head-wise divergence. Quantitatively, hallucinated tokens consistently yield higher $H^{l}_t$ than real tokens. Moreover, the entropy distribution of real tokens is unimodal, whereas hallucinated tokens exhibit a bimodal, long-tailed distribution, indicating  diffuse and unstable visual perception.
 
Building on prior observations that hallucinated tokens are often accompanied by heightened cross-head disagreement, we further characterize the role of head-wise inconsistency by examining its statistical association with the confidence of token prediction. During decoding, we normalize $\bm{A}_{t}^{(l,h)}(a_t,i)$ to a probability vector $P^{(l,h)}_t$ with $\sum_{i=1}^{n}P^{(l,h)}_t(a_t,i)=1$. Then, we  compute the head-averaged probability distribution as ${P^{l}_t}=\frac{1}{H}\sum_{h=1}^{H}P^{(l,h)}_t$. From an information-theoretic perspective, head-wise inconsistency is quantified using the Kullback–Leibler divergence between each individual distribution and ${P^{l}_t}$:
\begin{equation} \small
D_{{kl}}
=\frac{1}{H}\sum_{h=1}^{H} P^{(
l,h)}_t\big(\log P^{(l,h)}_t-\log P^{l}_t\big),
\end{equation}
where $D_{{kl}}^l$ provides a principled measure of the extent to which individual attention heads deviate from the collective attention pattern, with larger values indicating weaker cross-head agreement. As for quantifying token-level uncertainty, we compute the following entropy of the output distribution:
\begin{equation}
E_{{tok}}=-\sum_{e}p(e)\log p(e),
\end{equation}
where $p(e)$ denotes the predicted probability of each candidate word $e$. Then, we sort tokens tokens according to the layer-averaged cross-head divergence $D^{l}_{{kl}}$, where $D_{{kl}}=\frac{1}{L}\sum_{l=1}^{L} D^{l}_{{kl}}$. The ranked tokens are subsequently divided into quantile-based groups. For each group, we compute the mean token-level predictive entropy $E_{tok}$ separately for real and hallucinated tokens, with bootstrap confidence intervals estimated from 2,000 resampling trials. As illustrated in Fig.~\ref{fig:Attention-Correction} (Left), increasing values of $D_{{kl}}$ tends to coincide with higher $E_{{tok}}$ for both hallucinated and real tokens. Moreover, correlation tests are conducted to indicate a positive correlation between cross-head divergence and predictive entropy, with small $p$-values supporting its statistical reliability.  These results indicate that stronger head-wise inconsistency is consistently accompanied by lower token prediction confidence and reduced reasoning accuracy.

\subsection{Finding 2: Temporal Fading of Visual Attention indicates Hallucination} \label{finding2}
\begin{figure}
    \centering
    \includegraphics[width=1\linewidth]{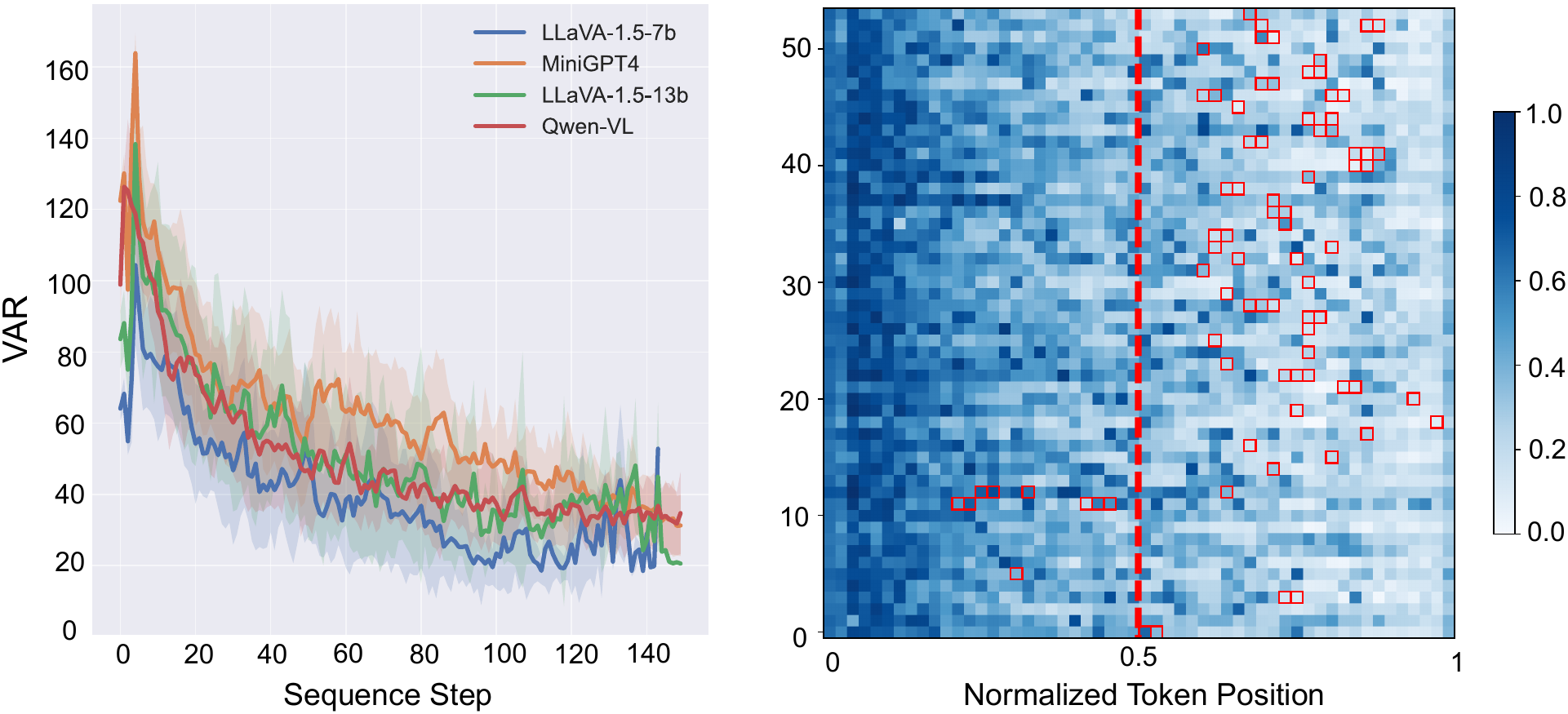}
    \caption{Temporal Fading of Visual Attention. \textbf{(Left)} VAR decreases as generation proceeds across multiple MLLMs, indicating progressive weakening of visual grounding in long-form responses. The shaded area represents the standard deviation band, evaluated on the COCO dataset. \textbf{(Right)} Heatmap shows VAR over normalized token position, with red boxes denoting hallucinated tokens.}
    \label{fig:fade-trend}
\end{figure}

Human's long-horizon reasoning relies on sustained attention to maintain coherent progress.
 However, in long-sequence visual question answering tasks performed by MLLMs, attention directed to visual regions typically diminishes as decoding advances \cite{tu2026attention}.
 To quantify this effect, we prompt the model to generate long-form responses while recording layer-wise attention distributions at each decoding step for VAR computation. These values are then averaged across layers to produce a VAR profile that tracks visual attention throughout the sequence of yielded  tokens. As illustrated in Fig.~\ref{fig:fade-trend} (Left), the resulting mean VAR exhibits a consistent downward trajectory during generation across all model backbones, which reflects a progressively reduced reliance on visual evidence as decoding proceeds.

Building on the preceding exploration, we further investigate the interplay between the attenuation of visual attention and hallucination during long-form generation. As decoding unfolds, the model gradually shifts focus away from visual inputs and increasingly relies on linguistic priors, increasing the likelihood of generating hallucinated tokens.
To quantitatively reveal this inherent association, we track the occurrence of real and hallucinated tokens during decoding, linking each token to its corresponding VAR value. Outputs are normalized by total length to account for varying response lengths, and tokens are positioned according to their relative location along the normalized trajectory.
The resulting patterns are visualized in  Fig.~\ref{fig:fade-trend} (Right), with color intensity representing VAR magnitude, which shows that hallucinated tokens predominantly cluster in low-VAR regions and emerge in the later stages of generation. In particular, tokens falling within the lowest 30\% of VAR values exhibit a markedly higher incidence of hallucination compared to the remainder. Taken together, these observations uncover a clear statistical association between visual blur and hallucination, suggesting that sustained reduction of visual grounding during extended responses coincides with increased hallucination risk.
 
\section{Object Hallucination Mitigation}
\label{method}
\subsection{Head-Level Attention Distraction Correction}
\begin{figure}
\centering\includegraphics[width=0.9\linewidth]{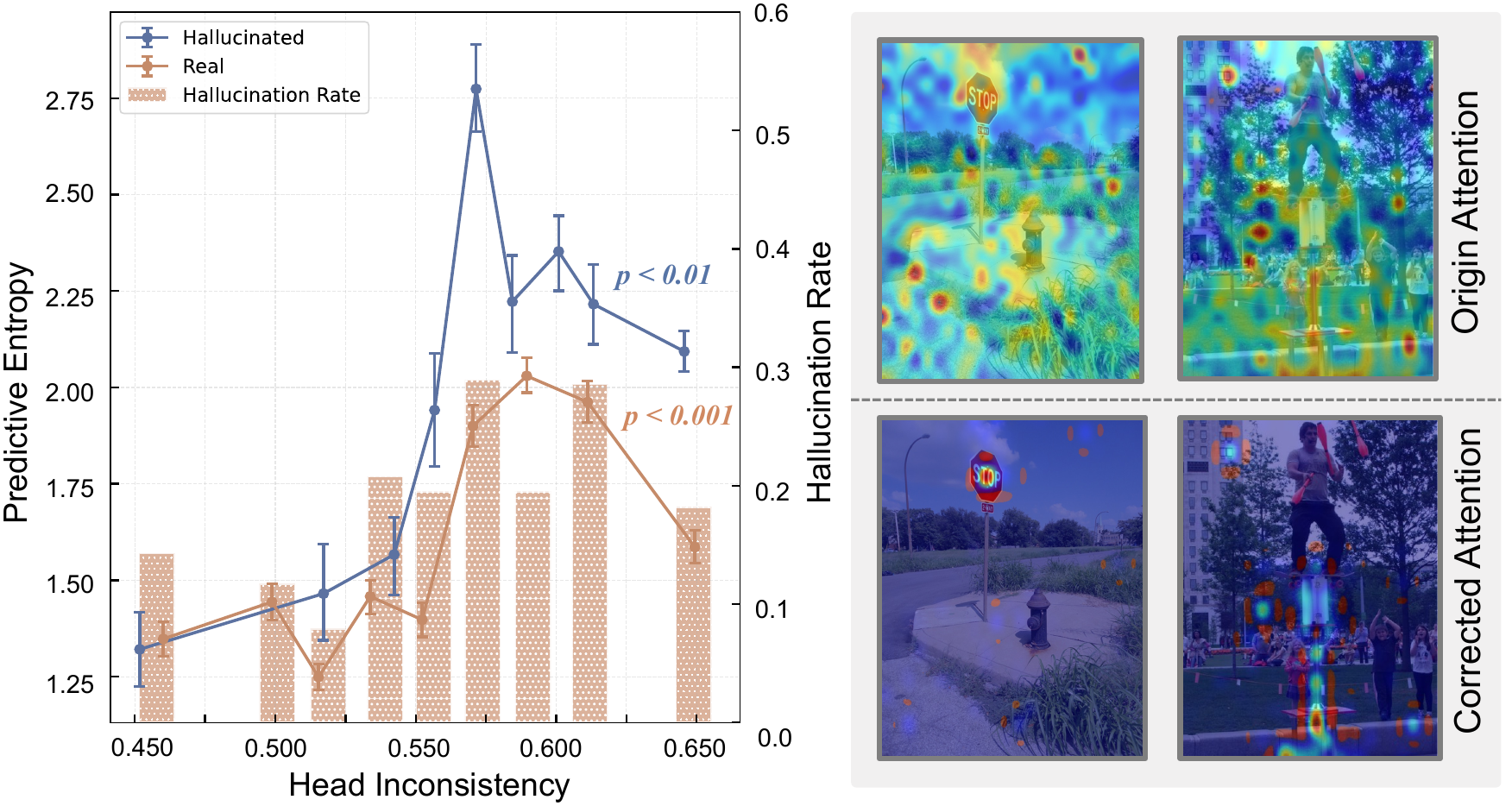}
\caption{\textbf{(Left)} Predictive entopy under varying head inconsistency. Increasing head inconsistency is associated with elevated uncertainty in token prediction, with hallucinated tokens consistently displaying higher entropy than real tokens. The bars denote the hallucination rate, which also increases as head inconsistency grows. \textbf{(Right)} Case study for attention correction. Attention maps before (top) and after (bottom) correction during the generation of sign and platform reveal that the correction enhances visual focus on the relevant regions. More cases are shown in Appendix~\ref{attn-corr}.}
    \label{fig:Attention-Correction}
\end{figure}
As established in the Section \ref{finding1}, the emergence of hallucinations 
primarily arises from attention distraction, characterized by diffuse attention allocation to irrelevant regions. Consequently, accurate grounding of target tokens requires a highly concentrated attention distribution over salient positions, with multiple heads jointly focusing on genuine visual objects. Therefore, the attention weights $\{{A}_{t}^{(l, h)}(a_{t}, i)\}_{i=1}^{n}$ assigned to image tokens should exhibit both high confidence and cross-head consistency. To this end, we aggregate strong attention signals from multiple heads to identify reliable perceptual cues. Specifically, the attention score matrix $\bm{B}_{t}^{(l, h)}$  before softmax operation
is extracted by
\begin{equation}\label{eq1}
    \bm{B}_{t}^{(l, h)}=\left(\frac{\bm{Q}_{l, h} \bm{K}_{l, h}^{\top}}{\sqrt{d_{t}}}\right)_{t},
\end{equation}
where $\bm{Q}_{l, h}$ and $\bm{K}_{l, h}\in \mathbb{R}^{a_{t} \times d_{t}}$ represent the query and key matrices of dimension $d_{t}$, respectively. To mitigate visual divergence caused by human-like attention shift,  the cross-head mean attention is used as the criterion for identifying faithful image tokens, and  the corresponding  variance is adopted as a regularization term to penalize insufficient concentration. Moreover, attention heads that manifest pronounced responses to image tokens can offer valuable cues for calibrating attention distributions, where such response intensity is measured by the magnitude of $|\bm{B}_{t}^{(l, h)}\left(a_{t}, i\right)|$. Then, the top-$k$ heads in each layer are selected based on the maximum absolute response each head exhibits across all image tokens, forming $\mathcal{H}_{k}^{(l)} \subseteq\{1,2, \ldots, H\}$ with $|\mathcal{H}_{k}^{(l)}|=k$. For image token $i$, the head-wise mean and variance formulations of the attention scores are given as follows:
\begin{equation}\label{eq2} \small
\left\{\begin{array}{l}
   Avg^{(l)}  = \frac{1}{H} \sum_{h \in \mathcal{H}_{k}^{(l)}}\left|\bm{B}_{t}^{(l, h)}\left(a_{t}, i\right)\right|  \\
  Var^{(l)}  = \frac{1}{H} \sum_{h \in \mathcal{H}_{k}^{(l)}}(\left|\bm{B}_{t}^{(l, h)}\left(a_{t}, i\right)\right|- Avg^{(l)})^2.
\end{array}\right.
\end{equation}
Formally, we adjust the visual attention scores by enriching beneficial information across multiple  heads and suppressing inconsistent attention responses:
\begin{equation}\label{eq6}
\bm{\hat{B}}_{t}^{(l, h)}\left(a_{t}, i\right)=\bm{B}_{t}^{(l, h)}\left(a_{t}, i\right)+ Avg^{(l)}-\alpha   Var^{(l)},
\end{equation}
where the incremental term $Avg^{(l)}-\alpha   Var^{(l)}$ is abbreviated as $\Delta_1^{(l)}$. As illustrated in Fig. \ref{fig:Attention-Correction} (Right), regulating visual attention by Eq. (\ref{eq6}) leads to a concentrated head-level attention distribution, which facilitates the model in aligning visual content with textual semantics and identifying the correct generated words.
\subsection{Dynamic  Historical Attention Enhancement}
\begin{figure}[htbp]
    \centering
    \begin{subfigure}[b]{0.235\textwidth}
        \centering
        \includegraphics[width=\linewidth]{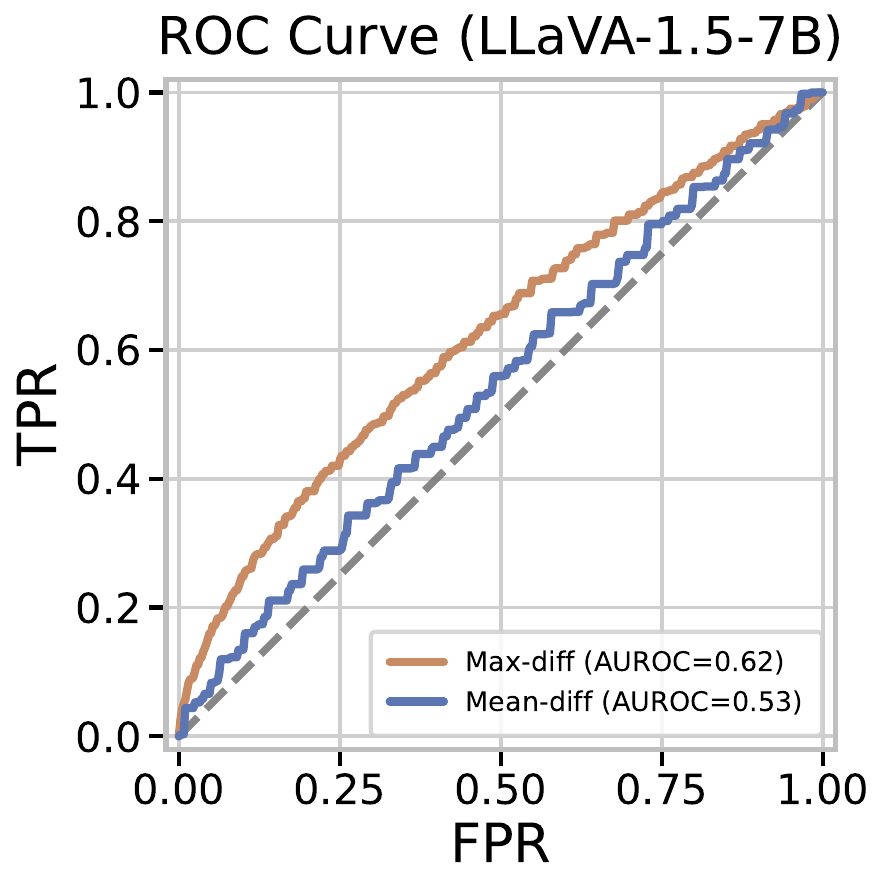}
        \caption{LLaVA-1.5-7b}
        \label{fig:roc-llava}
    \end{subfigure}
    \hfill
    \begin{subfigure}[b]{0.235\textwidth}
        \centering
        \includegraphics[width=\linewidth]{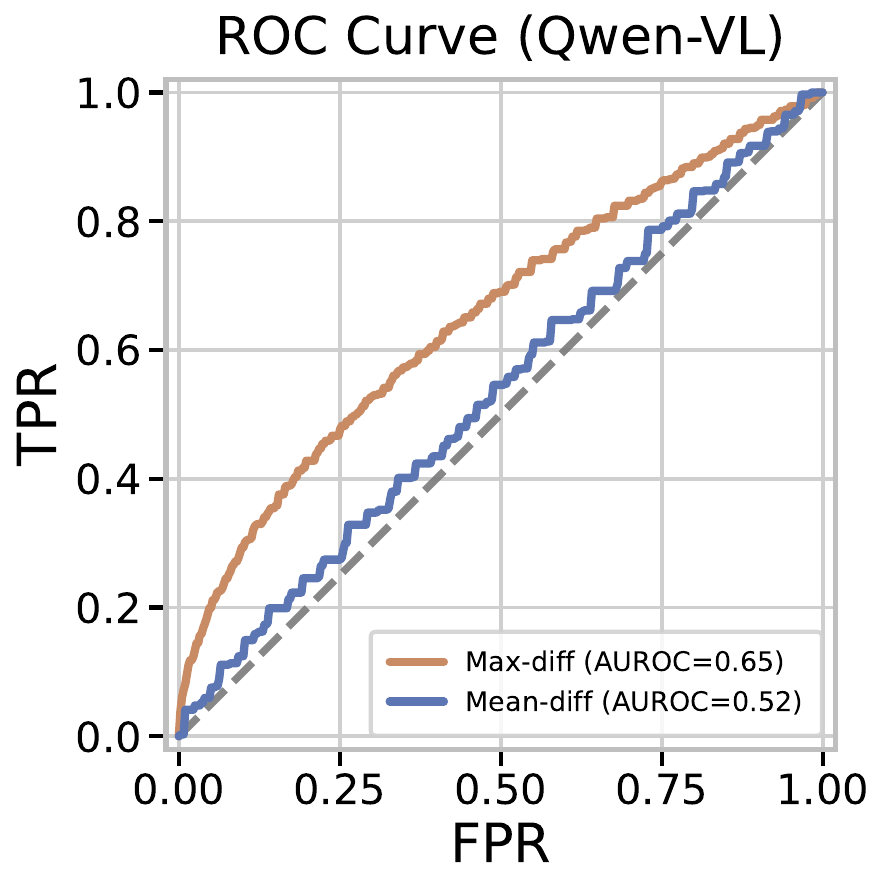}
        \caption{Qwen-VL}
        \label{fig:roc-qwen}
    \end{subfigure}
    \caption{Hallucination occurrence with different attention-based measures. The ROC (Receiver Operating Characteristic) curves of two models are presented  for hallucination detection.}
    \label{fig:roc}
\end{figure}

The progressive degradation of visual perception accompanied by attention distraction, as discussed in the Section \ref{finding2}, underscores the importance of reintegrating historical visual attention into the decoding process to correct visual blur and prevent the loss of image content. 
Accordingly, historical visual attention scores are first stored within a fixed-length history window as below:
\begin{equation}\label{eq4}
\mathcal{C}_{t}=\{\bm{C}^{(l, h)}_{t-i}\}_{i=1}^{\mathcal{W}_{t}}, \quad \bm{C}_{t-i}^{(l, h)}=\{\bm{B}_{t-i}^{(l, h)}(a_t,j)\}_{j=s}^{e},  
\end{equation}
where $\mathcal{W}_{t}$ denotes the size of the history window, fixed at 8 to constitute a sufficiently extended temporal context. 
Subsequently, the historical text token exerting the greatest influence on the decoding process is identified by
\begin{equation}\label{eq5}
  r_{t}^{(l, h)}=\arg \max _{j \in \mathcal{W}_{t}} \bm{B}_{t}^{(l, h)}\left(a_{t}, a_{t}-1-j\right).  
\end{equation}
Subsequently, the attention over the visual content conditioned on the most relevant historical text is fused into the current attention score to strengthen the comprehension of the visual information:
\begin{equation} \label{eq66} \small
\bm{\tilde{B}}_{t}^{(l, h)}\left(a_{t}, j\right)=\bm{\hat{B}}_{t}^{(l, h)}\left(a_{t}, j\right)+\gamma \bm{C}^{(l, h)}_{t-1-r_{t}^{(l, h)}}(s+j),  
\end{equation}
where the incremental term $\gamma \bm{C}^{(l, h)}_{t-1-r_{t}^{(l, h)}}(s+j)$ is denoted as $\Delta_2^{(l,h)}$. As shown in  Appendix \ref{Attention Enhancement}, the incorporation of $\Delta_2^{(l,h)}$ effectively strengthens visual attention and mitigates the temporal perceptual decay, which ensures that visual representations remain accurately interpreted over time.

Although the application of Eq. (\ref{eq66}) substantially strengthens the model’s sensitivity to image tokens, it is crucial to mitigate abnormal generation behaviors caused by excessive visual retracing, while maintaining   awareness to necessary textual context.
 To this end, we propose a dynamic gating mechanism that controls the activation conditions of attention adjustment. 
 Specifically, we compute the maximum attention score  over the visual  contents as $V^{(l,h)}_t=\max _{j \in (s,e )} \bm{B}_{t}^{(l, h)}(a_t,j)$. Similarly, the maximum attention score  over the textual regions can be obtained by
\begin{equation}\label{eq7} \small
    T^{(l,h)}_t=\max \left(\max _{j \in(0, s)} \bm{B}_{t}^{(l, h)}(a_t,j), \max _{j \in[e, a_t)} \bm{B}_{t}^{(l, h)}(a_t,j)\right).
\end{equation}
Based on $V^{(l,h)}_t$ and $T^{(l,h)}_t$, we define the text–visual dominance gap as $D^{(l,h)}_t=T^{(l,h)}_t-V^{(l,h)}_t$. A larger $D^{(l,h)}_t$ indicates stronger reliance on textual context and may correspond to a decoding regime where subsequent token generation is tightly constrained by the existing texts, as manifested by structural markers or end-of-sequence tokens. As shown in Fig. \ref{fig:roc}, we investigate the relationship between hallucination occurrence and two attention-based measures, namely the disparity between peak attention over visual and textual regions and the corresponding gap in their average attention. The results indicate that the peak-level measure possesses greater discriminative capability for identifying hallucinations. Consequently, selecting 
$D_t$ as the gating criterion is well motivated, as it provides an accurate indicator of hallucinations induced by textual inertia. Then, we introduce a gating function $G_{t}^{(l, h)}=\mathbb{I}(D_{t}^{(l, h)}<\tau)$ to selectively regulate attention adjustment. Accordingly, in regimes where textual attention outweighs visual perception, reinforcing visual attention helps alleviate hallucination risks arising from textual inertia. However, when textual attention becomes overly dominant, i.e., $D_{t}^{(l, h)} \geq \tau$, the decoding process should proceed freely to ensure fluent generation while avoiding repetitive outputs. After incorporating the dynamic gating mechanism, the attention modulation scheme for each image token can be expressed as
\begin{equation} \label{eq8} \small
\bm{\tilde{B}}_{t}^{(l, h)}\left(a_{t}, j\right)=\bm{{B}}_{t}^{(l, h)}\left(a_{t}, j\right)+G_{t}^{(l, h)}(\Delta_1^{(l)}+\Delta_2^{(l,h)}).
\end{equation}
It is noteworthy that we only perform the  correction operation in Eq. (\ref{eq8})  at the textual compilation layers, i.e., $l \in[l_e, L]$. Prior work has shown that earlier layers primarily function as modality fusion stages, which facilitates cross-modal alignment rather than supports high-level semantic reasoning~\cite{yin2025lifting,chen2025multimodal}. Guided by this insight, our approach operates attention adjustment only on the   intermediate and deeper layers. This design preserves the intrinsic modality alignment established in the shallow layers while calibrating the attention patterns in the reasoning-dominant layers. We further conduct a layer-wise ablation study in Appendix~\ref{layer-select}, which confirms that AFIP attains its optimal performance when applied to layers with $l \in[l_e, L]$. In summary, the procedure of our proposed AFIP is presented in Algorithm \ref{algorithm 1}. 
\begin{algorithm}[!ht]\caption{Attention-Focused Approach for Improved Image Perception (AFIP)} 
\label{algorithm 1}
\KwRequire MLLM $\mathcal{F}_{\theta}$, input image tokens $\{{v}_{1}, \ldots,{v}_{n}\}$, prompt tokens $\{{t}_{1}, \ldots, {t}_{m}\}$.

 \KwOutput Model response $y_t$ at the decoding step $t$. 

Store historical visual attention scores in the window $\mathcal{C}_{t}$.

\ForEach{$l$ in $[l_e,L]$}{ Identify the the top-$k$ attention heads $\mathcal{H}_{k}^{(l)}$;

Compute the mean and variance  value of the attention scores by Eq. (\ref{eq2});

Compute the term $\Delta_1^{(l)}$ by 
$Avg^{(l)}-\alpha   Var^{(l)}$;

Determine the most reliant text by Eq. (\ref{eq5});

Compute the term $\Delta_2^{(l,h)}$ by $\gamma \bm{C}^{(l, h)}_{t-1-r_{t}^{(l, h)}}(s+j)$;

Obtain the maximum attention score over the visual and textual region as $V^{(l,h)}_t$ and $T^{(l,h)}_t$, respectively;

\If{$T^{(l,h)}_t-V^{(l,h)}_t<\tau$}{
Conduct attention modulation for each layer and attention head by Eq. (\ref{eq8})}}

MLLM $\mathcal{F}_{\theta}$ decoding, obtain the current token $y_t$.
\end{algorithm}

\section{The Theoretical Discovery of Hallucination}
In this section, we offer a theoretical explanation for the emergence of hallucinations. By examining the structural properties of the attention layers and interpreting multi-head attention as diverse information from multiple sources, we demonstrate that inconsistencies in attention distributions across different heads can impair the model’s generalization capability, thereby increasing its vulnerability to hallucination behaviors.  The following conclusion encapsulates our central theoretical insights:
\begin{table*}[htbp]
\centering
\small
\setlength{\tabcolsep}{5pt}
\renewcommand{\arraystretch}{1.2}
\caption{CHAIR hallucination evaluation on five models with \textit{max new token} set to 512. ${\dagger}$ denotes using the greedy decoding strategy. $\Delta\%$ denotes the relative performance improvement with respect to the second-best method.}
\resizebox{\textwidth}{!}{
\begin{tabular}{l|
                c c c
                c c c
                c c c
                c c c
                c c c|
                c c}
\toprule[1.2pt]
\multirow{2}{*}{\textbf{Methods}} &
\multicolumn{3}{c}{\textbf{LLaVA-1.5-7B}} &
\multicolumn{3}{c}{\textbf{LLaVA-1.5-13B}} &
\multicolumn{3}{c}{\textbf{Shikra-7B}} &
\multicolumn{3}{c}{\textbf{MiniGPT-4-7B}} &
\multicolumn{3}{c|}{\textbf{Qwen-VL}} &
\multicolumn{2}{c}{\textbf{Avg.}} \\
\cmidrule(lr){2-4}\cmidrule(lr){5-7}\cmidrule(lr){8-10}
\cmidrule(lr){11-13}\cmidrule(lr){14-16}\cmidrule(lr){17-18}
& $C_S\downarrow$ & $C_I\downarrow$ & $\mathrm{Precision}\uparrow$
& $C_S\downarrow$ & $C_I\downarrow$ & $\mathrm{Precision}\uparrow$
& $C_S\downarrow$ & $C_I\downarrow$ & $\mathrm{Precision}\uparrow$
& $C_S\downarrow$ & $C_I\downarrow$ & $\mathrm{Precision}\uparrow$
& $C_S\downarrow$ & $C_I\downarrow$ & $\mathrm{Precision}\uparrow$
& $C_S\downarrow$ & $C_I\downarrow$ \\
\midrule

\rowcolor{gray!15}
\multicolumn{18}{c}{\textit{Decoding Strategy}} \\
\midrule
Greedy & 55.4 & 14.4 & 74.0 & 49.7 & 14.7 & 71.0 & 62.0 & 17.5 & 70.9 & 39.4 & 11.0 & 80.5 & 28.2 & 8.9 & 84.9 & 46.9 & 13.3 \\
Beam   & 56.2 & 15.1 & 74.0 & 50.4 & 15.2 & 76.3 & 59.2 & 16.2 & 74.2 & 39.2 &  12.2 & 81.7 & 30.0 & 10.7 & 83.8 & 47.0 & 13.9 \\
OPERA  & 48.2 & 13.1 & 76.9 & 41.3 & 14.1 & 77.2 & 41.9 & 13.8 & 72.1 & 30.9 &  11.2 & 77.5 & 29.6 & 9.5 & 79.4 & 38.4 & 12.3 \\
\midrule

\rowcolor{gray!15}
\multicolumn{18}{c}{\textit{Contrastive Decoding}} \\
\midrule
VCD$^{\dagger}$ & 54.6 & 17.3 & 70.1 & 57.2 & 16.1 & 71.3 & 56.4 & 15.5 & 75.2 & 41.4 & 12.6 & 68.2 & 45.4 & 18.1 & 70.0 & 51.0 & 15.9 \\
\midrule

\rowcolor{gray!15}
\multicolumn{18}{c}{\textit{Hidden states-intervention}} \\
\midrule
MemVR & 50.4 & 14.3 & 74.2 & 53.2 & 13.5 & 75.6 & 48.8 & 16.5 & 70.9 & 33.6 & 9.7 & 83.5 & 38.2 & 13.6 & 74.9 & 44.8 & 13.5 \\
\midrule

\rowcolor{gray!15}
\multicolumn{18}{c}{\textit{Attention-intervention}} \\
\midrule
HGAI$^{\dagger}$ & 24.6 & 6.4 & 86.8 & 27.1 & 7.9 & 85.1 & 32.8 & 10.4 & 81.0 & 28.0 & 7.5 & 85.1 & 20.9 & 5.2 & 89.1 & 26.7 & 7.5 \\
\midrule[1.1pt]

\textbf{Ours}$^{\dagger}$ &
\cellcolor{bestgreen}\textbf{16.8} & \cellcolor{bestgreen}\textbf{4.4} & \cellcolor{bestgreen}91.3 &
\cellcolor{bestgreen}\textbf{18.4} & \cellcolor{bestgreen}\textbf{6.4} & \cellcolor{bestgreen}89.2 &
\cellcolor{bestgreen}\textbf{18.0} & \cellcolor{bestgreen}\textbf{6.6} & \cellcolor{bestgreen}89.8 &
\cellcolor{bestgreen}\textbf{17.8} & \cellcolor{bestgreen}\textbf{7.3} & \cellcolor{bestgreen}89.5 &
\cellcolor{bestgreen}\textbf{16.6} & \cellcolor{bestgreen}\textbf{5.0} & \cellcolor{bestgreen}91.1 &
\cellcolor{bestgreen}\textbf{17.5} & \cellcolor{bestgreen}\textbf{5.9} \\
$\Delta\%$ &
\textcolor{mygreen}{$\downarrow$7.8\%} & \textcolor{mygreen}{$\downarrow$2.0\%} & \textcolor{mygreen}{$\uparrow$4.5\%} &
\textcolor{mygreen}{$\downarrow$8.7\%} & \textcolor{mygreen}{$\downarrow$1.5\%} & \textcolor{mygreen}{$\uparrow$4.1\%} &
\textcolor{mygreen}{$\downarrow$14.8\%} & \textcolor{mygreen}{$\downarrow$3.8\%} & \textcolor{mygreen}{$\uparrow$8.8\%} &
\textcolor{mygreen}{$\downarrow$10.2\%} & \textcolor{mygreen}{$\downarrow$0.2\%} & \textcolor{mygreen}{$\uparrow$4.4\%} & \textcolor{mygreen}{$\downarrow$4.3\%}
 & \textcolor{mygreen}{$\downarrow$0.2\%} &
\textcolor{mygreen}{$\uparrow$2.0\%} & \textcolor{mygreen}{$\downarrow$9.2\%} & \textcolor{mygreen}{$\downarrow$1.6\%} \\
\bottomrule[1.1pt]
\end{tabular}
}
\label{tab:main}
\vspace{-1.0em}
\end{table*}

\begin{theorem}
  \label{theorem1} 
Let $\mathcal{F}$ denote the function class induced by a Transformer architecture with self-attention as its core mechanism. The input sequence is 
$\bm{X}=\left[\bm{x}_{1}, \bm{x}_{2}, \ldots, \bm{x}_{t}\right]^{\top} \in \mathbb{R}^{t \times d}$, augmented with a classification token $\bm{x}_{cls} \in \mathbb{R}^{d}$ to produce a scalar output. For the $h$-th head, the value and query–key projections are represented by $\bm{W}_{V}^{(h)}  \in \mathbb    {R}^{d \times d_{v}}$ and $\bm{W}_{Q K}^{(h)} \in \mathbb{R}^{d \times d}$. The scalar  output is given by
\begin{equation} \small
    g(\bm{X})=\bm{w}^{\top} \bm{W}_{C}^{\top} \sigma\left(\bm{W}_{O}^{\top} \operatorname{concat}\left(z^{(1)}(\bm{X}), \ldots, z^{(H)}(\bm{X})\right)\right), 
\end{equation}
where the updated representation of the  $h$-th head is  $z^{(h)}(\bm{X})=\bm{W}_{V}^{(h)^{\top}} \bm{X}^{\top} \alpha^{(h)}(\bm{X})$, with attention weights defined by $\alpha^{(h)}(\bm{X})=\operatorname{softmax}\left(\bm{X} \bm{W}_{Q K}^{(h)^{\top}} \bm{x}_{cls}\right)$. Assume the parameter norms satisfy $\|\bm{w}\|_{2} \leq B_{w}$, $\left\|\bm{W}_{C}\right\|_F \leq B_{W C}$, $\left\|\bm{W}_{O}\right\|_F \leq B_{O}$, $\|\bm{W}_{V}^{(h)}\|_F \leq B_{W V}^{(h)}$, and $\|\bm{W}_{Q K}^{(h)}\|_F \leq B_{W Q K}^{(h)}$, which aligns with standard regularization practices to ensure stable training. Suppose the activation function $\sigma$ is  $L_{\sigma}$-Lipschitz and all input tokens are bound as
$ \left\|\bm{x}_{t}\right\|_{2} \leq R$. Then, the upper bound of the  Rademacher complexity $\hat{\Re}_{D}(\mathcal{F}) $ can be expressed as
\begin{equation}\label{eq34}
\hat{\Re}_{D}(\mathcal{F}) \leq \frac{4 B_{w} B_{W C} B_{O} L_{\sigma}R^3}{\sqrt{n}}\sqrt{H \bar{a}^{2}+\sum_{h=1}^{H}\left(a_{h}-\bar{a}\right)^{2}},
\end{equation} 
where $a_h=B_{W V}^{(h)}B_{W Q K}^{(h)}$ and 
$\bar{a}=\frac{1}{H} \sum_{h=1}^{H} a_{h}$. The upper bound  attains its minimum if and only if $a_{1}=\cdots=a_{H}=\bar{a}$, corresponding to uniform contributions from all attention heads. Any deviation from this regime introduces additional dispersion, enlarges the complexity and weakens generalization guarantees. This  finding formalizes the principle that maintaining homogeneity across multi-head attention supports optimal generalization performance.
\end{theorem}

\begin{theorem}
 \label{theorem2}
From a generalization analysis perspective, the representation captured at the $h$-th attention head is $z^{(h)}(\bm{X})\in \mathbb{R}^{d_{v}}$, and the target token to be predicted is denoted as $\bm{Y}$. After concatenating the  representations from all attention heads, the model produces a scalar output through the subsequent network $\psi: \mathbb{R}^{H d_{v}} \rightarrow \mathbb{R}$, yielding  $f(\bm{X})=\psi(z(\bm{X}))$. The training dataset $D=\{(\bm{X}_{i}, \bm{Y}_{i}): i \in[n]\}$ is drawn from a distribution over $\mathcal{X} \times \mathcal{Y}$, where $\mathcal{Y}$ denotes the corpus of $C$ words. Besides,
the expected risk and  empirical risk  w.r.t. the training dataset $D$ can be denoted as  $R({f})=\mathbb{E}_{(\bm{X}, \bm{Y}) \sim \mathcal{X} \times \mathcal{Y}}[l({f}(\bm{X}), \bm{Y})]$ and $\widehat{R}_{D}({f})=\frac{1}{N} \sum_{i=1}^{N}  l(f( \bm{X}_{i}), \bm{Y}_{i})$, respectively. With probability at least $1-\delta$, we have the following generalization error bound:
\begin{equation*}
\begin{aligned}
&R({f})-\widehat{R}_{D}({f})  \leq \frac{\widetilde{\mathcal{K}}_{1}}{n^{1/2}H^{-1/2}}+ \frac{\widetilde{\mathcal{K}}_{2}}{n^{3/4}H^{-1/4}}
\\&+\widetilde{\mathcal{K}}_{3}\sqrt{\frac{ \left(-\sum_{h=1}^{H}I(z^{(h)}(\bm{X}); \bm{Y})+\widetilde{\mathcal{K}}_{4}\right)}{nH}},
\end{aligned}
\end{equation*}
where $\widetilde{\mathcal{K}}_{1}=\mathcal{O}(H)$,  $\widetilde{\mathcal{K}}_{2}=\mathcal{O}(\sqrt{C \log C})$,
$\widetilde{\mathcal{K}}_{3}=\mathcal{O}(\sqrt{C}H)$,
$\widetilde{\mathcal{K}}_{4}$  is constant of order  $\widetilde{\mathcal{O}}(1)$ as $n, H \rightarrow \infty$.
Moreover, the generalization error bound becomes tighter as the visual evidence captured by each attention head aligns more closely with the target token. This conclusion suggests that strong generalization emerges when multiple heads maintain highly consistent semantic focus.
\end{theorem}

\section{Experiment}
\subsection{Experimental Setup}

\begin{table}[t]
\centering
\setlength{\tabcolsep}{5pt}
\renewcommand{\arraystretch}{1.1}
\caption{POPE hallucination evaluation  on five models with \textit{max new token} set to 512. ${\dagger}$ denotes using the greedy decoding strategy.}

\begin{adjustbox}{width=1\columnwidth}
\begin{tabular}{l|
                c c
                c c
                c c|
                c c}
\toprule[1.1pt]
\multirow{2}{*}{\textbf{Methods}} &
\multicolumn{2}{c}{\textbf{Random}} &
\multicolumn{2}{c}{\textbf{Popular}} &
\multicolumn{2}{c|}{\textbf{Adversarial}} &
\multicolumn{2}{c}{\textbf{Average}} \\
\cmidrule(lr){2-3}\cmidrule(lr){4-5}\cmidrule(lr){6-7}\cmidrule(lr){8-9}
& $\mathrm{Accuracy}\uparrow$ & $\mathrm{F1}\uparrow$
& $\mathrm{Accuracy}\uparrow$ & $\mathrm{F1}\uparrow$
& $\mathrm{Accuracy}\uparrow$ & $\mathrm{F1}\uparrow$
& $\mathrm{Accuracy}\uparrow$ & $\mathrm{F1}\uparrow$ \\
\midrule
\multicolumn{9}{c}{\textbf{LLaVA-1.5-7B}} \\
\midrule
Greedy & 89.20 & 89.12 & 85.77 & 84.33 & 79.63 & 81.26 & 84.87 & 84.90 \\
Beam   & 87.40 & 87.13 & 86.27 & 86.55 & 81.70 & 81.86 & 85.12 & 85.18 \\
VCD$^{\dagger}$ &
87.73 & 87.16 & 85.38 & 85.06 & 80.88 & 81.33 & 84.66 & 84.52 \\
MemVR  &
89.37 & 89.56 & 86.33 & 86.88 & 79.83 & 81.78 & 85.18 & 86.07 \\
HGAI$^{\dagger}$ & 89.70 & 89.42 & 86.97 & 86.98 & 80.50 & 81.63 & 85.72 & 86.01 \\
Ours$^{\dagger}$ &
\cellcolor{bestgreen}\textbf{90.77} & \cellcolor{bestgreen}\textbf{90.90} &
\cellcolor{bestgreen}\textbf{87.70} & \cellcolor{bestgreen}\textbf{88.24} &
\cellcolor{bestgreen}\textbf{82.87} & \cellcolor{bestgreen}\textbf{84.33} &
\cellcolor{bestgreen}\textbf{87.11} & \cellcolor{bestgreen}\textbf{87.82} \\

\midrule
\multicolumn{9}{c}{\textbf{LLaVA-1.5-13B}} \\
\midrule
Greedy & 89.93 & 90.16 & 85.97 & 86.80 & 81.23 & 83.29 & 85.71 & 86.75 \\
Beam   & 89.13 & 89.08 & 84.80 & 85.37 & 82.40 & 83.80 & 85.44 & 86.08 \\
VCD$^{\dagger}$ &
87.39 & 86.55 & 85.74 & 85.06 & 81.92 & 81.78 & 85.02 & 84.46 \\
MemVR  &
88.20 & 88.53 & 87.10 & 85.33 & 81.23 & 82.40 & 85.51 & 85.42 \\
HGAI$^{\dagger}$ & 89.93 & 90.10 & 84.87 & 85.82 & 80.67 & 82.57 & 85.16 & 86.16 \\
Ours$^{\dagger}$ &
\cellcolor{bestgreen}\textbf{91.20} & \cellcolor{bestgreen}\textbf{91.29} &
\cellcolor{bestgreen}\textbf{87.97} & \cellcolor{bestgreen}\textbf{88.46} &
\cellcolor{bestgreen}\textbf{83.57} & \cellcolor{bestgreen}\textbf{84.87} &
\cellcolor{bestgreen}\textbf{87.80} & \cellcolor{bestgreen}\textbf{86.90} \\

\midrule

\multicolumn{9}{c}{\textbf{Shikra-7B}} \\
\midrule
Greedy & 84.50 & 84.76 & 83.53 & 83.96 & 79.67 & 80.87 & 82.57 & 83.20 \\
Beam   & 84.90 & 82.75 & 83.46 & 84.26 & 79.93 & 80.94 & 82.76 & 82.65 \\
VCD$^{\dagger}$ & 84.50 & 84.76 & 83.53 & 83.96 & 79.67 & 80.87 & 82.79 & 83.20 \\
MemVR  &
84.76 & 85.04 & 83.86 & 84.30 & 79.76 & 81.01 & 82.79 & 83.45 \\
HGAI$^{\dagger}$ & 87.06 & 86.36 & 84.13 & 83.77 & 82.43 & 82.32 & 84.54 & 84.15 \\
Ours$^{\dagger}$ &
\cellcolor{bestgreen}\textbf{87.47} & \cellcolor{bestgreen}\textbf{87.08} &
\cellcolor{bestgreen}\textbf{86.17} & \cellcolor{bestgreen}\textbf{85.93} &
\cellcolor{bestgreen}\textbf{87.07} & \cellcolor{bestgreen}\textbf{86.36} &
\cellcolor{bestgreen}\textbf{86.90} & \cellcolor{bestgreen}\textbf{86.46} \\
\midrule

\multicolumn{9}{c}{\textbf{MiniGPT-4-7B}}  \\
\midrule
Greedy & 81.73 & 79.36 & 75.50 & 72.14 & 74.20 & \textbf{73.27} & 77.14 & 74.92 \\
Beam   & 80.63 & 77.36 & \textbf{76.40} & 74.86 & 72.00 & 68.84 & 76.34 & 73.69 \\
VCD$^{\dagger}$ &
\textbf{83.33} & \textbf{83.30} & 72.80 & \textbf{75.35} & 69.87 & 73.43 & 75.33 & \textbf{77.36} \\
MemVR  &
83.00 & 81.25 & 75.40 & 74.97 & 72.13 & 72.61 & 76.84 & 76.28 \\
HGAI$^{\dagger}$ & 80.27 & 80.87 & 71.23 & 74.35 & 69.20 & 73.09 & 73.57 & 76.10 \\
Ours$^{\dagger}$ &
\cellcolor{bestgreen}82.17 & \cellcolor{bestgreen}79.73 &
\cellcolor{bestgreen}76.30 & \cellcolor{bestgreen}74.74 &
\cellcolor{bestgreen}\textbf{74.23} & \cellcolor{bestgreen}73.24 &
\cellcolor{bestgreen}\textbf{77.57} & \cellcolor{bestgreen}75.90 \\
\bottomrule[1.1pt]
\end{tabular}
\end{adjustbox}

\label{tab:methods_ordered}
\vspace{-1.5em}
\end{table}

\noindent \textbf{Model Backbones.} We assess the efficacy of our approach on four representative MLLMs, i.e.,  LLaVA-1.5~\cite{liu2024improved}, Shikra~\cite{chen2023shikra}, MiniGPT-4~\cite{zhu2023minigpt}, and Qwen-VL~\cite{bai2023qwenvlversatilevisionlanguagemodel}. To probe the influence of model scale, we evaluate both the 7B and 13B variants of LLaVA-1.5. Furthermore, by encompassing models with diverse alignment mechanisms, such as Q-Former-based designs and encoder-based approaches, we can further demonstrate the robustness of our method.

\noindent \textbf{Baselines.}
Our method is compared with baselines covering different categories of hallucination mitigation strategies.  Greedy decoding and beam search serve as standard decoding methods~\cite{sutskever2014sequence}. Additional baselines include OPERA~\cite{huang2024opera}, a beam-search-based method, HGAI~\cite{jiang2025devils}, an attention intervention approach, VCD~\cite{leng2024mitigating}, a contrastive decoding strategy, and MemVR~\cite{zou2024look}, which performs hidden-state intervention. Beam search and OPERA employ a beam size of 5, while VCD, MemVR and HGAI are applied at inference with default hyperparameters.

\noindent \textbf{Benchmark}
Following~\cite{huang2024opera,jiang2025devils}, we evaluate our method on complementary benchmarks with distinct task types. We adopt POPE~\cite{li2023evaluating}, a standardized protocol for short-horizon binary question answering. For long-form generation, we employ CHAIR~\cite{rohrbach2018object}, which measures hallucination severity based on the proportion of generated objects absent from ground-truth annotations. Additional details are provided in Appendix~\ref{benchmarks}.
 
\subsection{Comparative Experiment}
We conduct hallucination evaluations on the CHAIR and POPE benchmarks, and present the corresponding results in Tables~\ref{tab:main} and \ref{tab:methods_ordered}. Besides, to  evaluate the impact of generated token length,  experiments are performed by varying the \textit{max new token} among $\{64,128,256\}$, with outcomes reported in Table~\ref{tab:token_len_precision}. From these comparative results, we can derive the following observations: 
(i) Our method substantially reduces hallucination compared with original model responses.  As shown in Table \ref{tab:main}, relative to the vanilla decoding strategies Greedy and Beam, our approach achieves over 50\% improvements on both $C_I$ and $C_S$ metrics across LLaVA-1.5 and Qwen-VL. (ii) Under diverse evaluation criteria, AFIP demonstrates consistently superior performance. It ranks first across all metrics and models on the CHAIR benchmark,  and attains leading results on POPE  under the Random, Popular, and Adversarial settings for all models except MiniGPT-4. (iii) As reported in Table \ref{tab:token_len_precision}, AFIP remains robust across varying generation lengths. Besides, it achieves more pronounced gains in long-sequence tasks. This finding shows that continuously reinforcing visual perception during long-form generation is essential for achieving stronger resistance to hallucination. (iv) Compared with different baselines, our approach exhibits clear advantages, reflecting its comprehensive consideration of hallucination causes across both spatial and temporal dimensions. Unlike VCD and HGAI, which focus on contrastive decoding and  attention control, our method reinjects historical visual information to counteract temporal degradation and maintain sustained visual grounding. Compared with OPERA and MemVR, AFIP corrects attention inconsistencies across multiple heads to enhance visual reliability.

\subsection{Parameter Sensitivity Analysis}
\begin{figure}[htbp]
    \centering
    \begin{subfigure}[b]{0.23\textwidth}
        \centering
\includegraphics[width=\linewidth]{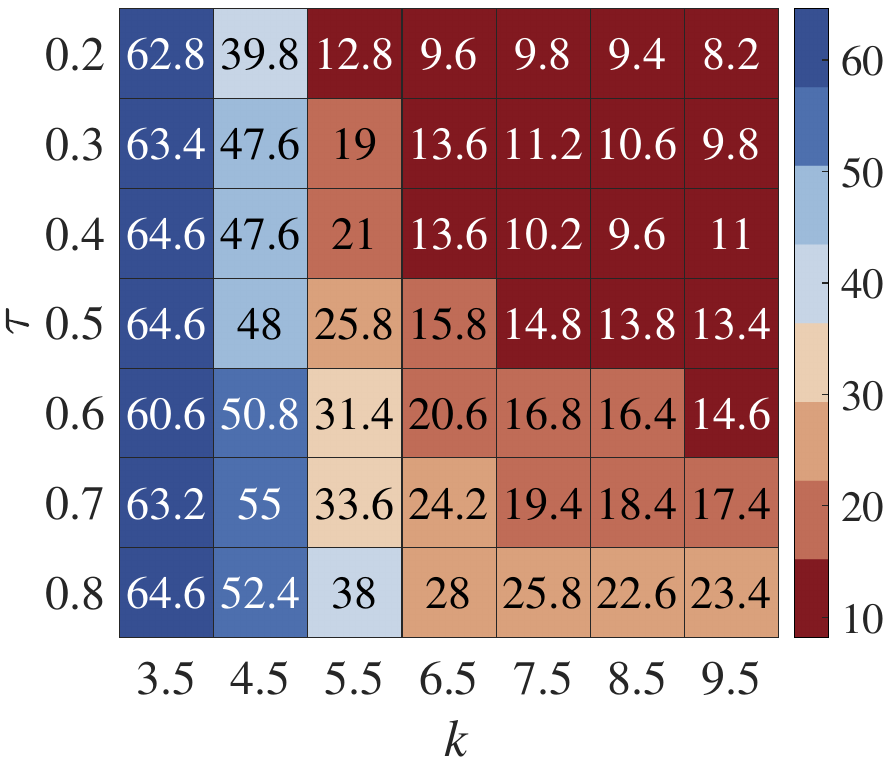}
         \caption{LLaVA-1.5-7b}
        \label{fig:var7b}
    \end{subfigure}
    \hfill
    \begin{subfigure}[b]{0.23\textwidth}
        \centering        \includegraphics[width=\linewidth]{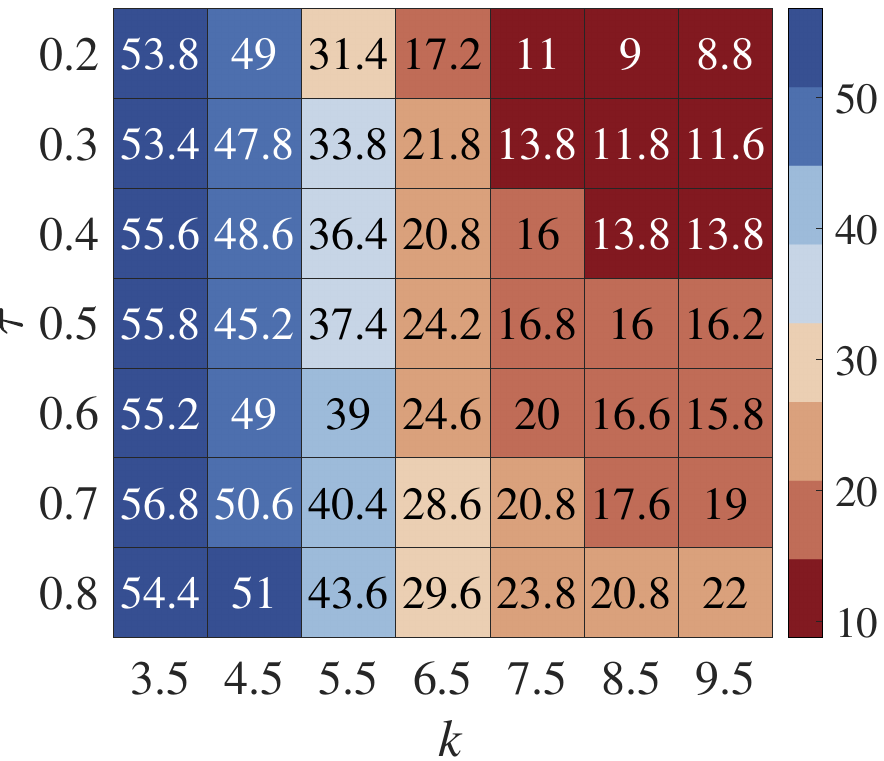}
        \caption{LLaVA-1.5-13b}
        \label{fig:var13b}
    \end{subfigure}
    \caption{Parameter sensitivity analysis of $k$ and $\tau$.}
    \label{Parameter}
\end{figure}

In our proposed AFIP, there are four hyperparameters in total, i.e., the coefficient 
$\alpha$ for the variance regularization term, the coefficient $\gamma$ for the historical attention enhancement term,  the Top-$k$ head selection and the threshold $\tau$ for the dynamic gating mechanism. As evidenced by the parameter analysis results in Appendix~\ref{para}, AFIP is largely insensitive to 
$\alpha$ and $\gamma$. Consequently, we focus on investigating the joint effect of 
$k$ and 
$\tau$ on the effectiveness of hallucination mitigation. Specifically, $k$ is evaluated over the discrete set $\{3.5,4.5,5.5,6.5,7.5,8.5,9.5\}$, while $\tau$ is varied across $\{0.2,0.3,0.4,0.5,0.6,0.7,0.8\}$. The results under different parameter combinations are summarized in the heatmap, as shown in Fig.~\ref{Parameter}.
Analysis of these results indicates that optimal performance is generally obtained when $k$ falls within $[5.5,7.5]$ and $\tau$ lies in the range $[0.4,0.6]$. Careful tuning of $k$ and $\tau$ is crucial, as the confidence levels of the top-ranked attention heads and the corresponding threshold estimates can vary substantially across different models.

\begin{table}[t]
\centering
\small
\setlength{\tabcolsep}{5pt}
\caption{CHAIR hallucination evaluation on LLaVA-1.5-7B and Qwen-VL with varying \textit{max new token} in \{64, 128, 256\}. ${\dagger}$ denotes using the greedy decoding strategy.}
\renewcommand{\arraystretch}{1.2}
\begin{adjustbox}{width=1\columnwidth}
\begin{tabular}{l
                c c c
                c c c
                c c c|
                c c}
\toprule
\multirow{2}{*}{\textbf{Method}} &
\multicolumn{3}{c}{\textbf{\textit{max new token 64}}} &
\multicolumn{3}{c}{\textbf{\textit{max new token 128}}} &
\multicolumn{3}{c|}{\textbf{\textit{max new token 256}}} &
\multicolumn{2}{c}{\textbf{Avg.}} \\
\cmidrule(lr){2-4}\cmidrule(lr){5-7}\cmidrule(lr){8-10}\cmidrule(lr){11-12}
& $C_S\downarrow$ & $C_I\downarrow$ & Precision$\uparrow$
& $C_S\downarrow$ & $C_I\downarrow$ & Precision$\uparrow$
& $C_S\downarrow$ & $C_I\downarrow$ & Precision$\uparrow$
& $C_S\downarrow$ & $C_I\downarrow$ \\
\midrule
\multicolumn{12}{c}{\textbf{LLaVA-1.5-7B}} \\
\midrule
Greedy &
22.4 & 6.4 & \cellcolor{gray!15}88.9 &
49.4 & 13.1 & \cellcolor{gray!15}76.4 &
50.0 & 13.1 & \cellcolor{gray!15}76.2 &
40.6 & 10.9 \\

Beam &
19.2 & 5.9 & \cellcolor{gray!15}90.4 &
51.6 & 13.8 & \cellcolor{gray!15}75.0 &
52.4 & 14.5 & \cellcolor{gray!15}73.8 &
41.1 & 11.4 \\

VCD$^{\dagger}$ &
29.2 & 9.7 & \cellcolor{gray!15}84.5 &
58.8 & 17.5 & \cellcolor{gray!15}70.1 &
54.6 & 17.3 & \cellcolor{gray!15}71.3 &
47.5 & 14.8 \\

MemVR &
24.0 & 7.3 & \cellcolor{gray!15}87.4 &
42.6 & 14.2 & \cellcolor{gray!15}74.7 &
43.4 & 14.3 & \cellcolor{gray!15}77.2 &
36.7 & 11.9 \\

HGAI$^{\dagger}$ &
18.6 & 5.3 & \cellcolor{gray!15}90.5 &
25.2 & 6.9 & \cellcolor{gray!15}70.1 &
26.6 & 7.4 & \cellcolor{gray!15}86.2 &
23.5 & 6.5 \\

\midrule
Ours$^{\dagger}$ &
\textbf{13.0} & \textbf{4.0} & \cellcolor{gray!15}92.6 &
\textbf{21.6} & \textbf{4.9} & \cellcolor{gray!15}89.3 &
\textbf{15.0} & \textbf{3.6} & \cellcolor{gray!15}92.0 &
\textbf{16.5} & \textbf{4.2} \\
\midrule
\multicolumn{12}{c}{\textbf{Qwen-VL}} \\
\midrule
Greedy &
19.6 & 10.6 & \cellcolor{gray!15}86.0 &
21.8 & 11.7 & \cellcolor{gray!15}84.1 &
28.2 & 14.7 & \cellcolor{gray!15}79.4 &
23.2 & 12.3 \\

Beam &
20.2 & 7.2 & \cellcolor{gray!15}88.9 &
23.2 & 7.5 & \cellcolor{gray!15}85.5 &
27.2 & 8.5 & \cellcolor{gray!15}82.5 &
23.5 & 7.7 \\

VCD$^{\dagger}$ &
19.2 & 9.8 & \cellcolor{gray!15}83.6 &
39.0 & 15.8 & \cellcolor{gray!15}76.5 &
41.4 & 15.6 & \cellcolor{gray!15}73.8 &
33.2 & 13.7 \\

MemVR &
25.0 & 7.7 & \cellcolor{gray!15}87.1 &
27.4 & 13.2 & \cellcolor{gray!15}75.5 &
32.3 & 11.6 & \cellcolor{gray!15}75.7 &
28.2 & 10.8 \\

HGAI$^{\dagger}$ &
20.8 & 10.7 & \cellcolor{gray!15}85.8 &
16.4 & 7.4 & \cellcolor{gray!15}90.2 &
15.4 & 6.7 & \cellcolor{gray!15}90.6 &
17.5 & 8.3 \\

\midrule
Ours$^{\dagger}$ &
\textbf{14.4} & \textbf{6.6} & \cellcolor{gray!15}90.0 &
\textbf{15.6} & \textbf{7.2} & \cellcolor{gray!15}89.7 &
\textbf{12.2} & \textbf{6.0} & \cellcolor{gray!15}91.7 &
\textbf{14.1} & \textbf{6.6} \\

\bottomrule
\end{tabular}
\end{adjustbox}
\label{tab:token_len_precision}
\end{table}

\section{Conclusion}
This paper proposes an Attention-Focused Approach for Improved Image Perception named AFIP to mitigate hallucinations in MLLMs. Extensive statistical analyses reveal that model hallucinations mainly arise from spatial inconsistencies in multi-head visual attention and temporal degradation in image perception, a pattern analogous to visual blur induced by attention distraction  in human perception. We further theoretically demonstrate that attention distraction undermines the model’s discriminative capacity during vocabulary selection. Moreover, AFIP is introduced as a plug-and-play module that incurs no additional inference latency, while extensive experiments demonstrate its effectiveness in correcting head-level attention deviations and alleviating the temporal decay of visual attention.
\section*{Acknowledgments}
This work was partially supported by the NSF for Distinguished Young Scholars under Grant No. 62425607, the Key
NSF of China under Grant No. 62136005,  Hunan Provincial Natural Science Outstanding Youth Fund under Grant No. 2026JJ20066 and the National Natural Science Foundation of China Grant No.\ 62376281.

\section*{Impact Statement}
This paper aims to advance the field of machine learning. As our work focuses on mitigating hallucination in Multimodal large language models (MLLMs), it may give rise to a range of societal implications. Overall, the primary impact of this research is to promote the safer and more reliable deployment of MLLMs across diverse application domains. We do not anticipate that the proposed method introduces additional ethical concerns or poses risks related to privacy.
\nocite{langley00}

\bibliography{ref}
\bibliographystyle{icml2026}

\newpage
\appendix
\onecolumn
\section{Proof of The Theorem \ref{theorem1}}
We begin the proof of Theorem \ref{theorem1} by introducing the following lemmas:
\begin{lemma}[\cite{edelman2022inductive}]
\label{lemma1}
For vectors $\theta_{1}, \theta_{2} \in \mathbb{R}^{p}$, we have
\begin{equation}
\left\|\operatorname{softmax}\left(\theta_{1}\right)-\operatorname{softmax}\left(\theta_{2}\right)\right\|_{1} \leq 2\left\|\theta_{1}-\theta_{2}\right\|_{\infty}.    
\end{equation}
\end{lemma}
\begin{lemma}[\cite{lust1991non}] \label{lemma2}
Consider vectors $\bm{v}_{1}, \ldots, \bm{v}_{n} \in \mathcal{H}$, where  $\mathcal{H}$ is a Hilbert space with $\|\cdot\|$ being the associated $p$-th norm. Let $\epsilon_{1}, \ldots, \epsilon_{n}$ be a  sequence of independent Rademacher variables. $\bm{v}_{1}, \ldots, \bm{v}_{n} \in \mathcal{H}$, where  $\mathcal{H}$ is a Hilbert space with $\|\cdot\|$ being the associated $p$-th norm. Let $\epsilon_{1}, \ldots, \epsilon_{n}$ be a  sequence of independent Rademacher variables. Then, for any $p \geq 1$, we have
\begin{equation}
 \min (\sqrt{p-1}, 1)\left[\sum_{i=1}^{n}\left\|\bm{v}_{i}\right\|^{2}\right]^{\frac{1}{2}} \leq\left[\mathbb{E}_{\bm{\epsilon}}\left\|\sum_{i=1}^{n} \epsilon_{i} \bm{v}_{i}\right\|^{p}\right]^{\frac{1}{p}} \leq \max (\sqrt{p-1}, 1)\left[\sum_{i=1}^{n}\left\|\bm{v}_{i}\right\|^{2}\right]^{\frac{1}{2}},   
\end{equation}
and
\begin{equation}
\mathbb{E}_{\bm{\epsilon}}\left\|\sum_{i=1}^{n} \epsilon_{i} \bm{v}_{i}\right\| \geq 2^{-\frac{1}{2}}\left[\sum_{i=1}^{n}\left\|\bm{v}_{i}\right\|^{2}\right].
\end{equation}

\end{lemma}
We adopt the definitions and notation for self-attention and Transformers introduced in \cite{edelman2022inductive}. Accordingly, we disregard other architectural components of the Transformer, such as normalization and feed-forward networks, and concentrate solely on the role of the attention layer in shaping generalization behavior. The input sequence is denoted as $\bm{X}=\left[\bm{x}_{1}, \bm{x}_{2}, \ldots, \bm{x}_{t}\right]^{\top} \in \mathbb{R}^{t \times d}$, which contains $t$ $d$-dimensional tokens.  Additionally, for notational convenience, the $t$-th token of the 
$i$-th sample is denoted by $\bm{x}^i_t$. Suppose there are 
$H$ attention heads, For the 
$h$-th head, the query, key, and value are denoted as $\bm{W}_{Q}^{(h)} \in \mathbb{R}^{d \times d_{v}}$, $\bm{W}_{K}^{(h)} \in \mathbb{R}^{d \times d_{v}}$, $\bm{W}_{V}^{(h)} \in \mathbb{R}^{d \times d_{v}}$, respectively. Moreover, we combine the query and key and write them as $\bm{W}_{Q K}^{(h)} \in \mathbb{R}^{d \times d}$. To enable a principled generalization analysis, we further adopt the standard classification-token (CLS) construction. Specifically, we prepend a special token $\bm{x}_{{cls}} \in \mathbb{R}^d$ to the input sequence and define the model output solely based on the representation associated with this token. Concretely, the CLS token serves as a global query that attends to all input tokens through the self-attention mechanism, thereby aggregating sequence-level information into a single vector. As a consequence, the norm of the aggregated representation can be controlled independently of the sequence length. From a functional perspective, introducing the CLS token does not reduce the expressive power of the model for sequence-level tasks. Any attention-based pooling or global aggregation over tokens can be equivalently implemented by an appropriately parameterized CLS query. Instead, the CLS construction internalizes the pooling operation into the attention layer itself, yielding a single vector representation that naturally supports scalar-valued prediction. This scalar-output formulation is crucial for complexity analysis, as it allows the Transformer to be viewed as a real-valued function class, enabling the application of Rademacher complexity and related generalization bounds. Such a setup is standard in practice and is employed in BERT \cite{devlin2019bert}. Then, we can obtain the attention weights for the $h$-th head as
\begin{equation}
\alpha^{(h)}(\bm{X})=\operatorname{softmax}\left(\bm{X} \bm{W}_{Q K}^{(h)^{\top}} \bm{x}_{cls}\right) \in \mathbb{R}^{t},   
\end{equation}
where the  $\operatorname{softmax}$ function is applied row-wise, which ensures that $\alpha_{t}^{(h)}(\bm{X}) \geq 0$ and $ \sum_{t=1}^{T} \alpha_{t}^{(h)}(\bm{X})=1$. Hence, the output corresponding to the $h$-th head can be expressed as
\begin{equation}
 z^{(h)}(\bm{X})=\bm{W}_{V}^{(h)^{\top}} \bm{X}^{\top} \alpha^{(h)}(\bm{X}) \in \mathbb{R}^{d_{v}}.  
\end{equation}
By concatenating the outputs of all heads, we can obtain
\begin{equation}
z(\bm{X})=\operatorname{concat}\left(z^{(1)}(\bm{X}), \ldots, z^{(H)}(\bm{X})\right) \in \mathbb{R}^{H d_{v}}.   
\end{equation}
Subsequently, applying the projection matrix of the multi-head output layer yields the following output:
\begin{equation}
u(\bm{X})=\bm{W}_{O}^{\top} z(\bm{X}) \in \mathbb{R}^{d}, 
\end{equation}
where $\bm{W}_{O} \in \mathbb{R}^{H d_{v} \times d}$.
Finally, through the combined effect of the activation function and the classification head, the following scalar output is obtained:
\begin{equation}
g(\bm{X})=\bm{w}^{\top} \bm{W}_{C}^{\top} \sigma(u(\bm{X})),    
\end{equation}
where $\bm{w} \in \mathbb{R}^{d}$ and $ \bm{W}_{C} \in \mathbb{R}^{d \times d}$. Besides, $\sigma$ denotes an element-wise activation function.

Next, we impose appropriate bounds on the input magnitude and the norms of the model parameters:
\begin{assumption}
  \label{assumption1}
There exists a constant $R>0$, such that, for all input tokens, the embeddings satisfy
\begin{equation} \left\|\bm{x}_{t}\right\|_{2} \leq R, \quad\left\|\bm{x}_{cls}\right\|_{2} \leq R .
\end{equation}
\end{assumption}

\begin{assumption}
 \label{assumption2}
There exist positive constants $B_w$, $B_{WC}$, and $B_O$, as well as head-specific constants $B^{(h)}_{WV}$ and $B^{(h)}_{WQK}$, such that the corresponding norms of the model parameters are bounded as follows:
\begin{equation}\label{eq18}
\left\{\begin{array}{l}
 \|\bm{w}\|_{2} \leq B_{w}, \quad\left\|\bm{W}_{C}\right\|_F \leq B_{W C}, \quad\left\|\bm{W}_{O}\right\|_F \leq B_{O},  \\
\|\bm{W}_{V}^{(h)}\|_F \leq B_{W V}^{(h)}, \quad\|\bm{W}_{Q K}^{(h)}\|_F \leq B_{W Q K}^{(h)}, \quad h=1, \ldots, H .
\end{array}\right.
\end{equation}
\end{assumption}

\begin{assumption}
  \label{assumption3}
  The activation function $\sigma: \mathbb{R} \rightarrow \mathbb{R}$ is  $L_{\sigma}$-Lipschitz and $\sigma(0)=0$.
\end{assumption}

Let $\mathcal{F}$ denote the real-valued function class induced by a Transformer architecture with self-attention as its core mechanism.
Then,  we can get the Rademacher complexity of the scalar one layer Transformer  as below:
\begin{equation}
 \hat{\Re}_{D}(\mathcal{F})=\mathbb{E}_{\bm{\epsilon}}\left[\sup _{f(\bm{X}_i)\in \mathcal{F}} \frac{1}{n} \sum_{i=1}^{n} \epsilon_{i} \bm{w}^{\top} W_{C}^{\top} \sigma\left(\bm{W}_{O}^{\top}\operatorname{concat}\left(z^{(1)}(\bm{X}_i), \ldots, z^{(H)}(\bm{X}_i)\right) \right)\right].   
\end{equation}
According to the Assumption \ref{assumption2} and the Cauchy–Schwarz inequality, we can upper bound $ \hat{\Re}_{D}(\mathcal{F})$:
\begin{equation}
\begin{aligned}
\hat{\Re}_{D}(\mathcal{F})&=\mathbb{E}_{\bm{\epsilon}}\left[\sup _{\|\bm{w}\|_2 \leq B_{\bm{w}}, \sigma} \frac{1}{n} \sum_{i=1}^{n} \epsilon_{i}\left\langle\bm{w}, \bm{W}_{C}^{\top} \sigma\left(\bm{W}_{O}^{\top}\operatorname{concat}\left(z^{(1)}(\bm{X}_i), \ldots, z^{(H)}(\bm{X}_i)\right) \right)\right\rangle\right] \\ &
\leq B_{\bm{w}} \mathbb{E}_{\bm{\epsilon}} \left[\sup _{\left\|\bm{W}_{C}\right\|_F \leq B_{WC}, \sigma} \frac{1}{n}\left\|\sum_{i=1}^{n} \epsilon_{i}\bm{W}_{C}^{\top} \sigma\left(\bm{W}_{O}^{\top}\operatorname{concat}\left(z^{(1)}(\bm{X}_i), \ldots, z^{(H)}(\bm{X}_i)\right) \right)\right\|_2\right]\\ &
\leq \frac{B_{\bm{w}} B_{W_{C}}}{n} \mathbb{E}_{\bm{\epsilon}} \left[\sup _{\sigma}\left\|\sum_{i=1}^{n} \epsilon_{i} \sigma
\left(\bm{W}_{O}^{\top}\operatorname{concat}\left(z^{(1)}(\bm{X}_i), \ldots, z^{(H)}(\bm{X}_i)\right) \right)
\right\|_2\right].
\end{aligned}
\end{equation}
According to the Assumption \ref{assumption3}, we can further get
\begin{equation}\label{eq22}
\begin{aligned}
 \hat{\Re}_{D}(\mathcal{F}) &\leq \frac{2B_{\bm{w}} B_{W_{C}}L_{\sigma}}{n} \mathbb{E}_{\bm{\epsilon}} \left[\sup _{\left\|\bm{W}_{O}\right\|_F \leq B_{O}}\left\|\sum_{i=1}^{n} \epsilon_{i}
\bm{W}_{O}^{\top}\operatorname{concat}\left(z^{(1)}(\bm{X}_i), \ldots, z^{(H)}(\bm{X}_i)\right) 
\right\|_2\right]\\&
\leq \frac{2B_{\bm{w}} B_{W_{C}}B_{O}L_{\sigma}}{n} \mathbb{E}_{\bm{\epsilon}} \left[\sup _{}\left\|\sum_{i=1}^{n} \epsilon_{i}
\operatorname{concat}\left(z^{(1)}(\bm{X}_i), \ldots, z^{(H)}(\bm{X}_i)\right) 
\right\|_2\right].
\end{aligned}
\end{equation}
Then, we define $S_h$ as the aggregate sum over all samples at the $h$-th head, with respect to the function $z^{h}$, i,e., $S_{h}=\sum_{i=1}^{n} \varepsilon_{i} z^{(h)}\left(\bm{X}_{i}\right) \in \mathbb{R}^{d_{v}}$. Since $z(\bm{X}_i)=\operatorname{concat}\left(z^{(1)}(\bm{X}_i), \ldots, z^{(H)}(\bm{X}_i)\right)$, we have
\begin{equation}
 \left\|\sum_{i=1}^{n} \varepsilon_{i} z\left(\bm{X}_{i}\right)\right\|^{2}_2=\sum_{h=1}^{H}\left\|S_{h}\right\|^{2}_2.
\end{equation}
Based on Eq. (\ref{eq22}),  we arrive at the following inequality:
\begin{equation}
\sup \left\|\sum_{i=1}^{n} \varepsilon_{i} z\left(X_{i}\right)\right\|_2 \leq\left(\sum_{h=1}^{H}\left(\sup \left\|S_{h}\right\|_2^2\right)\right)^{1 / 2}.
\end{equation}
Consequently, we further obtain the following bound on $\hat{\Re}_{D}(\mathcal{F})$:
\begin{equation}\label{eq24}
   \hat{\Re}_{D}\left(\mathcal{F}\right) \leq \frac{2 B_{w} B_{W C} B_{O} L_{\sigma}}{n} \mathbb{E}_{\varepsilon}\left(\sum_{h=1}^{H}\left(\sup \left\|S_{h}\right\|_2^2\right)\right)^{1 / 2}. 
\end{equation}
Regarding the head-specific term $S_h$, we obtain the following estimate:
\begin{equation}
\begin{aligned}
\mathbb{E}_{\bm{\epsilon}} \left[\sup \left\|S_{h}\right\|_2\right] &=\mathbb{E}_{\bm{\epsilon}} \left[\sup \left\|\sum_{i=1}^{n} \varepsilon_{i} z^{(h)}\left(\bm{X}_{i}\right)\right\|_2\right]
\\&
=\mathbb{E}_{\bm{\epsilon}} \left[\sup \left\|\sum_{i=1}^{n} \varepsilon_{i} \bm{W}_{V}^{(h)^{\top}} \bm{X}^{\top}_i \alpha^{(h)}(\bm{X}_i) \right\|_2\right]
\\&
=\mathbb{E}_{\bm{\epsilon}} \left[\sup \left\|\sum_{i=1}^{n} \varepsilon_{i} \bm{W}_{V}^{(h)^{\top}} \bm{X}^{\top}_i \operatorname{softmax}\left(\bm{X}_i \bm{W}_{Q K}^{(h)^{\top}} \bm{x}_{cls}\right)\right\|_2\right] \\&
\leq  B_{W V}^{(h)}\mathbb{E}_{\bm{\epsilon}} \left[ \sup_{\|\bm{W}_{Q K}^{(h)}\|_F \leq B_{W Q K}^{(h)}}\left\|\sum_{i=1}^{n} \varepsilon_{i} \bm{X}^{\top}_i \operatorname{softmax}\left(\bm{X}_i \bm{W}_{Q K}^{(h)^{\top}} \bm{x}_{cls}\right)\right\|_2 \right]
\end{aligned}.
\end{equation}
Using the inequality $\|\bm{Q} \bm{x}\| \leq\|\bm{Q}\|_{2, \infty}\|\bm{x}\|_{1}$, we can get
\begin{equation}
\begin{aligned}
\mathbb{E}_{\bm{\epsilon}}\left[\sup \left\|S_{h}\right\|_2\right] & \leq B_{W V}^{(h)}\mathbb{E}_{\bm{\epsilon}}\left[ \sup _{\bm{W}_{Q K}^{(h)}\leq B_{W Q K}^{(h)}}\left\|\sum_{i=1}^{n} \epsilon_{i} \bm{X}_{i}^{\top}\right\|_{2, \infty}\left\|\operatorname{softmax}\left(\bm{X}_i \bm{W}_{Q K}^{(h)^{\top}} \bm{x}_{cls}\right)\right\|_{1}\right]\\ &
\leq B_{W V}^{(h)} \sup _{\bm{W}_{Q K}^{(h)}\leq B_{W Q K}^{(h)}}\left\|\operatorname{softmax}\left(\bm{X}_i \bm{W}_{Q K}^{(h)^{\top}} \bm{x}_{cls}\right)\right\|_{1}\mathbb{E}_{\bm{\epsilon}}\left[\left\|\sum_{i=1}^{n} \epsilon_{i} \bm{X}_{i}^{\top}\right\|_{2, \infty}\right].
\end{aligned}
\end{equation}
By applying the Lemma \ref{lemma2}, the following result is derived:
\begin{equation}\label{eq27}
\begin{aligned}
    \mathbb{E}_{\bm{\epsilon}} \left[\sup \left\|S_{h}\right\|_2\right]&  \leq 2B_{W V}^{(h)}\sup _{\bm{W}_{Q K}^{(h)}\leq B_{W Q K}^{(h)}}\left\|\bm{X}_i \bm{W}_{Q K}^{(h)^{\top}} \bm{x}_{cls}\right\|_{\infty}\mathbb{E}_{\bm{\epsilon}}\left[\left\|\sum_{i=1}^{n} \epsilon_{i} \bm{X}_{i}^{\top}\right\|_{2, \infty}\right]\\&
    =2B_{W V}^{(h)}\sup _{\bm{W}_{Q K}^{(h)}\leq B_{W Q K}^{(h)}}\max _{t}\left\|\bm{x}_{t}^{i^{\top}} W_{Q K}^{(h)^{\top}} \bm{x}_{cls}\right\|_2\mathbb{E}_{\bm{\epsilon}}\left[\left\|\sum_{i=1}^{n} \epsilon_{i} \bm{X}_{i}^{\top}\right\|_{2, \infty}\right] \\&
    \leq 2B_{W V}^{(h)}\sup _{\bm{W}_{Q K}^{(h)}\leq B_{W Q K}^{(h)}}\max _{t}
    \left\|W_{Q K}^{(h)}\bm{x}_{t}^{i}\right\|_2
\left\|\bm{x}_{cls}\right\|_2\mathbb{E}_{\bm{\epsilon}}\left[\left\|\sum_{i=1}^{n} \epsilon_{i} \bm{X}_{i}^{\top}\right\|_{2, \infty}\right] \\&
\leq 2B_{W V}^{(h)}B_{W Q K}^{(h)}\max _{t}\left\|\bm{x}_{t}^{i}\right\|_2\left\|\bm{x}_{cls}\right\|_2\mathbb{E}_{\bm{\epsilon}}\left[\left\|\sum_{i=1}^{n} \epsilon_{i} \bm{X}_{i}^{\top}\right\|_{2, \infty}\right].
    \end{aligned}
\end{equation}
According to the Assumption \ref{assumption3}, the result of Eq. (\ref{eq27})   can be further scaled to
\begin{equation}
\begin{aligned}
 \mathbb{E}_{\bm{\epsilon}}\left[\sup \left\|S_{h}\right\|_2\right]   &\leq 2B_{W V}^{(h)}B_{W Q K}^{(h)} R^2 \max _{t}\mathbb{E}_{\bm{\epsilon}}\left[\left\|\sum_{i=1}^{n} \epsilon_{i} \bm{X}_{i}^{\top}\right\|_{2, \infty}\right] \\&
 =2B_{W V}^{(h)}B_{W Q K}^{(h)} R^2 \max _{t}\mathbb{E}_{\bm{\epsilon}}\left\|\sum_{i=1}^{n} \epsilon_{i} \bm{x}_{t}^{i}\right\|_2.
 \end{aligned}
\end{equation}
By leveraging the Jensen’s Inequality and Lemma \ref{lemma2}, we obtain the following bound on $S_h$ for a single head:
\begin{equation}
\begin{aligned}
    \mathbb{E}_{\bm{\epsilon}}\left[\sup \left\|S_{h}\right\|_2\right]  & \leq 2B_{W V}^{(h)}B_{W Q K}^{(h)} R^2  \max _{t}\left[\mathbb{E}_{\bm{\epsilon}}\left\|\sum_{i=1}^{n} \epsilon_{i} \bm{x}_{t}^{i}\right\|^{2}_2\right]^{\frac{1}{2}}\\&
    \leq 2B_{W V}^{(h)}B_{W Q K}^{(h)} R^2  \max _{t}\left[\sum_{i=1}^{n}\left\|\bm{x}_{t}^{i}\right\|^{2}_2\right]^{\frac{1}{2}} \\&
    \leq  2B_{W V}^{(h)}B_{W Q K}^{(h)} R^3\sqrt{n}. 
    \end{aligned}
\end{equation}
Since the square root function is concave, we have 
\begin{equation}\label{eq30}
\mathbb{E}_{\varepsilon}\left(\sum_{h=1}^H\left(\sup \left\|S_{h}\right\|^2_2\right)\right)^{1 / 2} \leq\left(\mathbb{E}_{\varepsilon} \sum_{h=1}^H\left(\sup \left\|S_{h}\right\|^2_2\right)\right)^{1 / 2}=\left(\sum_{h=1}^H \mathbb{E}_{\varepsilon}\left(\sup \left\|S_{h}\right\|_2^2\right)\right)^{1 / 2}.
\end{equation}
Based  on Eqs. (\ref{eq24}) and (\ref{eq30}), we have the following bound on $\hat{\Re}_{D}(\mathcal{F})$:
\begin{equation}
\begin{aligned}
\hat{\Re}_{D}(\mathcal{F}) &\leq \frac{2 B_{w} B_{W C} B_{O} L_{\sigma}}{n} \left(\sum_{h=1}^{H}\mathbb{E}_{\varepsilon}\left(\sup \left\|S_{h}\right\|_2^2\right)\right)^{1 / 2}
\\ & \leq \frac{2 B_{w} B_{W C} B_{O} L_{\sigma}}{n} \left(\sum_{h=1}^{H}\left(2B_{W V}^{(h)}B_{W Q K}^{(h)} R^3\sqrt{n}\right)^2\right)^{1 / 2} \\&
\leq \frac{4 B_{w} B_{W C} B_{O} L_{\sigma}R^3}{\sqrt{n}}\left(\sum_{h=1}^{H}\left(B_{W V}^{(h)}B_{W Q K}^{(h)} \right)^2\right)^{1 /2}.
\end{aligned}
\end{equation}
Since gradient explosion does not occur during normal training, the norms of the model parameters are bounded and admit finite upper bounds, while not simultaneously collapsing to zero. Therefore, we denote $a_h=B_{W V}^{(h)}B_{W Q K}^{(h)}$, and the mean value of $a_h$  necessarily lies between the minimum value and the maximum value. Then, we can obtain the following equality:
\begin{equation}
\sum_{h=1}^{H} a_{h}^{2}=\sum_{h=1}^{H}\left(a_{h}-\bar{a}\right)^{2}+H \bar{a}^{2}, \quad \bar{a}=\frac{1}{H} \sum_{h=1}^{H} a_{h} .   
\end{equation}
Thus, the following result holds:
\begin{equation}\label{eq33}
    \left(\sum_{h=1}^{H} a_{h}^{2}\right)^{1 / 2} \geq \sqrt{H} \bar{a},
\end{equation}
where the condition for the equality to hold is that $a_{1}=\cdots=a_{H}=\bar{a}$. Therefore, we can get the final bound on $\hat{\Re}_{D}(\mathcal{F})$:
\begin{equation}\label{eq34}
\hat{\Re}_{D}(\mathcal{F}) \leq \frac{4 B_{w} B_{W C} B_{O} L_{\sigma}R^3}{\sqrt{n}}\sqrt{H \bar{a}^{2}+\sum_{h=1}^{H}\left(a_{h}-\bar{a}\right)^{2}}.
\end{equation}
Eq. (\ref{eq34}) shows that the upper bound of $\hat{\Re}_{D}(\mathcal{F})$ attains its minimum if and only if $a_{1}=\cdots=a_{H}=\bar{a}$, which corresponds to a high degree of alignment in the parameter scales of the query, key and value projection matrices across different heads. Consequently, increased heterogeneity in the attention distributions among heads leads to a looser complexity upper bound, which in turn is associated with weaker generalization performance and reduced accuracy in predicting the next  token.

\newpage
\section{Proof of The Theorem \ref{theorem2}} \label{sectionB}
We begin the proof of Theorem \ref{theorem2} by introducing the following lemmas:
\begin{lemma}\label{lemmaB1} \cite{kawaguchi2022robustness}
The vector $X=(X_{1}, \ldots, X_{k})$ is defined to 
 follow the multinomial distribution with parameters $p=\left(p_{1}, \ldots, p_{k}\right)$. Let $\bar{a}_{1}, \ldots, \bar{a}_{k} \geq 0$ be fixed such that $\sum_{i=1}^{k} \bar{a}_{i} p_{i} \neq 0$. Then, for any $\epsilon>0$, the following inequality holds:
\begin{equation*}
\mathbb{P}\left(\sum_{i=1}^{k} \bar{a}_{i}\left(p_{i}-\frac{X_{i}}{m}\right)>\epsilon\right) \leq \exp \left(-\frac{m \epsilon^{2}}{\beta}\right), 
\end{equation*}
where $\beta=2 \sum_{i=1}^{k} \bar{a}_{i}^{2} p_{i}$.
\end{lemma}
\begin{lemma}\label{lemmaB2}
For any $y \in \mathcal{Y}$, if the loss function $l(\cdot, y)$ is $L_{l}$-Lipschitz, the following inequality exists:
\begin{equation}
 |l(u, y)-l(v, y)| \leq L_{l}\|u-v\|_2, \quad \forall u, v \in \mathbb{R} .   
\end{equation}
\end{lemma}
\begin{lemma}\label{lemmaB3}
 If the function $\psi$ is $L_{l}$-Lipschitz, with respect to the Euclidean norm $\|\cdot\|_{2}$, the following inequality exists:
 \begin{equation}
 |\psi(a)-\psi(b)| \leq L_{\psi}\|a-b\|_{2}, \quad \forall a, b \in \mathbb{R}^{d_{v}}. 
 \end{equation}
\end{lemma}
We consider a Transformer model with multi-head self-attention and a single prediction target. Let $\bm{X}$ denote the input sequence and $\bm{Y}$ the target token to be predicted.  For each head $h \in\{1, \ldots, H\}$, we define a feature extraction mapping $\phi^{(h)}: \mathcal{X} \rightarrow \mathbb{R}^{d_{v}}$. The output representation of the 
$h$-th head is defined as $z^{(h)}(\bm{X})=\phi^{(h)}(\bm{X}) \in \mathbb{R}^{d_{v}}$. $z^{(h)}(\bm{X})$ summarizes information extracted from $\bm{X}$ through  the $h$-th attention mechanism. Since the Transformer fuses the multi-head representations by concatenation, the resulting concatenated representation is denoted by $z(\bm{X})=\operatorname{concat}\left(z^{(1)}(\bm{X}), \ldots, z^{(H)}(\bm{X})\right) \in \mathbb{R}^{H d_{v}}$. By defining the subsequent networks (a composition of linear layers and  activation functions) as $\psi: \mathbb{R}^{H d_{v}} \rightarrow \mathbb{R}$, the model’s scalar output is then given by $f(\bm{X})=\psi(z(\bm{X})) \in \mathbb{R}$.  Next, from the perspective of model training, we define the training dataset $D=\{(\bm{X}_{i}, \bm{Y}_{i}): i \in[n]\}$  drawn from the distribution over $\mathcal{X} \times \mathcal{Y}$, where $\mathcal{Y}$ denotes the training corpus of the model, consisting of $C$ words. The goal of learning is to find a hypothesis $f \in \mathcal{F}$ with good generalization performance by minimizing the loss $l$ on the dataset $D$. Then,
the expected risk and  empirical risk  w.r.t. the training dataset $D$ can be denoted as  $R({f})=\mathbb{E}_{(\bm{X}, \bm{Y}) \sim \mathcal{X} \times \mathcal{Y}}[l({f}(\bm{X}), \bm{Y})]$ and $\widehat{R}_{D}({f})=\frac{1}{N} \sum_{i=1}^{N}  l(f( \bm{X}_{i}), \bm{Y}_{i})$, respectively. The generalization error can be written as
\begin{equation} \label{eq39}
    \operatorname{Gen}(f)=R(f)-\hat{R}_{D}(f)=\mathbb{E}_{(\bm{X}, \bm{Y}) \sim \mathcal{X} \times \mathcal{Y}}[l({f}(\bm{X}), \bm{Y})]-\frac{1}{N} \sum_{i=1}^{N}  l(f( \bm{X}_{i}), \bm{Y}_{i}).
\end{equation}
Eq. (\ref{eq39}) can be expanded as 
\begin{equation}
\begin{aligned}
\operatorname{Gen}(f)&=\mathbb{E}_{(\bm{X}, \bm{Y}) \sim \mathcal{X} \times \mathcal{Y}}\left[l\left(\psi\left(\operatorname{concat}\left(z^{(1)}(\bm{X}), \ldots, z^{(H)}(\bm{X})\right)\right), \bm{Y}\right)\right]
\\&-\frac{1}{n} \sum^{n} l\left(\psi\left(\operatorname{concat}\left(z^{(1)}((\bm{X}_i)), \ldots, z^{(H)}\left(\bm{X}_{i}\right)\right)\right), \bm{Y}_{i}\right).
\end{aligned}
\end{equation}
According to Lemma \ref{lemmaB2} and in view of the Lipschitz continuity of $l(\cdot ; \bm{Y})$, it follows that
\begin{equation}\label{eq41}
l(\psi(z(\bm{X})), \bm{Y}) \leq l(\psi(0), \bm{Y})+L_{l}|\psi(z(\bm{X}))-\psi(0)|. 
\end{equation}
Since the activation function in $\psi$ is Lipschitz continuous, it follows that $\psi$ is also Lipschitz continuous. Moreover, applying Lemma \ref{lemmaB3} yields that 
\begin{equation}\label{eq42}
    |\psi(z(\bm{X}))-\psi(0)| \leq L_{\psi}\|z(\bm{X})\|_{2}. 
\end{equation}
Furthermore, noting that $z(\bm{X})=\operatorname{concat}\left(z^{(1)}(\bm{X}), \ldots, z^{(H)}(\bm{X})\right)$, we obtain the following norm relation induced by concatenation:
\begin{equation}\label{eq43}
\|z(\bm{X})\|_{2}=\left(\sum_{h=1}^{H}\left\|z^{(h)}(\bm{X})\right\|_{2}^{2}\right)^{1 / 2} \leq \sum_{h=1}^{H}\left\|z^{(h)}(\bm{X})\right\|_{2}. 
\end{equation}
Combining Eqs. (\ref{eq41}), (\ref{eq42}) and (\ref{eq43}), we can obtain the following result:
\begin{equation}
l(\psi(z(\bm{X})), \bm{Y}) \leq l(\psi(0), \bm{Y})+L_{l} L_{\psi} \sum_{h=1}^{H}\left\|z^{(h)}(\bm{X})\right\|_{2}.
\end{equation}
Then,  the head-wise surrogate loss is denoted as follows:
\begin{equation}\label{eq45}
\hat{l}\left({\psi} \circ \phi^{(h)}(\bm{X}), \bm{Y}\right)=\frac{1}{H} l(\psi(0), \bm{Y})+L_{l} L_{\psi}\left\|z^{(h)}(\bm{X})\right\|_{2}, \quad \phi^{(h)}(\bm{X})=z^{(h)}(\bm{X}) .   
\end{equation}
Besides,  the mean surrogate loss across multiple heads is given by
\begin{equation}\label{eq46}
\hat{l}_{\mathrm{avg}}((\bm{X}, \bm{Y}) ; {\psi}, \phi)=\frac{1}{H} \sum_{h=1}^{H} \hat{l}\left({\psi} \circ \phi^{(h)}(\bm{X}), \bm{Y}\right).    
\end{equation}
Based on Eqs. (\ref{eq45}) and (\ref{eq46}), we can get the following equality:
\begin{equation}
H \hat{l}_{\mathrm{avg}}((\bm{X}, \bm{Y}) ; {\psi}, \phi)=l(\psi(0), \bm{Y})+L_{l} L_{\psi} \sum_{h=1}^{H}\left\|z^{(h)}(\bm{X})\right\|_{2}.    
\end{equation}
Consequently, we can obtain the following relation:
\begin{equation}\label{eq48}
l(\psi(z(\bm{X})), \bm{Y}) \leq H \hat{l}_{\mathrm{avg}}((\bm{X}, \bm{Y}) ; {\psi}, \phi).
\end{equation}
From Eq. (\ref{eq48}), we can get the following bound of the expected risk $R({f})$:
\begin{equation}\label{eq49}
R(f)=\mathbb{E}_{(\bm{X}, \bm{Y}) \sim \mathcal{X} \times \mathcal{Y}}[l(\psi(z(\bm{X})), \bm{Y})] \leq H \cdot \mathbb{E}_{(\bm{X}, \bm{Y}) \sim \mathcal{X} \times \mathcal{Y}}\left[\hat{l}_{\mathrm{avg}}((\bm{X}, \bm{Y}) ; {\psi}, \phi)\right] .    
\end{equation}
Substituting the upper bound in Eq. (\ref{eq49}) back into (\ref{eq39}), we can obtain: 
\begin{equation}\label{eq50}
\operatorname{Gen}(f) \leq H \mathbb{E}_{(\bm{X}, \bm{Y}) \sim \mathcal{X} \times \mathcal{Y}}\left[\hat{l}_{\mathrm{avg}}((\bm{X}, \bm{Y}) ; {\psi}, \phi)\right] -\hat{R}_{D}(f).    
\end{equation}
Next, we add and subtract the same term $H \cdot \frac{1}{n} \sum_{i=1}^{n} \hat{l}_{\mathrm{avg}}(X_{i}, Y_{i})$ on the right-hand side of Eq. (\ref{eq50}), and obtain the following expression via algebraic manipulation:
\begin{equation}
\begin{aligned}
&H  \mathbb{E}_{(\bm{X}, \bm{Y}) \sim \mathcal{X} \times \mathcal{Y}}\left[\hat{l}_{\mathrm{avg}}((\bm{X}, \bm{Y}) ; {\psi}, \phi)\right] -\hat{R}_{D}(f)
\\&=H\left(\mathbb{E}_{(\bm{X}, \bm{Y}) \sim \mathcal{X} \times \mathcal{Y}}\left[\hat{l}_{\mathrm{avg}}((\bm{X}, \bm{Y}) ; {\psi}, \phi)\right]-\frac{1}{n} \sum_{i=1}^{n} \hat{l}_{\text {avg }}\left(\bm{X}_{i}, \bm{Y}_{i}\right)\right)+\underbrace{\left(H  \frac{1}{n} \sum_{i=1}^{n} \hat{l}_{\text {avg }}\left(\bm{X}_{i}, \bm{Y}_{i}\right)-\hat{R}_{D}(f)\right)}_{\Delta_{D}(f)}. 
\end{aligned}
\end{equation}
Based on the inequality (\ref{eq48}), we can obtain the  result $H\hat{l}_{\text {avg }}\left(\bm{X}_{i}, \bm{Y}_{i}\right) \geq l(f( \bm{X}_{i}), \bm{Y}_{i}) $ for each sample point. Therefore, the conclusion $\hat{R}_{D}(f) \geq 0$ holds. Then, the upper bound of 
$\operatorname{Gen}(f)$ can be reformulated as
\begin{equation}\label{eq52}
\begin{aligned}
    \operatorname{Gen}(f) &\leq H\left(\mathbb{E}_{(\bm{X}, \bm{Y}) \sim \mathcal{X} \times \mathcal{Y}}\left[\hat{l}_{\mathrm{avg}}((\bm{X}, \bm{Y}) ; {\psi}, \phi)\right]-\frac{1}{n} \sum_{i=1}^{n} \hat{l}_{\text {avg }}\left(\bm{X}_{i}, \bm{Y}_{i}\right)\right) \\&
    =H\left(\mathbb{E}_{(\bm{X}, \bm{Y}) \sim \mathcal{X} \times \mathcal{Y}}\left[\hat{l}_{\mathrm{avg}}((\bm{X}_i, \bm{Y}_i) ; {\psi}, \phi)\right]-\frac{1}{n} \sum_{i=1}^{n} \frac{1}{H} \sum_{h=1}^{H} \hat{l}\left({\psi} \circ \phi^{(h)}(\bm{X}_i), \bm{Y}_i\right)\right).
    \end{aligned}
\end{equation}
The representation extracted via the multi-head attention mechanism can be expressed as 
$\bm{S}=\phi(\bm{X})=(\bm{S}^{(1)}, \ldots, \bm{S}^{(H)})=(\phi^{(1)}(\bm{X}), \ldots, \phi^{(H)}(\bm{X}))=(z^{(1)}(\bm{X}), \ldots, z^{(H)}(\bm{X}))$. Let the row input data $\bm{X}$ be generated with a hidden category-specific function $\theta^{c}$ by $\bm{X}=\theta^{c}(\bm{y}^{c}, \bm{V})$, where $\bm{y}^{c} \in \mathcal{Y}$ denotes the $c$-th word in training corpus, and $\bm{V}=\{\bm{V}^{(h)}=(\bm{V}_{1}^{(h)}, \ldots, \bm{V}_{d}^{(h)})\}_{h=1}^{H} \in  \mathbb{R}^{H \times d}$ are nuisance variables.  The conditional random variables of $\bm{X}$ and $\bm{S}$ given the category $\bm{Y} = \bm{y}^c$ are denoted  as $\bm{X}_{\bm{y}^c}$ and $\bm{S}_{\bm{y}^c}$, respectively. For any $\bm{y}^c \in \mathcal{Y}$, we define the sensitivity $c_{\phi}^{\bm{y}^c}$ of the mapping $\phi=\{\phi^{(h)}\}_{h=1}^{H}$ with respect to the nuisance variable $\bm{V}_{i}^{(h)}$ is defined as 
 \begin{equation}\label{eq100}
    \begin{aligned}
     c_{\phi}^{\bm{y}^{c}}= & \sup _{h \in[H]} \sup _{\bm{v}_{1}^{(h)}, \ldots, {\bm{\hat{v}}}_{i}^{(h)}, \ldots, \bm{v}_{d}^{(h)}}|\log (p_{\bm{s}}(\phi^{(h)} \circ \theta_{{\bm{y}_{c}}}(\bm{v}_{1}^{(h)}, \ldots, \bm{v}_{i}^{(h)}, \ldots, \bm{v}_{d}^{(h)}))) \\ &
     -\log (p_{\bm{s}}(\phi^{(h)} \circ \theta_{\bm{y}_{c}}\left(\bm{v}_{1}^{(h)}, \ldots, \hat{\bm{v}}_{i}^{(h)}, \ldots, \bm{v}_{d}^{(h)}\right)))|,
    \end{aligned}
 \end{equation}
where $\theta_{\bm{y}^c}(\bm{v}^{(h)})=\theta(\bm{y}^c, \bm{v}^{(h)})$ and $p_{\bm{s}}(\bm{s})=\mathbb{P}(\bm{S}=\bm{s})$. Following Eq. (\ref{eq100}), the global sensitivity of $\phi$ is defined as the supremum of its class-wise sensitivities, i.e., $c_{\phi}=\sup _{c \in[C]} c_{\phi}^{\bm{y}^{c}}$. Let $\mathcal{S}^{x}_{c}=\{\theta_{\bm{y}^c}(\bm{v}), \phi \circ \theta_{\bm{y}^c}(\bm{v}): \bm{v} \in \mathcal{V}, \bm{y}^c \in \mathcal{Y}\}$ denote the complete set of row data and its corresponding head-wise representations. For any $\gamma>0$, we construct the following category-specific typical representation subset \cite{wen2025towards}:
\begin{equation}
\mathcal{S}_{c,\gamma}^{x}=\left\{ \bm{X}, \bm{S} \in \mathcal{S}_{c}^{x}:-\log p_{\bm{s}|\bm{y}^{c}}(\bm{s})-H(\bm{S}_{\bm{y}^{c}}) \leq c_{\phi} \sqrt{\frac{d \log (\sqrt{nh} / \gamma)}{2}}\right\}.
\end{equation}
Define the function $h^{\prime}_{\bm{y}^{c}}(\bm{v})=-\log p_{\bm{s} \mid \bm{y}^{c}}(h^{\prime\prime}_{\bm{y}^{c}}(\bm{v}))$, where $h^{\prime\prime}_{\bm{y}^{c}}(\bm{v})=\phi \circ(\theta_{\bm{y}^c}(\bm{v}))$.
  Let $p_{\bm{v}}(\bm{v})={P}(\bm{V}=\bm{v})$ and $h_{\bm{y}^{c}}^{-1}(\bm{s})=\left\{\bm{v} \in \mathcal{V}: h_{\bm{y}^{c}}(\bm{v})=\bm{s}\right\}$, we can get
\begin{equation*}
 \begin{aligned}
\mathbb{E}_{\bm{V}}\left[h^{\prime}_{\bm{y}^{c}}(\bm{V})\right] & =-\sum_{\bm{v} \in \mathcal{V}} p_{\bm{v}}(\bm{v}) \log p_{\bm{s} \mid \bm{y}^{c}} \left(h^{\prime\prime}_{\bm{y}^{c}}(\bm{v})\right) \\
& =-\sum_{\bm{s} \in \mathcal{S}_{c}^{x}} \sum_{\bm{v} \in h^{-1}_{\bm{y}^{c}}(\bm{s})} p_{\bm{v}}(\bm{v}) \log p_{\bm{s} \mid \bm{y}^{c}}\left(h^{\prime\prime}_{\bm{y}^{c}}(\bm{v})\right) \\
& =-\sum_{\bm{s} \in \mathcal{S}_{c}^{x}}\left(\sum_{v \in h^{-1}_{\bm{y}^{c}}(\bm{s})} p_{\bm{v}}(\bm{v})\right) \log p_{\bm{s} \mid \bm{y}^{c}}(\bm{s}) \\
& =-\sum_{\bm{s} \in \mathcal{R}_{c}^{x}} p_{\bm{s} \mid \bm{y}^{c}}(\bm{s}) \log p_{\bm{s} \mid \bm{y}^{c}}(\bm{s}) \\
& =H\left({S}_{\bm{y}^{c}}\right).
\end{aligned}   
\end{equation*}
Using McDiarmid’s inequality for the function $h^{\prime}_{\bm{y}^{c}}(\bm{V})$, we derive
\begin{equation*}
    \mathbb{P}\left(-\log p_{\bm{s}|\bm{y}^{c}}(\bm{s})-H(\bm{S}_{\bm{y}^{c}}) \geq \epsilon\right) \leq \exp \left(-\frac{2 \epsilon^{2}}{d(c_{\phi})^{2}}\right).
\end{equation*}
Taking $\delta=\exp \left(-\frac{2 \epsilon^{2}}{d(c_{\phi})^{2}}\right)$, we have $
\epsilon=c_{\phi} \sqrt{\frac{d \log (1 / \delta)}{2}}$. By setting $\delta=\gamma / \sqrt{n H}$, we can obtain $\mathbb{P}\left(\bm{X}, \bm{S} \notin \mathcal{S}_{c, \gamma}^{x}\right) \leq \delta=\frac{\gamma}{\sqrt{nH}}$. Since
$\exp (-H(\bm{S}_{\bm{y}^{c}})-\epsilon) \leq  p_{\bm{s}|\bm{y}^{c}}(\bm{s})$
, we can further derive the following transformation:
\begin{equation}\label{eq55}
\left|\mathcal{S}_{c,\gamma}^{x}\right| \exp (-H(\bm{S}_{\bm{Y}^{c}})-\epsilon)=\sum_{\bm{s} \in \mathcal{S}_{c,\gamma}^{x}} \exp (-H(\bm{S}_{\bm{y}^{c}})-\epsilon)\leq \sum_{r \in \mathcal{S}_{c,\gamma}^{x}} p_{\bm{s}}(\bm{s})=1.
\end{equation}
From Eq. (\ref{eq55}), we have the following result:
\begin{equation} \label{eq156}
\left|\mathcal{S}_{c,\gamma}^{x}\right| \leq \exp \left(H(\bm{S}_{\bm{y}^{c}})+c_{\phi} \sqrt{\frac{d \log (\sqrt{n H} / \gamma)}{2}}\right).
\end{equation}
We further define $T^{\bm{y}^{c}}=|\mathcal{S}_{c, \gamma}^{x}|$, where the typical subset $\mathcal{S}_{c,\gamma}^{x}$ is given by $\mathcal{S}_{c,\gamma}^{x}=\{(a_{c,1}^{x}, a_{c,1}^{s}), \ldots,\left(a_{c,T}^{x}, a_{c,T}^{s}\right)\}$. To facilitate the subsequent  derivations, we denote $\bm{x^{(h)}_i}$=$\bm{X}_i$. Based on this notation,  we introduce the  following category-specific sets:
\begin{equation}
    \begin{aligned}
\mathcal{U}^{\bm{y}^{c}} & =\left\{i \in[N], h \in[H]: \bm{x}_{i}^{(h)}, \phi^{(h)}(\bm{x}_{i}^{(h)}) \notin \mathcal{S}_{c, \gamma}^{x}, \bm{Y}_{i}=\bm{y}^{c}\right\}, \\
\mathcal{U}_{k}^{\bm{y}^{c}} & =\left\{i \in[N], h \in[H]: \bm{x}_{i}^{(h)}=a_{k}^{c,x}, \phi^{(h)}(\bm{x}_{i}^{(h)})=a_{k}^{c,s}, \bm{Y}_{i}=\bm{y}^{c}\right\}.
\end{aligned}
\end{equation}
Then, we conduct an analysis of the cumulative loss for individual heads:
\begin{equation}\label{eq57}
\begin{aligned}
&
    \frac{1}{nH} \sum_{i=1}^{n} \sum_{h=1}^{H} \hat{l}\left({\psi} \circ \phi^{(h)}(\bm{X}_i), \bm{Y}_i\right)
    \\&
    =\frac{1}{nH}  \sum_{\bm{y}^{c} \in \mathcal{Y}}\left(\sum_{i, h \in \mathcal{U}^{\bm{y}^{c}}} \hat{l}(\psi \circ \phi^{(h)}(\bm{x}_{i}^{(h)}), \bm{y}^{c})+\sum_{k=1}^{T^{\bm{y}^{c}}} \sum_{i, v \in \mathcal{U}_{k}^{\bm{y}^{c}}} \hat{l}(\psi \circ \phi^{(h)}(\bm{x}_{i}^{(h)}), \bm{y}^{c})\right)
    \\&
    = \frac{1}{nH} \sum_{\bm{y}^{c} \in \mathcal{Y}} \sum_{i, h \in \mathcal{U}^{\bm{y}^{c}}} \hat{l}(\psi \circ \phi^{(h)}(\bm{x}_{i}^{(h)}), \bm{y}^{c})+ \frac{1}{nH}\sum_{\bm{y}^{c} \in \mathcal{Y}} \sum_{k=1}^{T^{\bm{y}^{c}}}\sum_{i, v \in \mathcal{U}_{k}^{\bm{y}^{c}}} \hat{l}(\psi(a_{k}^{c,s}) , \bm{y}^{c}) \\&
=\frac{1}{nH} \sum_{\bm{y}^{c} \in \mathcal{Y}} \sum_{i, h \in \mathcal{U}^{\bm{y}^{c}}} \hat{l}(\psi \circ \phi^{(h)}(\bm{x}_{i}^{(h)}), \bm{y}^{c})+ \frac{\left|\mathcal{U}_{k}^{y^{c}}\right|}{nH}\sum_{\bm{y}^{c} \in \mathcal{Y}} \sum_{k=1}^{T^{\bm{y}^{c}}} \hat{l}(\psi(a_{k}^{c,s}) , \bm{y}^{c}).
    \end{aligned}
\end{equation}
Similarly, the population risk $\mathbb{E}_{(\bm{X}, \bm{Y}) \sim \mathcal{X} \times \mathcal{Y}}\left[\hat{l}_{\mathrm{avg}}((\bm{X}, \bm{Y}) ; {\psi}, \phi)\right]$ can be decomposed into the following components:
\begin{equation}\label{eq58}
\begin{aligned}
&
   \mathbb{E}_{(\bm{X}, \bm{Y}) \sim \mathcal{X} \times \mathcal{Y}}\left[\hat{l}_{\mathrm{avg}}((\bm{X}, \bm{Y}); {\psi}, \phi)\right]
   \\&
=\sum_{\bm{y}^{c} \in \mathcal{Y}}\mathbb{P}(\bm{Y}=y^{c})\mathbb{E}_{(\bm{X}, \bm{Y}^{c}) \sim \mathcal{X} \times \mathcal{Y}} [\hat{l}_{\mathrm{avg}}((\bm{X}, \bm{Y}) ; {\psi}, \phi)|\bm{Y}=y^{c}]
\\ &
= \sum_{\bm{y}^{c} \in \mathcal{Y}}\mathbb{P}(\bm{Y}=y^{c}, (\bm{X}, \bm{S}) \notin \mathcal{S}_{c, \gamma}^{x})\mathbb{E}_{(\bm{X}, \bm{Y}) \sim \mathcal{X} \times \mathcal{Y}} [\hat{l}_{\mathrm{avg}}((\bm{X}, \bm{Y}) ; {\psi}, \phi)|\bm{Y}=y^{c},(\bm{X}, \bm{S}) \notin \mathcal{S}_{c, \gamma}^{x}]
\\ &
+ \sum_{\bm{y}^{c} \in \mathcal{Y}}\mathbb{P}(\bm{Y}=y^{c}, (\bm{X}, \bm{S}) \in \mathcal{S}_{c, \gamma}^{x})\mathbb{E}_{(\bm{X}, \bm{Y}) \sim \mathcal{X} \times \mathcal{Y}} [\hat{l}_{\mathrm{avg}}((\bm{X}, \bm{Y}) ; {\psi}, \phi)|\bm{Y}=y^{c},(\bm{X}, \bm{S}) \in \mathcal{S}_{c, \gamma}^{x}]\\
&
=  \sum_{\bm{y}^{c} \in \mathcal{Y}}\mathbb{P}(\bm{Y}=y^{c}, (\bm{X}^{(j)}, \phi^{j}(\bm{X}^{(j)})) \notin \mathcal{S}_{c, \gamma}^{x})
\mathbb{E}_{(\bm{X}, \bm{Y}) \sim \mathcal{X} \times \mathcal{Y}, \bm{X}^{j} \sim \bm{X}}[\hat{l}(\psi \circ \phi^{j}(\bm{X}^{j}), y^{c})|\bm{Y}=y^{c},(\bm{X}^{(j)}, \\
&  \phi^{j}(\bm{X}^{(j)})) \notin \mathcal{S}_{c, \gamma}^{x}]+ \sum_{\bm{y}^{c} \in \mathcal{Y}}\sum_{k=1}^{T^{\bm{y}^{c}}}\mathbb{P}(\bm{Y}=y^{c}, \bm{X}^{(j)}=a_{k}^{c, x}, \phi^{j}(\bm{X}^{(j)})=a_{k}^{c,s})\hat{l}(\psi a_{k}^{c,s}, y^{c})).
   \end{aligned}
\end{equation}
Putting Eqs. (\ref{eq57}) and (\ref{eq58}) back into Eq. (\ref{eq52}), we can obtain 
\begin{equation} \label{eq205}
\begin{aligned}
& \mathbb{E}_{(\bm{X}, \bm{Y}) \sim \mathcal{X} \times \mathcal{Y}}\left[\hat{l}_{\mathrm{avg}}((\bm{X}_i, \bm{Y}_i) ; {\psi}, \phi)\right]-\frac{1}{n} \sum_{i=1}^{n} \frac{1}{H} \sum_{h=1}^{H} \hat{l}\left({\psi} \circ \phi^{(h)}(\bm{X}_i), \bm{Y}_i\right)
     \\ &
\leq  \sum_{\bm{y}^{c} \in \mathcal{Y}}\mathbb{P}(\bm{Y}=y^{c}, (\bm{X}^{(j)}, \phi^{j}(\bm{X}^{(j)})) \notin \mathcal{S}_{c, \gamma}^{x})
\mathbb{E}_{(\bm{X}, \bm{Y}^{c}) \sim \mathcal{X} \times \mathcal{Y}, \bm{X}^{j} \sim \bm{X}}[\hat{l}(\psi \circ \phi^{j}(\bm{X}^{j}), y^{c})|\bm{Y}=y^{c},(\bm{X}^{(j)}, \\
&  \phi^{j}(\bm{X}^{(j)})) \notin \mathcal{S}_{c, \gamma}^{x}] 
- \sum_{\bm{y}^{c} \in \mathcal{Y}}\mathbb{P}(\bm{Y}=y^{c}, (\bm{X}^{(j)}, \phi^{j}(\bm{X}^{(j)})) \notin \mathcal{S}_{c, \gamma}^{x}) \frac{1}{|\mathcal{U}^{\bm{y}^{c}}|}\sum_{i, h \in \mathcal{U}^{\bm{y}^{c}}}\hat{l}(\psi \circ \phi^{(h)}(\bm{x}_{i}^{(h)}), \bm{y}^{c})\\ &
+ \sum_{\bm{y}^{c} \in \mathcal{Y}}\mathbb{P}(\bm{Y}=y^{c}, (\bm{X}^{(j)}, \phi^{j}(\bm{X}^{(j)})) \notin \mathcal{S}_{c, \gamma}^{x}) \frac{1}{|\mathcal{U}^{\bm{y}^{c}}|}\sum_{i, h \in \mathcal{U}^{\bm{y}^{c}}}\hat{l}(\psi \circ \phi^{(h)}(\bm{x}_{i}^{(h)}), \bm{y}^{c}) \\ &
-\frac{1}{nH}   \sum_{\bm{y}^{c} \in \mathcal{Y}} \sum_{i, h \in \mathcal{U}^{\bm{y}^{c}}} \hat{l}(\psi \circ \phi^{(h)}(\bm{x}_{i}^{(h)}), \bm{y}^{c}) \\ &
+ \sum_{\bm{y}^{c} \in \mathcal{Y}}\sum_{k=1}^{T^{\bm{y}^{c}}}\mathbb{P}(\bm{Y}=y^{c}, \bm{X}^{(j)}=a_{k}^{c, x}, \phi^{j}(\bm{X}^{(j)})=a_{k}^{c,s})\hat{l}(\psi(a_{k}^{c,s}), y^{c}))\\ &
-   \sum_{\bm{y}^{c} \in \mathcal{Y}} \sum_{k=1}^{T^{\bm{y}^{c}}}\frac{\left|\mathcal{U}_{k}^{y^{c}}\right|}{nH}\hat{l}(\psi(a_{k}^{c,s}), \bm{y}^{c})
\end{aligned}
\end{equation}
Let $M_{x, y}=\ \sup _{\left(\bm{X}, \bm{Y}\right) \in \mathcal{X} \times \mathcal{Y}} \sum_{h=1}^{H} \hat{l}\left(f\left(\bm{X}^{(h)}\right), \bm{Y}\right)$ represent the maximum attainable loss and  $M_{x, y}^{s}= \sup _{i \in[n]}  \sum_{h=1}^{H} \hat{l}\left(f(\bm{X}_{i}^{(h)}), \bm{Y}_{i}\right)$  denote the
maximum instance-level loss.
The expression in Eq. (\ref{eq205}) can be partitioned into three  distinct terms. The upper bound corresponding to the first  term is derived as follows:
\begin{equation}
\begin{aligned}
&
\mathcal{P}_1= \sum_{\bm{y}^{c} \in \mathcal{Y}}\mathbb{P}(\bm{Y}=y^{c}, (\bm{X}^{(j)}, \phi^{j}(\bm{X}^{(j)})) \notin \mathcal{S}_{c, \gamma}^{x})(\mathbb{E}_{(\bm{X}, \bm{Y}^{c}) \sim \mathcal{X} \times \mathcal{Y}, \bm{X}^{j} \sim \bm{X}}[\hat{l}(\psi \circ \phi^{j}(\bm{X}^{j}), y^{c})|\bm{Y}=y^{c},  \\ & (\bm{X}^{(j)},   \phi^{j}(\bm{X}^{(j)})) \notin \mathcal{S}_{c, \gamma}^{x}]
-\frac{1}{|\mathcal{U}^{\bm{y}^{c}}|}\sum_{i, h \in \mathcal{U}^{\bm{y}^{c}}}\hat{l}(\psi \circ \phi^{(h)}(\bm{x}_{i}^{(h)}), \bm{y}^{c})
) \\ &
\leq  \sum_{\bm{y}^{c} \in \mathcal{Y}}\mathbb{P}(\bm{Y}=y^{c}, (\bm{X}, \bm{S}) \notin \mathcal{S}_{c, \gamma}^{x}) \mathbb{E}_{(\bm{X}, \bm{Y}^{c}) \sim \mathcal{X} \times \mathcal{Y}, \bm{X}^{j} \sim \bm{X}}[\sum_{h=1}^{H} \hat{l}(f(\bm{X}^{(h)}, \bm{Y}|\bm{Y}=y^{c},(\bm{X}, \bm{S}) \notin \mathcal{S}_{c, \gamma}^{x}] \\ &
\leq  \sum_{\bm{y}^{c} \in \mathcal{Y}}\mathbb{P}(\bm{Y}=y^{c})  \frac{\gamma}{\sqrt{nH}} \mathcal{M}_{x,y}=\frac{ \gamma}{\sqrt{nH}} \mathcal{M}_{x,y}.
\end{aligned}
\end{equation}
Define $p_k^{c}=\mathbb{P}(\bm{Y}=y^{c}, \bm{X}^{(j)}=a_{k}^{c, x}, \phi^{j}(\bm{X}^{(j)})=a_{k}^{c,s})$ for $k \in [T^{\bm{y}^{c}}]$ and $p_{T^{\bm{y}^{c}}+1}^c=\mathbb{P}(\bm{Y}=y^{c}, (\bm{X}^{(j)}, \phi^{j}(\bm{X}^{(j)})) \notin \mathcal{S}_{c, \gamma}^{x})$. Moreover, the associated losses  are  denoted as $b_k^{c}=\hat{l}(\psi(a_{k}^{c,s}), y^{c})$, where $k \in [T^{\bm{y}^{c}}+1]$. Then, we can get the following term:
\begin{equation}
\mathcal{P}_{3,k}^{c}=\sum_{t=1}^{T^{\bm{y}^{c}}}(p_{t}^{c}-\frac{|\mathcal{U}_{t}^{\bm{y}^{c}}|}{nH}) b_{t}^{c}-(p_{k}^{c}-\frac{|\mathcal{U}_{k}^{\bm{y}^{c}}|}{nH}) b_{k}^{c}.
\end{equation}
By applying Lemma \ref{lemma2}, we have the following result for any $\epsilon>0$ and $k \in\left[T^{\bm{y}^{c}}\right]$: 
\begin{equation} \label{eq210}
\mathbb{P}\left(\mathcal{P}_{3,k}^{c} \geq \epsilon\right) \leq \exp \left(-\frac{nH \epsilon^{2}}{2\left(\sum_{t=1}^{T^{\bm{y}^{c}}} p_{t}^{c}\left(b_{t}^{c}\right)^{2}-p_{k}^{c}\left(b_{k}^{c}\right)^{2}\right)}\right).   
\end{equation}
Besides, we can get the following inequality regarding $k=T^{\bm{y}^{c}}+1$ : 
\begin{equation}\label{eq211}
\mathbb{P}\left(p_{T^{\bm{y}^{c}}+1}^{c}-\frac{\left|\mathcal{U}^{\bm{y}^{c}}\right|}{nH} \geq \epsilon\right) \leq \exp \left(-\frac{nH \epsilon^{2}}{2 p_{T^{\bm{y}^{c}}+1}^{c}}\right).
\end{equation}
Upon assigning $\delta$ to the right-hand sides of Eqs. (\ref{eq210}) and (\ref{eq211}), respectively, the following variants emerge:
\begin{equation}
\left\{
\begin{array}{c}
\mathbb{P}\left(\mathcal{P}_{3,k}^{c} \geq \sqrt{\sum_{t=1}^{T^{\bm{y}^{c}}} p_{t}^{c}\left(b_{t}^{c}\right)^{2}-p_{k}^{c}\left(b_{k}^{c}\right)^{2}} \sqrt{\frac{2 \log (1 / \delta)}{nH}}\right) \leq \delta, \, \forall k \in [T^{\bm{y}^{c}}] \\\\
\mathbb{P}\left(p_{T^{\bm{y}^{c}}+1}^{c}-\frac{\left|\mathcal{U}^{\bm{y}^{c}}\right|}{nH} \geq \sqrt{\frac{2 p^{c}_{T^{\bm{y}^{c}}+1} \log (1 / \delta)}{nH}}\right) \leq \delta.
\end{array}\right.
\end{equation}
By taking union bounds over all $\bm{y}^{c} \in \mathcal{Y}$ $(|\mathcal{Y}|=C)$ and $k \in [T^{\bm{y}^{c}}]$, we can get
\begin{equation}
\left\{
\begin{array}{c}
\mathbb{P}\left(\mathcal{P}_{3,k}^{c} \geq \sqrt{\sum_{t=1}^{T^{\bm{y}^{c}}} p_{t}^{c}\left(b_{t}^{c}\right)^{2}-p_{k}^{c}\left(b_{k}^{c}\right)^{2}} \sqrt{\frac{2 \log (CT^{\bm{y}^{c}} / \delta)}{nH}}\right) \leq \delta, \, \forall k \in [T^{\bm{y}^{c}}] \\\\
\mathbb{P}\left(p_{T^{\bm{y}^{c}}+1}^{c}-\frac{\left|\mathcal{U}^{\bm{y}^{c}}\right|}{nH} \geq \sqrt{\frac{2 p_{T^{\bm{y}^{c}}+1}^{c} \log (C / \delta)}{nH}}\right) \leq \delta.
\end{array}\right.
\end{equation}
Then, for any $\delta>0$, with probability at least $1-\delta$, the second term in Eq. (\ref{eq205}) admits the following scaling, as a consequence of Jensen’s inequality:
\begin{equation}
\begin{aligned}
&
\mathcal{P}_2=\frac{1}{|\mathcal{U}^{\bm{y}^{c}}|} \sum_{y^{c}
\in \mathcal{Y}}\left(\mathbb{P}\left(\bm{Y}=y^{c}, (\bm{X}^{(j)}, \phi^{j}(\bm{X}^{(j)}))\notin \mathcal{S}_{c, \gamma}^{x}\right) -\frac{|\mathcal{U}^{\bm{y}^{c}}|}{       nH}\right)\sum_{i, h \in \mathcal{U}^{\bm{y}^{c}}}\hat{l}(\psi \circ \phi^{(h)}(\bm{x}_{i}^{(h)}), \bm{y}^{c}) \\ &
\leq  \sum_{y^{c}
\in \mathcal{Y}} \sqrt{\mathbb{P}(\bm{Y}=y^{c}, (\bm{X}^{(j)}, \phi^{j}(\bm{X}^{(j)})) \notin \mathcal{S}_{c, \gamma}^{x})} \frac{\sum_{i, j \in \mathcal{U}^{\bm{y}^{c}}} \hat{l}(\psi \circ \phi^{(h)}(\bm{x}_{i}^{(h)}),  \bm{y}^{c})}{\left|\mathcal{U}^{\bm{y}^{c}}\right|} \sqrt{\frac{2 \log (C/ \delta)}{nH}} \\ &
=  \sum_{y^{c}
\in \mathcal{Y}} \sqrt{\mathbb{P}(\bm{Y}=y^{c})} \sqrt{\mathbb{P}(\bm{X}^{(j)}, \phi^{j}(\bm{X}^{(j)})) \notin \mathcal{S}_{c, \gamma}^{x})}\frac{\sum_{i, j \in \mathcal{U}^{\bm{y}^{c}}} \hat{l}(\psi \circ \phi^{(h)}(\bm{x}_{i}^{(h)}),  \bm{y}^{c})}{\left|\mathcal{U}^{\bm{y}^{c}}\right|} \sqrt{\frac{2 \log (C/ \delta)}{nH}} \\ &
\leq  \frac{\sqrt{\gamma}}{(nH)^{1 / 4}}\frac{\sum_{i, j \in \mathcal{U}^{\bm{y}^{c}}} \hat{l}(\psi \circ \phi^{(h)}(\bm{x}_{i}^{(h)}),  \bm{y}^{c})}{\left|\mathcal{U}^{\bm{y}^{c}}\right|}\sqrt{|\mathcal{Y}|}\sqrt{\sum_{y^{c}
\in \mathcal{Y}}\mathbb{P}(\bm{Y}=y^{c})}\sqrt{\frac{2 \log (C/ \delta)}{nH}}\\ &
\leq {M}_{x,y}^{s} \frac{\sqrt{\gamma}}{(nH)^{1 / 4}} \sqrt{\frac{2C\log (C/ \delta)}{nH}}.
\end{aligned}
\end{equation}
Similarly, with probability at least $1-\delta$, we have the following comparison for $\mathcal{P}_{3,k}^{c}$:

\begin{equation}
\begin{aligned}
\mathcal{P}_{3,k}^{c} &\leq
\sqrt{\sum_{t=1}^{T^{\bm{y}^{c}}}p_{t}^{y}\left(b_{t}^{y}\right)^{2}-p_{k}^{c}\left(b_{k}^{y}\right)^{2}} \sqrt{\frac{2 \log (CT^{\bm{y}^{c}} / \delta)}{nH}} \\ &
\leq M_{x,y} \sqrt{\sum_{t=1}^{T^{\bm{y}^{c}}}p_{t}^{y}-p_{k}^{c}} \sqrt{\frac{2 \log (CT^{\bm{y}^{c}} / \delta)}{nH}} \\ &
= M_{x, y} \sqrt{\mathbb{P}\left(\bm{Y}^{c}=\bm{y}^{c} \bigcap(\bm{X}, \bm{S}) \in \mathcal{S}_{c, \gamma}^{x} \bigcap\left(\bm{X}^{(j)}, \phi^{(j)}\left(\bm{X}^{(j)}\right)\right) \neq\left(a_{k}^{c, x}, a_{k}^{c, s}\right)\right)} \sqrt{\frac{2 \log (CT^{\bm{y}^{c}} / \delta)}{nH}} \\ &
\leq M_{x, y}\sqrt{\mathbb{P}(\bm{Y}^{c}=\bm{y}^{c} )}\sqrt{\frac{2 \log (CT^{\bm{y}^{c}} / \delta)}{nH}}.  
\end{aligned}
\end{equation}
Next, we can derive the scaling result for the third term:
\begin{equation}\label{eq119}
\begin{aligned}
\mathcal{P}_{3}&= \sum_{\bm{y}^{c} \in \mathcal{Y}}\sum_{k=1}^{T^{\bm{y}^{c}}}\left(\mathbb{P}(\bm{Y}=y^{c}, \bm{X}^{(j)}=a_{k}^{c, x}, \phi^{j}(\bm{X}^{(j)})=a_{k}^{c,s})-\frac{|\mathcal{U}_{k}^{\bm{y}^{c}}|}{N C}\right)\hat{l}(\psi(a_{k}^{c,s}), \bm{y}^{c})\\ &
= \sum_{\bm{y}^{c} \in \mathcal{Y}} \frac{1}{T^{\bm{y}^{c}}-1} \sum_{k=1}^{T^{\bm{y}^{c}}} \mathcal{P}_{3,k}^{c} \\ &
\leq  \sum_{\bm{y}^{c} \in \mathcal{Y}} \frac{T^{\bm{y}^{c}}}{T^{\bm{y}^{c}}-1} M_{x, y}\sqrt{\mathbb{P}(\bm{Y}^{c}=\bm{y}^{c} )}\sqrt{\frac{2 \log (CT^{\bm{y}^{c}} / \delta)}{nH}}\\ &
\leq
2 \sum_{\bm{y}^{c} \in \mathcal{Y}}M_{x, y}\sqrt{\mathbb{P}(\bm{Y}^{c}=\bm{y}^{c} )}\sqrt{\frac{2 \log (CT^{\bm{y}^{c}} / \delta)}{nH}}.
\end{aligned}
\end{equation}
Based on Eq. (\ref{eq156}), the  property  of  $T^{\bm{y}^{c}}$ can be expressed as
\begin{equation}\label{eq220}
T^{\bm{y}^{c}}=|\mathcal{S}_{c,\gamma}^{x}|  \leq \exp \left(H(\bm{S}_{\bm{Y}^{c}})+c_{\phi} \sqrt{\frac{d \log (\sqrt{nH} / \gamma)}{2}}\right).
\end{equation}
Combining Eqs. (\ref{eq119}) and (\ref{eq220}), we can get the following result by applying Jensen’s inequality:
\begin{equation}\label{eq230}
\begin{aligned}
\mathcal{P}_{3} &\leq 2  \mathcal{M}_{x, y} \sum_{\bm{y}^{c} \in \mathcal{Y}} \sqrt{\mathbb{P}(\bm{Y}=\bm{y}^{c})} \sqrt{\frac{2\left(H(\bm{S}_{\bm{Y}^{c}}) +c_{\phi} \sqrt{\frac{d \log (\sqrt{nH} / \gamma)}{2}}\right)+2 \log (C / \delta)}{nH}} \\&
\leq 2 \mathcal{M}_{x, y}\sqrt{|\mathcal{Y}|}\sqrt{\sum_{y^{c}
\in \mathcal{Y}}\mathbb{P}(\bm{Y}=y^{c})} \sqrt{\frac{2\left(H(\bm{S}_{\bm{Y}^{c}}) +c_{\phi} \sqrt{\frac{d \log (\sqrt{nH} / \gamma)}{2}}\right)+2 \log (C / \delta)}{nH}} \\ &
= 2\sqrt{2C}  \mathcal{M}_{x, y}\sqrt{\frac{H(\bm{S}_{\bm{Y}^{c}}) +c_{\phi} \sqrt{\frac{d \log (\sqrt{nH} / \gamma)}{2}}+ \log (C/ \delta)}{nH}} \\ &
\leq 2\sqrt{2C} \mathcal{M}_{x, y}\sqrt{ \frac{H(\bm{S}|{\bm{Y}^{c}}) +c_{\phi} \sqrt{\frac{d \log (\sqrt{nH} / \gamma)}{2}}+ \log (C/ \delta)}{nH}}.
\\ &
\leq 
2\sqrt{2C} \mathcal{M}_{x, y}\sqrt{ \frac{H(\bm{S})-I\left(\bm{S} ; \bm{Y}\right) +c_{\phi} \sqrt{\frac{d \log (\sqrt{nH} / \gamma)}{2}}+ \log (C/ \delta)}{nH}}.
\end{aligned}
\end{equation} 
Therefore, with probability at least $1-\delta$, the generalization error  bound of our model is
\begin{equation} \label{eq172}
\begin{aligned}
&
    \operatorname{Gen}(f)=R(f)-\hat{R}_{D}(f)=\mathbb{E}_{(\bm{X}, \bm{Y}) \sim \mathcal{X} \times \mathcal{Y}}[l({f}(\bm{X}), \bm{Y})]-\frac{1}{N} \sum_{i=1}^{N}  l(f( \bm{X}_{i}), \bm{Y}_{i})
    \\ &
    \leq 
    =H\left(\mathbb{E}_{(\bm{X}, \bm{Y}) \sim \mathcal{X} \times \mathcal{Y}}\left[\hat{l}_{\mathrm{avg}}((\bm{X}_i, \bm{Y}_i) ; {\psi}, \phi)\right]-\frac{1}{n} \sum_{i=1}^{n} \frac{1}{H} \sum_{h=1}^{H} \hat{l}\left({\psi} \circ \phi^{(h)}(\bm{X}_i), \bm{Y}_i\right)\right)
    \\ &
\leqq \frac{H \gamma}{\sqrt{nH}} \mathcal{M}_{x,y}+{M}_{x,y}^{s} \frac{H\sqrt{\gamma}}{(nH)^{1 / 4}} \sqrt{\frac{2C\log (C/ \delta)}{nH}}
\\ &
+2\sqrt{2C} H\mathcal{M}_{x, y}\sqrt{ \frac{H(\bm{S})-I\left(\bm{S} ; \bm{Y}\right) +c_{\phi} \sqrt{\frac{d \log (\sqrt{nH} / \gamma)}{2}}+ \log (C/ \delta)}{nH}}\\ &
= \frac{\widetilde{\mathcal{K}}_{1}}{n^{1/2}H^{-1/2}}+ \frac{\widetilde{\mathcal{K}}_{2}}{n^{3/4}H^{-1/4}}+\widetilde{\mathcal{K}}_{3}\sqrt{\frac{ \left(-\sum_{h=1}^{H}I(\bm{S}^{(h)}; \bm{Y})+\widetilde{\mathcal{K}}_{4}\right)}{nH}},
    \end{aligned}
\end{equation}
where 
\begin{equation}
\begin{aligned}
&
\widetilde{\mathcal{K}}_{1} =H\gamma\mathcal{M}_{x,y}
\\ &
\widetilde{\mathcal{K}}_{2} ={M}_{x,y}^{s}\sqrt{2\gamma C\log (C/ \delta)}
\\ &
\widetilde{\mathcal{K}}_{3}=2\sqrt{2C} H\mathcal{M}_{x, y}
\\ &
\widetilde{\mathcal{K}}_{4}=H(\bm{S}) +c_{\phi} \sqrt{\frac{d \log (\sqrt{nH} / \gamma)}{2}}+ \log (C/ \delta).
\end{aligned}
\end{equation}
Based on the term $-\sum_{h=1}^{H}I(\bm{S}^{(h)}$ in Eq. (\ref{eq172}), it follows that constraining the mutual information between each representation extracted by the multi-head attention mechanism and the target $\bm{Y}$ leads to lower generalization error. Under this condition, the model attains greater precision in retrieving the correct lexical roots from the vocabulary. In other words, greater consistency in the image information attended to by different heads leads to stronger overall classification generalization and facilitates the selection of correct descriptive terms, thereby suppressing hallucination.
\section{Details of benchmarks}
\label{benchmarks}
To supplement the main text, we provide the evaluation details for the two hallucination benchmarks used in our captioning study. We follow the standard setup and evaluate on 500 images randomly sampled from the MSCOCO 2014 validation split, which provides ground-truth annotations for 80 object categories. For CHAIR~\cite{rohrbach2018object}, we obtain captions by prompting the MLLM with ``Please describe this image in detail'' and compute both sentence-level and object-level hallucination rates as 
$$
\scriptsize
C_I = \frac{|\{\text{hallucinated objects}\}|}{|\{\text{all mentioned objects}\}|},
C_S = \frac{|\{\text{captions with hallucinated objects}\}|}{|\{\text{all captions}\}|}
$$

For POPE~\cite{li2023evaluating}, we probe object existence using binary VQA queries of the form ``Is [object] in this image?'' and append ``Please answer yes or no.'' to enforce binary outputs. Following the protocol, the queried objects are constructed under three sampling strategies (random, popular, and adversarial), and we report results across all sampling options.

\section{Additional Results}
\subsection{Functionality of Dynamic Historical Attention Enhancement}  \label{Attention Enhancement}
\begin{figure}[htbp]
    \centering
    \begin{subfigure}[b]{0.35\textwidth}
        \centering
\includegraphics[width=\linewidth]{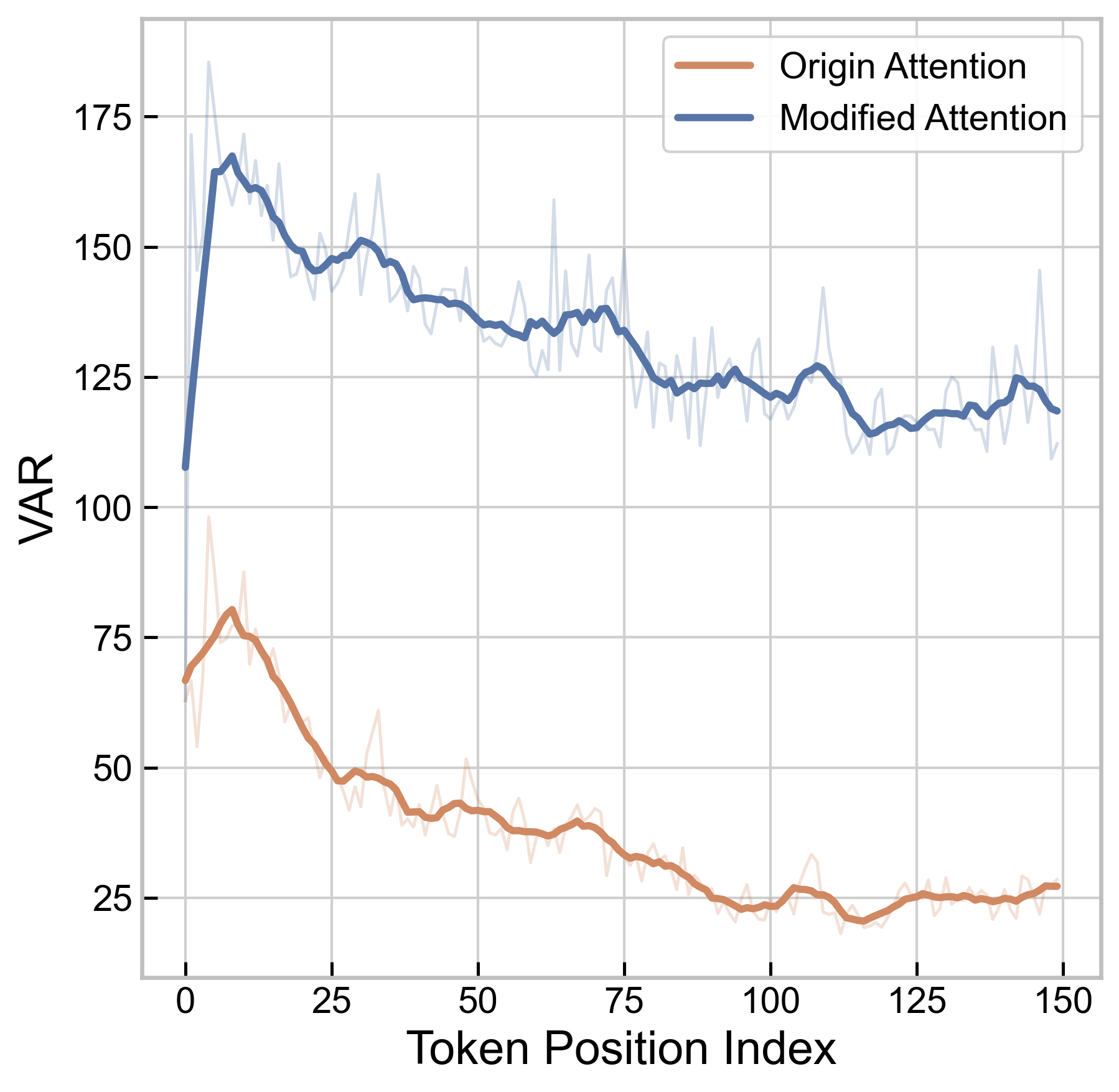}
        \label{fig:var1}
    \end{subfigure}
    \hspace{0.03\textwidth}
    \begin{subfigure}[b]{0.35\textwidth}
        \centering        \includegraphics[width=\linewidth]{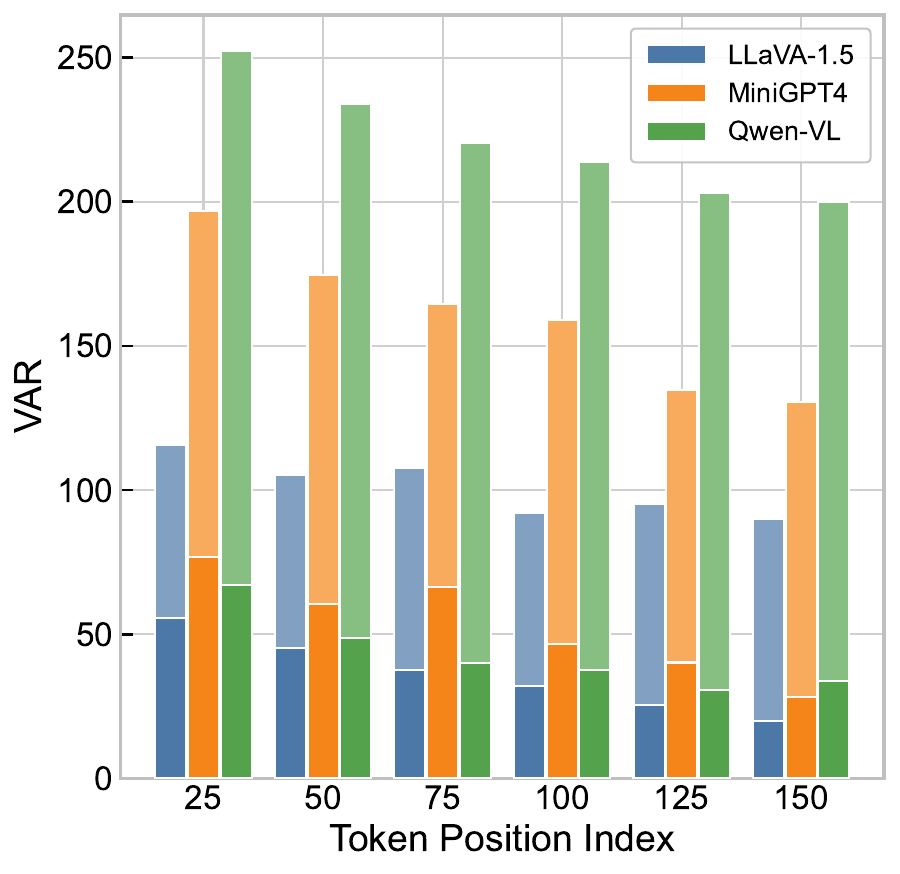}
        \label{fig:var2}
    \end{subfigure} \caption{Visualization of visual attention retention with dynamic historical attention enhancement.}
    \label{var-compare}
\end{figure}

Fig. ~\ref{var-compare} illustrates the impact of dynamic historical attention enhancement on visual attention retention during long-form generation. In the left panel, we plot VAR on LLaVA-1.5-7B as a function of token position and compare the standard forward propagation with the enhanced variant. Throughout the decoding process, the enhancement mechanism consistently yields higher VAR values, indicating that the model maintains a greater proportion of attention on visual tokens. 

On the right, we report VAR across multiple models (LLaVA-1.5-13B, MiniGPT-4, and Qwen-VL) under the same experimental setting. Solid bars represent the original attention mechanism, while the lighter segments denote the VAR gains introduced by the enhancement strategy. The results demonstrate that the proposed method leads to consistent improvements across diverse MLLMs and model scales.

\newpage

\subsection{Ablation Studies}
The ablation experiments are conducted to deeply investigate the effect of the three crucial modules of AFIP, i.e.,a dynamic gating mechanism driven by the discrepancy between attention allocated to visual and textual content ($S_{1}$), the integration of historical visual attention information ($S_{2}$), and a head-level attention distraction correction strategy ($S_{3}$).  Based on the ablation results reported in Table \ref{tab:ablation1}, we observe that removing any individual component consistently leads to performance degradation, thereby validating the effectiveness and careful design of the proposed AFIP. Moreover, the attention distraction correction strategy plays a pivotal role in boosting performance, as it  mitigates inconsistencies in attention distributions across multiple heads and reinforces the model’s ability to focus on the  salient visual content. In addition, the dynamic gating mechanism and the integration of historical visual attention further enhance performance by strengthening the model’s visual perception capability while preventing excessive attention intervention.
\begin{table}[H]
\centering
\caption{Ablation study on components across different MLLMs.}
\label{tab:ablation1}
\resizebox{0.8\linewidth}{!}{
\begin{tabular}{ccc|cc|cc|cc|cc|cc}
\toprule
\multirow{2}{*}{S1} & \multirow{2}{*}{S2} & \multirow{2}{*}{S3}
& \multicolumn{2}{c|}{\textbf{LLaVA-7b}}
& \multicolumn{2}{c|}{\textbf{Shikra}}
& \multicolumn{2}{c|}{\textbf{Qwen-VL}}
& \multicolumn{2}{c|}{\textbf{LLaVA-13b}}
& \multicolumn{2}{c}{\textbf{MiniGPT-4}} \\
\cmidrule(lr){4-5} \cmidrule(lr){6-7} \cmidrule(lr){8-9} \cmidrule(lr){10-11} \cmidrule(lr){12-13}
& & 
& $C_S$ & $C_I$
& $C_S$ & $C_I$
& $C_S$ & $C_I$
& $C_S$ & $C_I$
& $C_S$ & $C_I$ \\
\midrule
\ding{55} & \ding{51} & \ding{51}
& 22.6 & 8.8  & 27.4 & 11.4 & 22.8 & 7.6  & 26.4 & 11.6 & 27.3 & 9.5 \\
\ding{51} & \ding{55} & \ding{51}
& 19.3 & 7.2  & 22.6 & 9.3  & 19.1 & 6.9  & 22.5 & 10.2 & 21.6 & 8.4 \\
\ding{51} & \ding{51} & \ding{55}
& 49.6 & 13.2 & 46.8 & 14.3 & 25.8 & 8.2  & 45.3 & 13.8 & 32.4 & 10.2 \\
\rowcolor{bggray}
\ding{51} & \ding{51} & \ding{51}
& \textbf{16.8} & \textbf{4.4}
& \textbf{18.0} & \textbf{6.6}
& \textbf{16.6} & \textbf{5.0}
& \textbf{18.4} & \textbf{6.4}
& \textbf{17.8} & \textbf{7.3} \\
\midrule
\multicolumn{3}{c|}{\textbf{\texttt{Beam}}}
& 56.2 & 15.1 & 59.2 & 16.2 & 30.0 & 10.7 & 50.4 & 15.2 & 39.2 & 12.2 \\
\multicolumn{3}{c|}{\textbf{\texttt{Greedy}}}
& 55.4 & 14.4 & 62.0 & 17.5 & 28.2 & 8.9  & 49.7 & 14.7 & 39.4 & 11.0 \\
\bottomrule
\end{tabular}
}
\end{table}

\subsection{Layer Selection for Our Method}
\label{layer-select}
We apply our method to multiple candidate layer ranges $(0–32, 5–18, 18–32, 0–5 and 5–32)$ across five models under a unified experimental setting, and report the corresponding CHAIR results in Table~\ref{tab:layer-ranges}. From the results, we observe that the 5–32 configuration (Ours) consistently achieves the lowest $\mathcal{C}_S$ and $\mathcal{C}_I$, and demonstrates
a substantial improvement over the Greedy selection strategy. Notably, intervening in the early layers (0–5) clearly degrades performance, which is consistent with our analysis. Specifically, the early layers of MLLMs primarily perform modality aggregation and alignment, as discussed in the main text. These ablation results further validate the  practicality of our layer-selection strategy, showing that it can reliably identify the layers that require attention modulation and effectively reduce hallucinations.
\begin{table}[H]
\centering
\setlength{\tabcolsep}{9pt}
\renewcommand{\arraystretch}{1.05}
\caption{CHAIR hallucination evaluation across different layer ranges on various models.}
\resizebox{0.65\linewidth}{!}{
\begin{tabular}{l| l |ccccc|c}
\toprule
\textbf{Models} & \textbf{Layers} 
& \textbf{0--32} & \textbf{5--18} & \textbf{18--32} & \textbf{0--5} & \textbf{Greedy} & \textbf{5--32 (Ours)} \\
\midrule

\multirow{2}{*}{LLaVA-7B}
& $\mathcal{C}_S \downarrow$ & 28.9 & 23.7 & 36.7 & 53.2 & 55.4 & 16.8 \\
& $\mathcal{C}_I \downarrow$ & 8.8  & 7.2  & 10.9 & 13.7 & 14.4 & 4.4 \\
\midrule

\multirow{2}{*}{Shikra}
& $\mathcal{C}_S \downarrow$ & 26.1 & 21.5 & 42.8 & 58.6 & 62.0 & 18.0 \\
& $\mathcal{C}_I \downarrow$ & 10.9 & 8.9  & 13.7 & 15.9 & 17.5 & 6.6 \\
\midrule

\multirow{2}{*}{MiniGPT}
& $\mathcal{C}_S \downarrow$ & 25.3 & 20.5 & 29.4 & 37.5 & 39.4 & 17.8 \\
& $\mathcal{C}_I \downarrow$ & 8.5  & 7.9  & 9.9  & 10.6 & 11.0 & 7.3 \\
\midrule

\multirow{2}{*}{Qwen}
& $\mathcal{C}_S \downarrow$ & 21.8 & 19.5 & 23.6 & 26.3 & 28.2 & 16.6 \\
& $\mathcal{C}_I \downarrow$ & 6.9  & 6.3  & 7.7  & 8.4  & 8.9  & 5.0 \\
\bottomrule
\end{tabular}
}
\label{tab:layer-ranges}
\end{table}

\newpage
\subsection{Case study of Temporal Fading of Visual Attention}
\begin{figure}[H]
    \centering
    \includegraphics[width=0.85\linewidth]{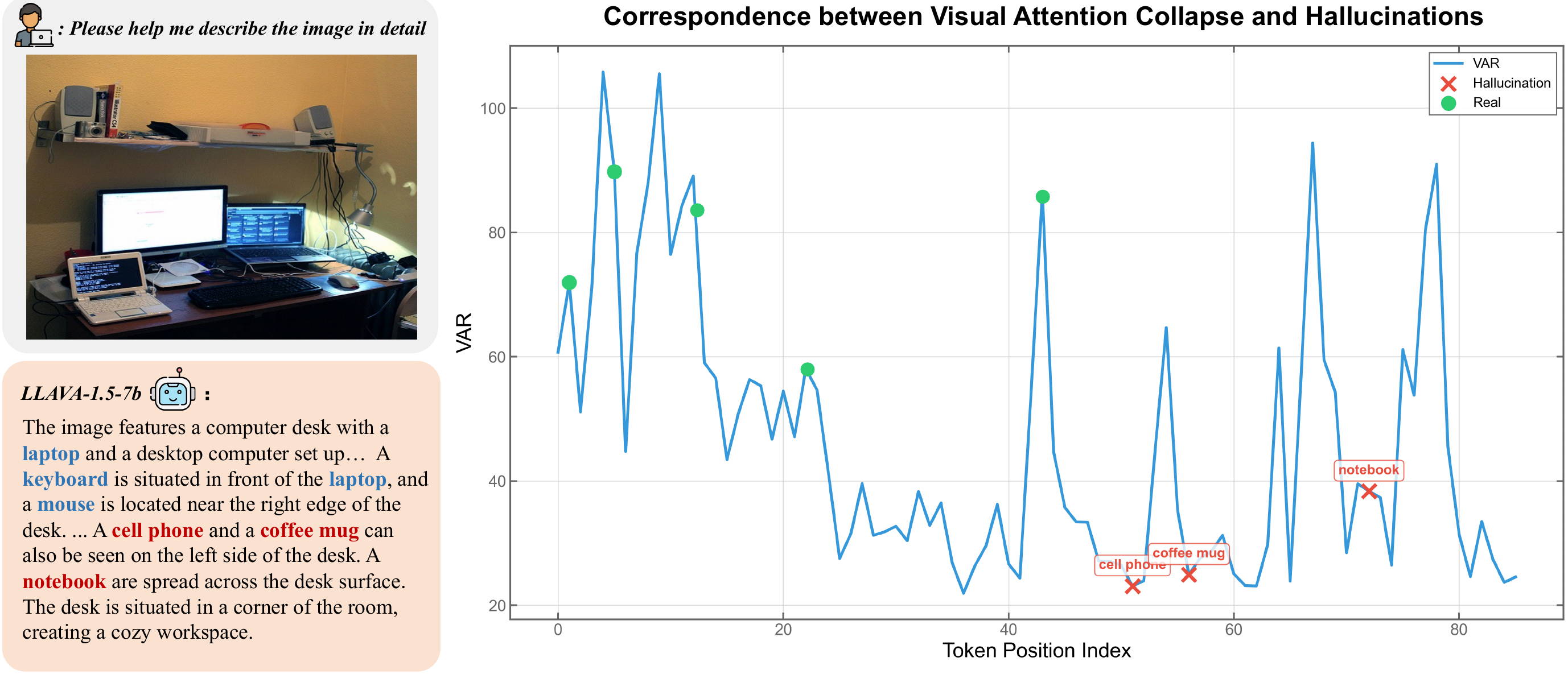}
    \caption{Case study of temporal fading of visual attention.}
    \label{fig:fading-case}
\end{figure}

The Fig. ~\ref{fig:fading-case} provides a qualitative case study of \emph{temporal fading of visual attention}. 
As the generation process unfolds, the proportion of visual attention assigned to successive tokens exhibits a clear overall downward trend. Notably, hallucinated tokens predominantly emerge in regions where visual attention attains local minima, revealing a strong correlation between diminished visual grounding and the onset of hallucinations.

\subsection{Parameter studies on $\alpha$ and $\gamma$}
We present the parameter studies on $\alpha$ and $\gamma$ here. From the result, we can observe that our AFIP is not sensitive to $\alpha$ and $\gamma$, and adjusting $\alpha$ and $\gamma$ does not lead to significant improvements in performance. 
\label{para}
\begin{figure}[htbp]
    \centering
    \begin{subfigure}[b]{0.45\textwidth}
        \centering
\includegraphics[width=\linewidth]{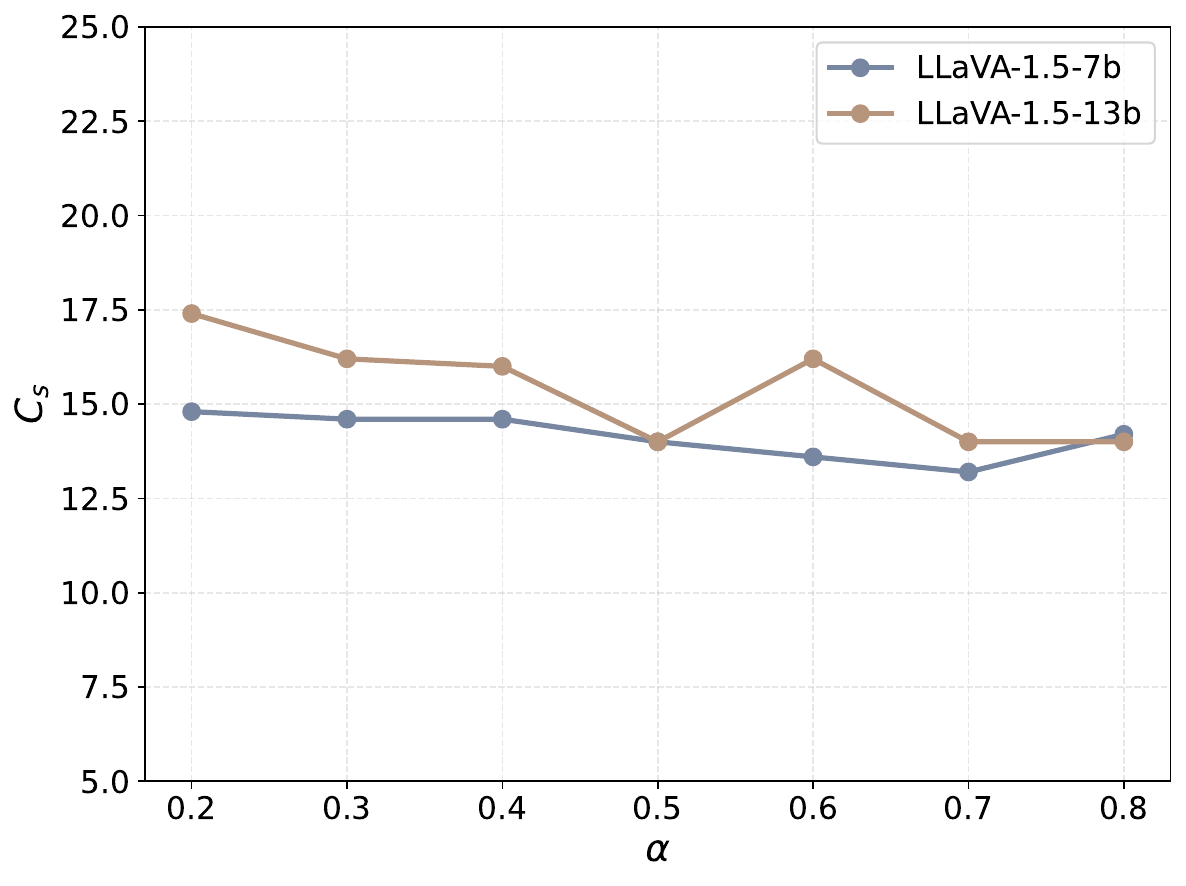}
         \caption{Performance change across different $\alpha$}
        \label{fig:chairs_comparison}
    \end{subfigure}
    \hspace{0.03\textwidth}
    \begin{subfigure}[b]{0.45\textwidth}
        \centering        \includegraphics[width=\linewidth]{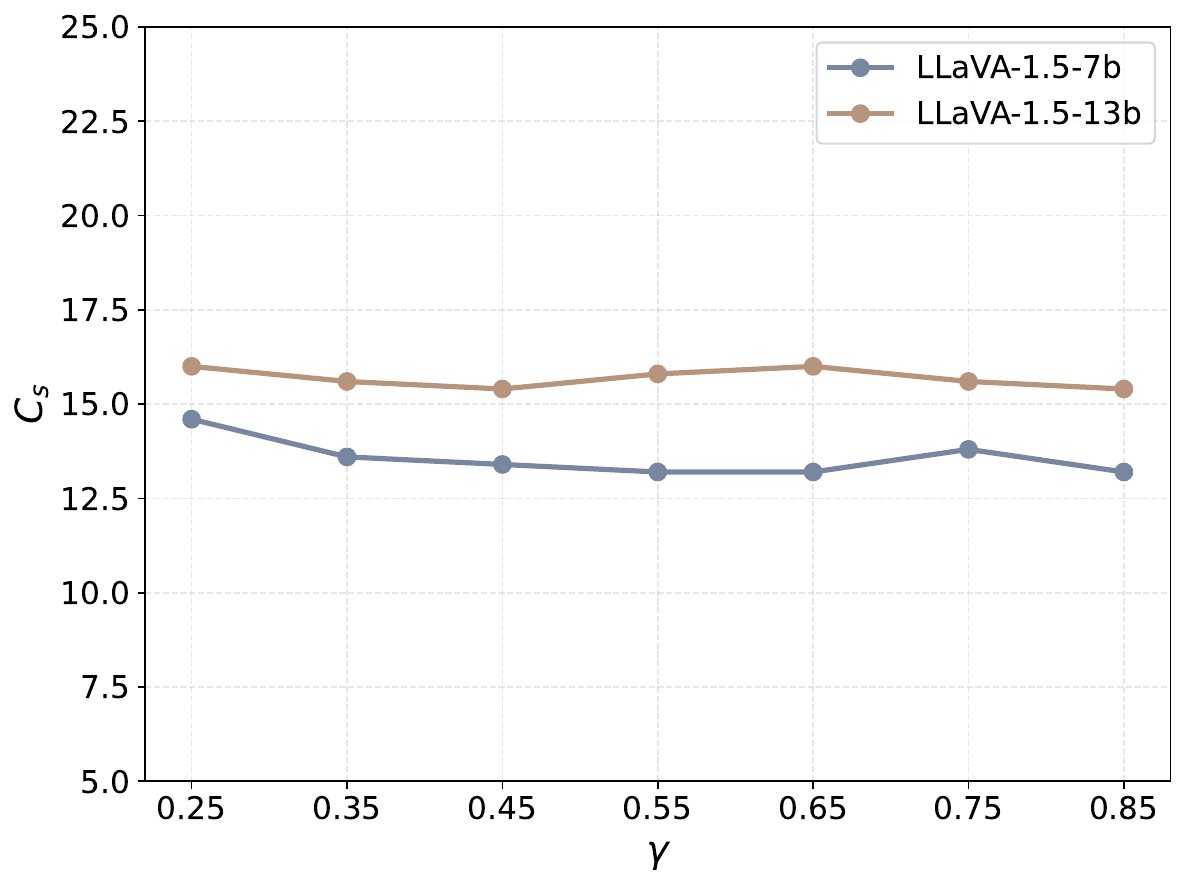}
        \caption{Performance change across different $\gamma$}
        \label{fig:chairs_comparison}
    \end{subfigure}
    \caption{Parameter sensitivity analysis of $\alpha$ and $\gamma$.}
    \label{Parameter1}
\end{figure}

\newpage
\subsection{Visualization of Attention Spatial Inconsistency}
\label{attn-incon}
We present additional visualization examples of LLaVA-1.5-7B to demonstrate that attention spatial inconsistency contributes to the generation of hallucinated object tokens.
\begin{figure}[H]
    \centering
    \includegraphics[width=0.85\linewidth]{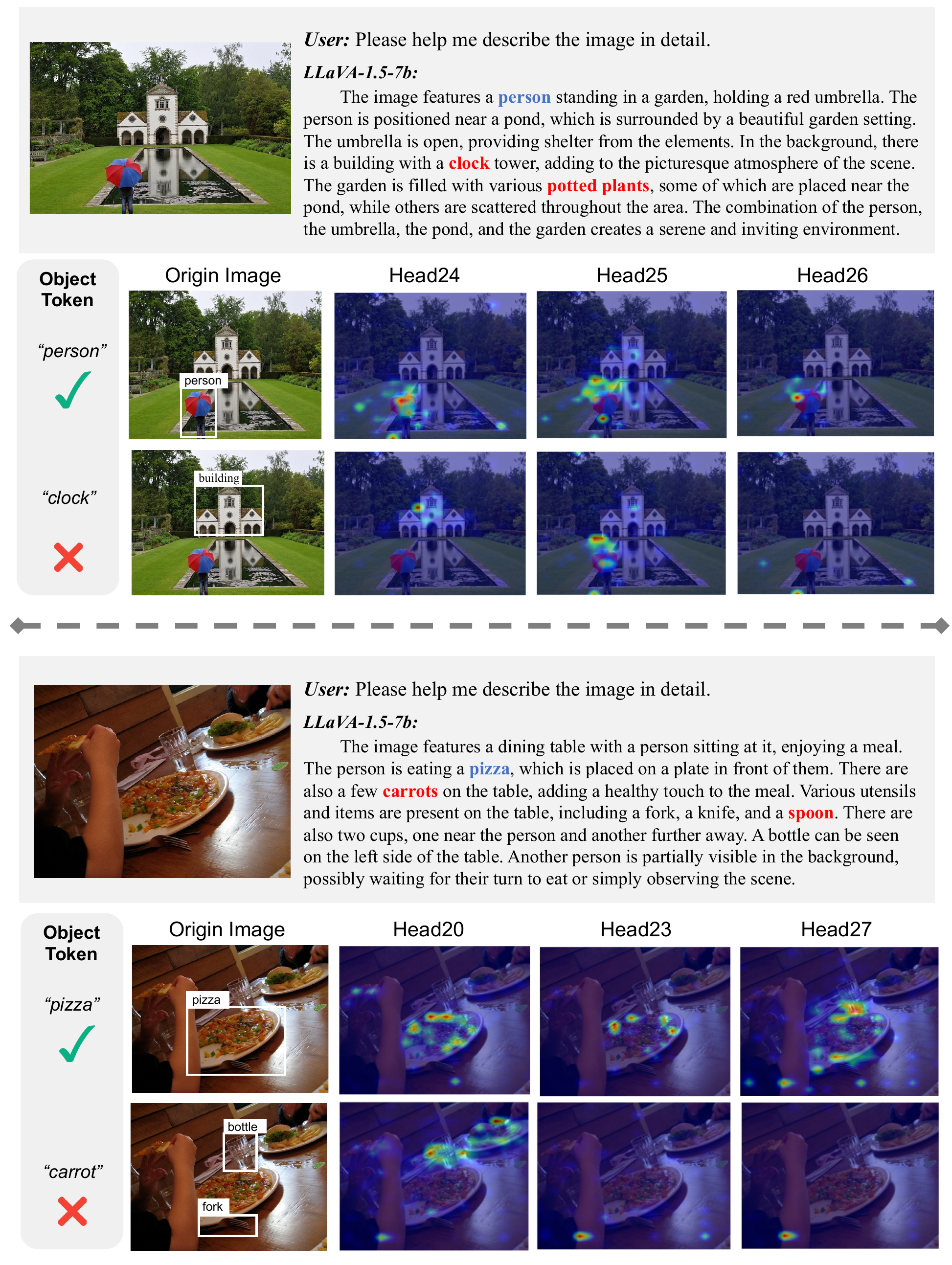}
    \caption{Visualization of attention maps over image for real and hallucinated object tokens.}
    \label{fig:multi3}
\end{figure}

\begin{figure}[H]
    \centering
    \includegraphics[width=0.9\linewidth]{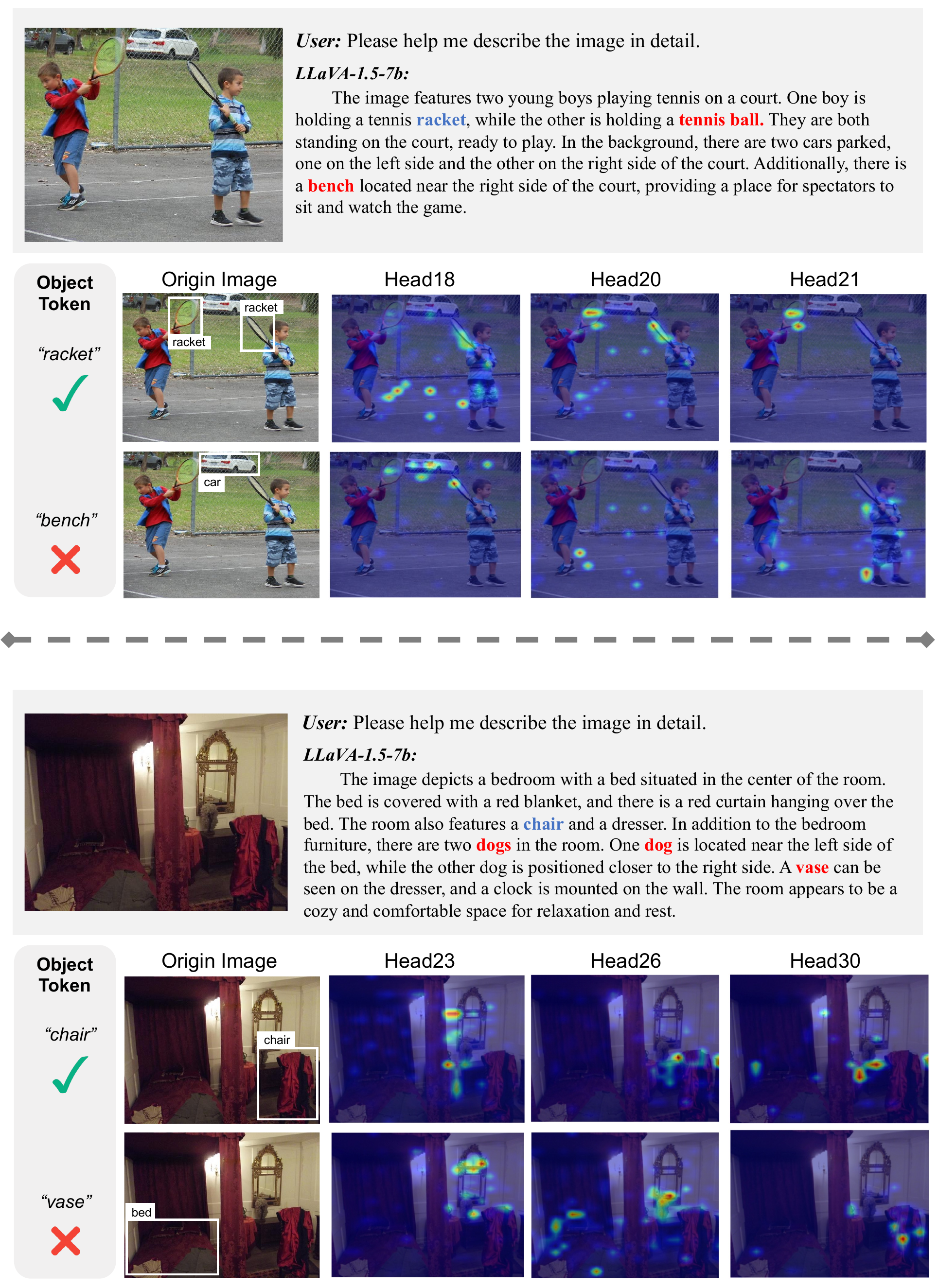}
    \caption{Visualization of attention maps over image for real and hallucinated object tokens.}
    \label{fig:multi2}
\end{figure}

\begin{figure}[H]
    \centering
    \includegraphics[width=0.9\linewidth]{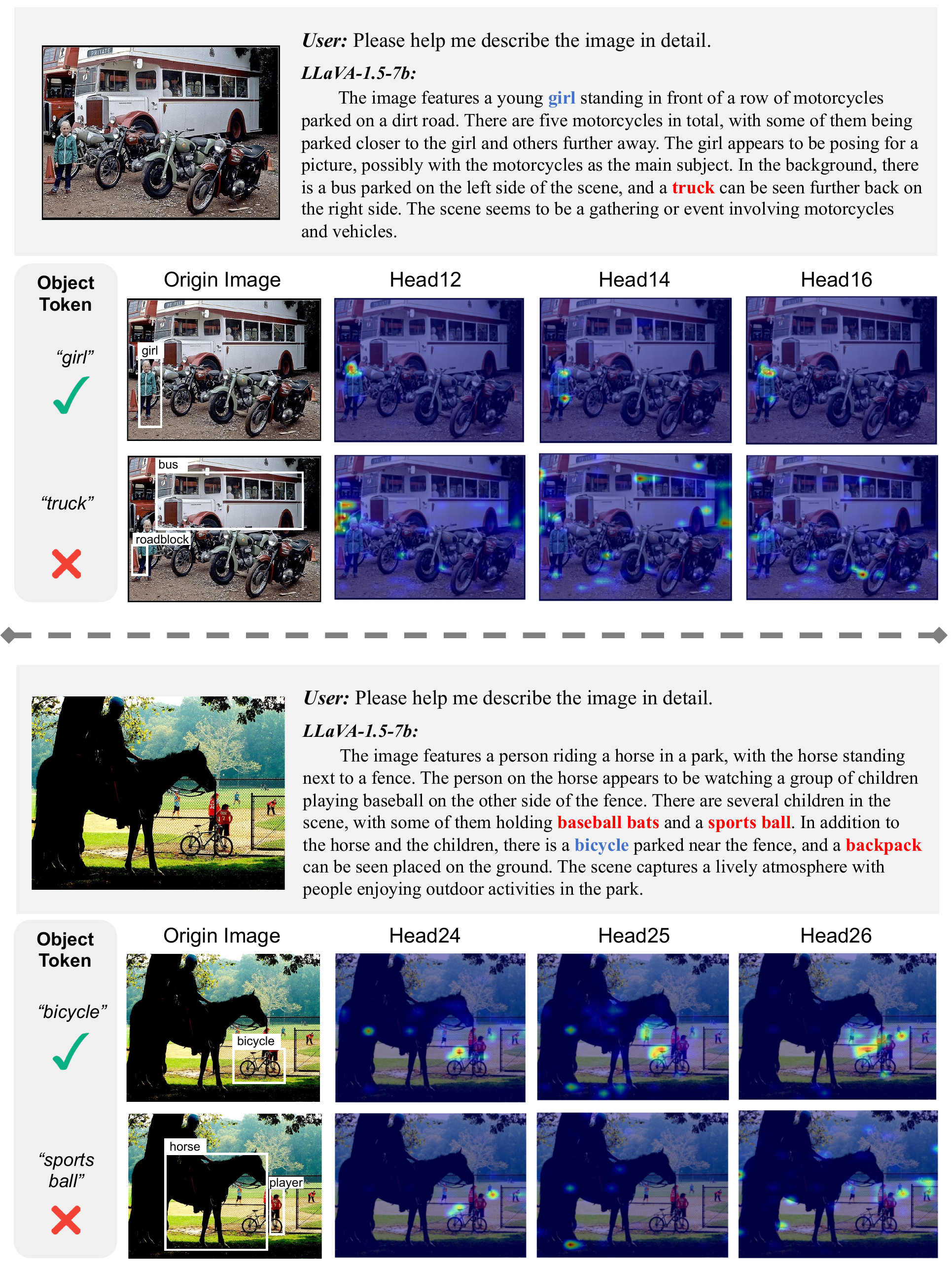}
    \caption{Visualization of attention maps over image for real and hallucinated object tokens.}
    \label{fig:multi1}
\end{figure}

\subsection{Case study for Attention Distraction Correction}
\label{attn-corr}
We present visualizations of the model’s visual attention before and after applying our head-level attention distraction correction method on LLaVA-1.5-7B. All  figures depict attention scores from the 20-th layer,  which demonstrates that  our approach effectively refines the model’s focus toward semantically relevant visual regions.
\begin{figure}[H]
    \centering
    \includegraphics[width=0.9\linewidth]{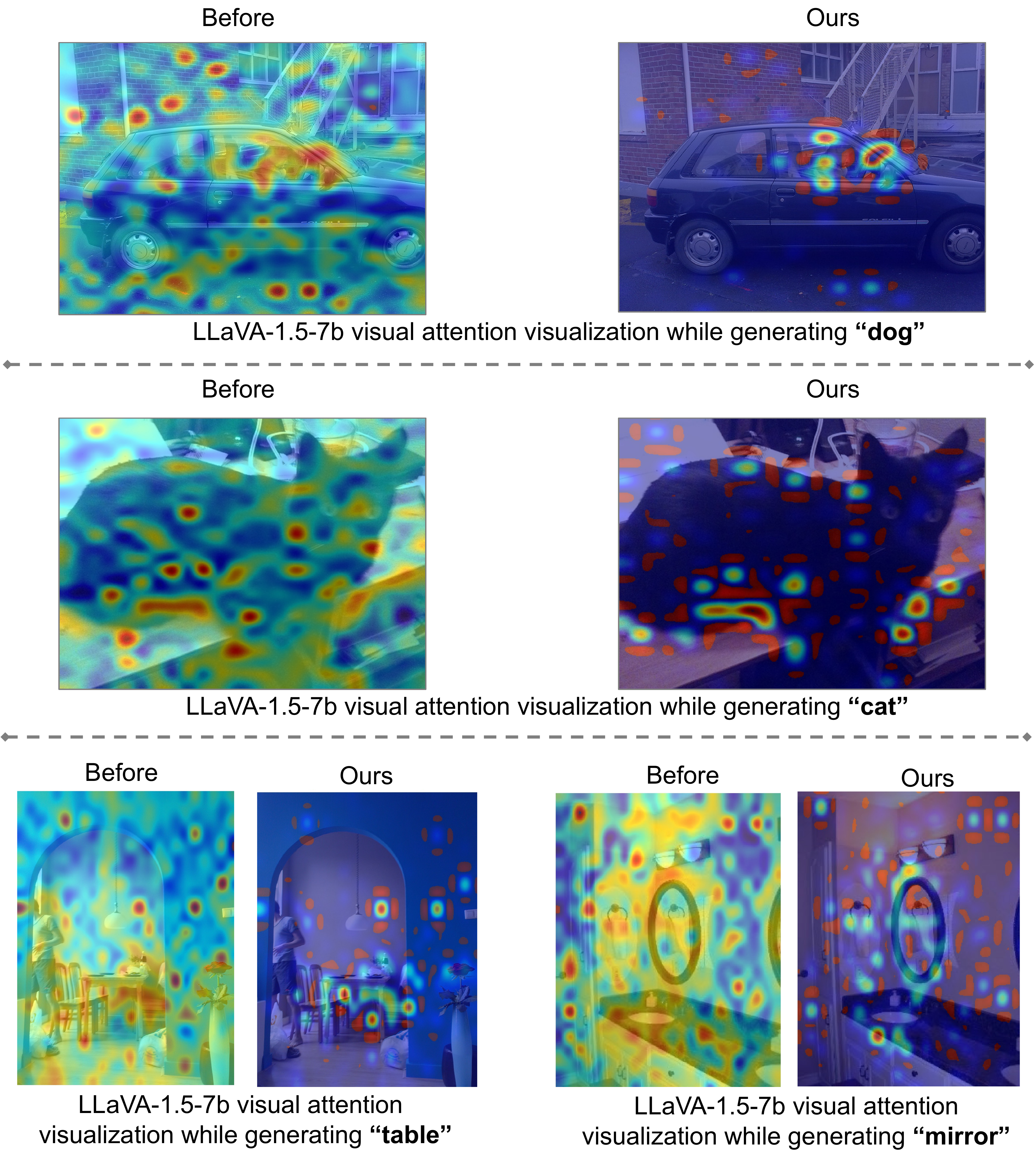}
    \caption{Qualitative comparison of visual attention maps before and after applying Head-Level Attention Distraction Correction.}
    \label{fig:modify-appendix-case}
\end{figure}
\newpage

\begin{figure}[H]
    \centering
    \includegraphics[width=1\linewidth]{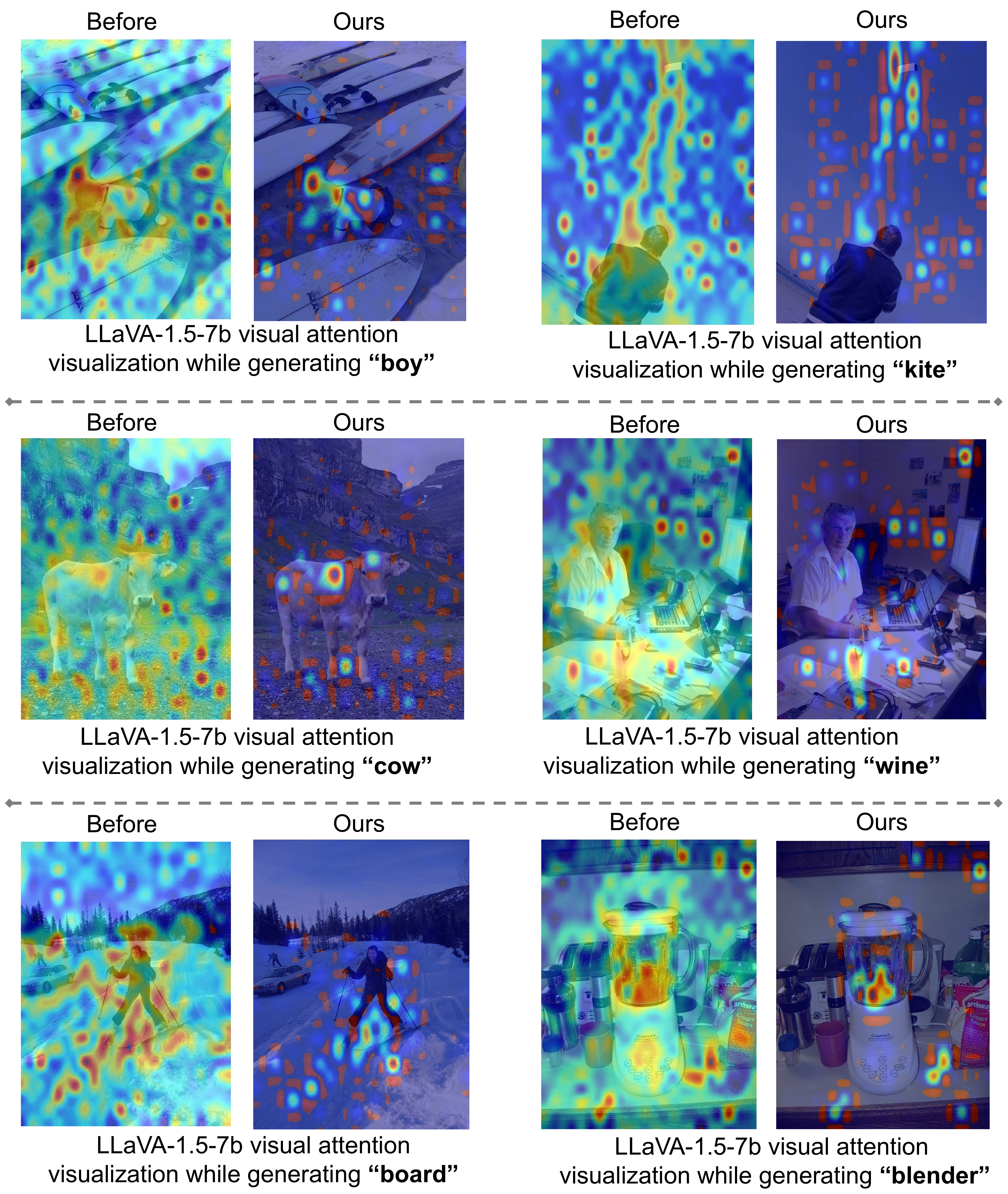}
    \caption{Qualitative comparison of visual attention maps before and after applying head-level attention distraction correction.}
    \label{fig:modify-appendix-case1}
\end{figure}

\subsection{Qualitative Results of Hallucination Mitigation}
We further present sample results from our hallucination mitigation method, as detailed in Section~\ref{method}. Our prompt is ``Please help me describe the image in detail'' for all samples. The real and hallucinated object words are marked in blue and red, respectively. 

\begin{figure}[H]
    \centering
    \includegraphics[width=0.9\linewidth]{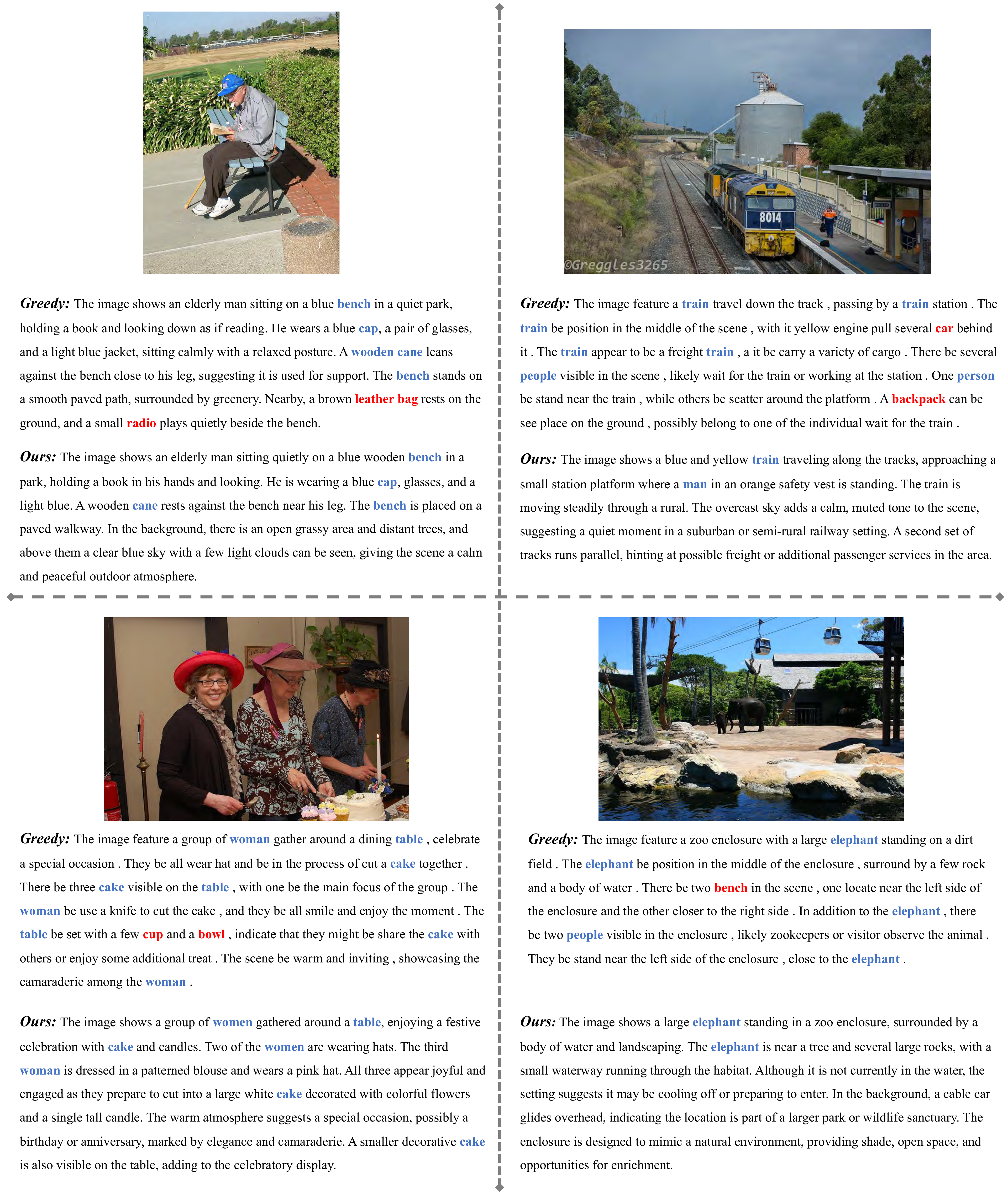}
    \caption{Qualitative results of hallucination mitigation on LLaVA-1.5-7B}
    \label{fig:llava-7b}
\end{figure}

\begin{figure}[H]
    \centering
    \includegraphics[width=\linewidth]{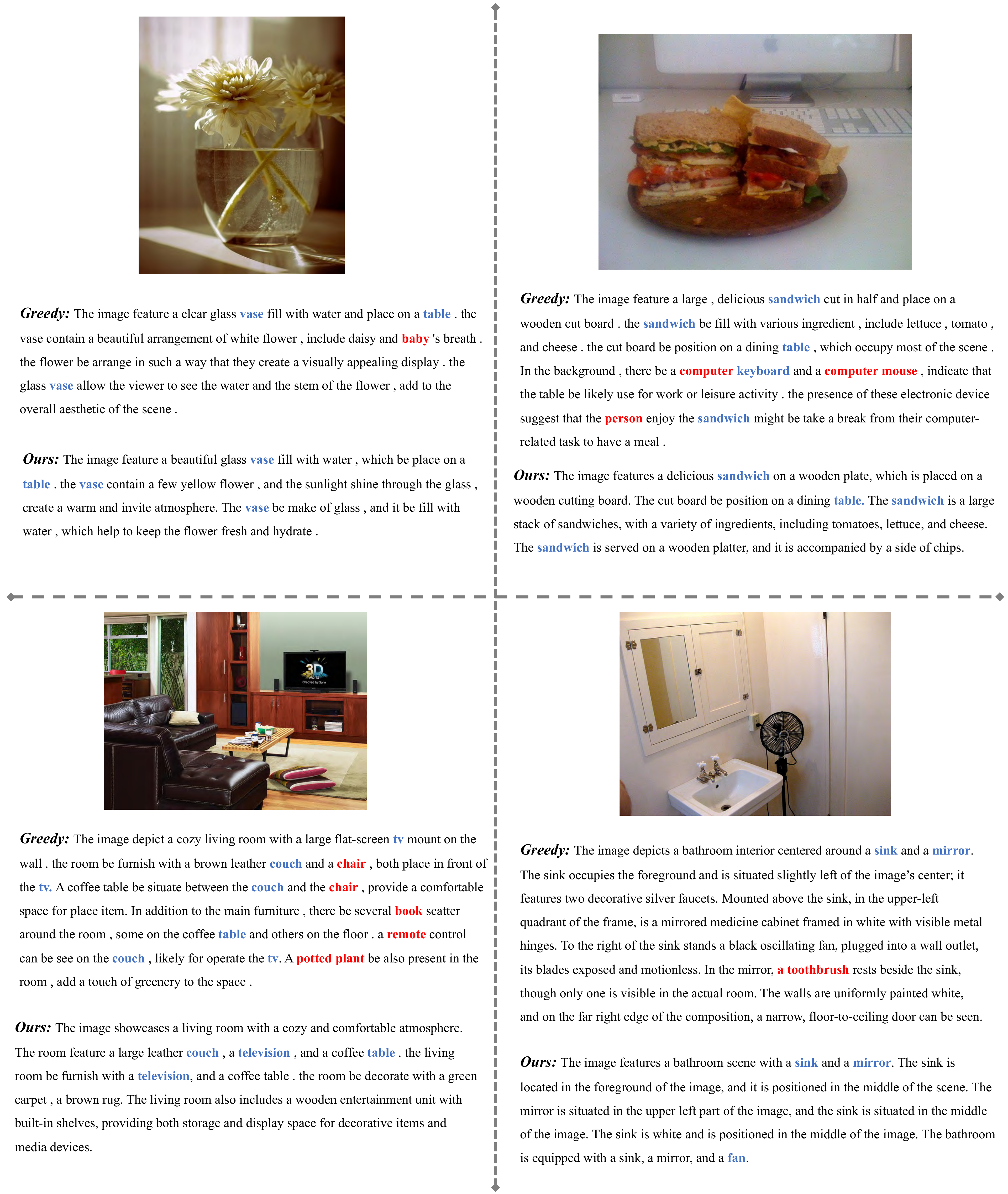}
    \caption{Qualitative results of hallucination mitigation on LLaVA-1.5-13B}
    \label{fig:llava-13b}
\end{figure}

\begin{figure}[H]
    \centering
    \includegraphics[width=\linewidth]{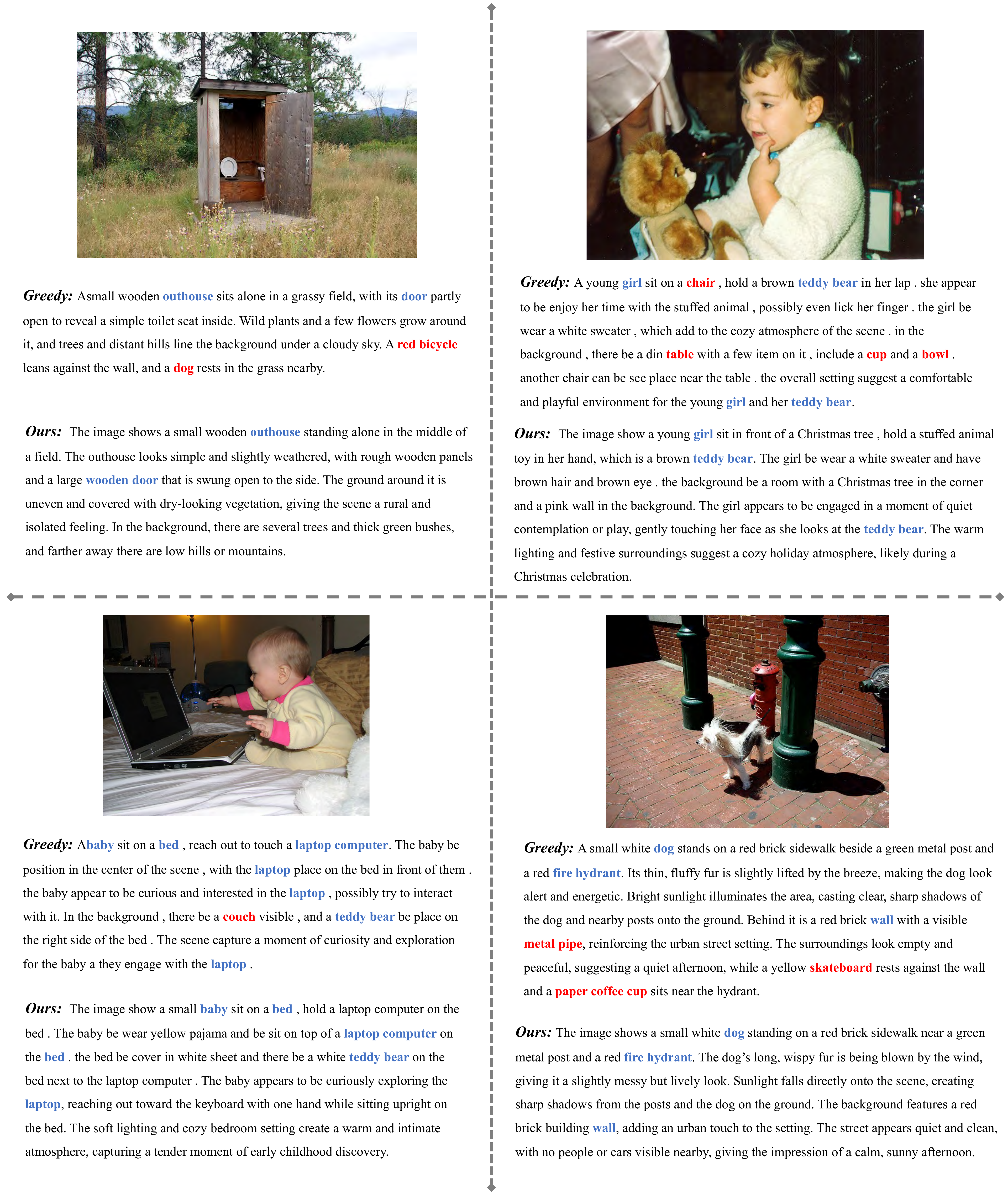}
    \caption{Qualitative results of hallucination mitigation on MiniGPT-4}
    \label{fig:minigpt}
\end{figure}

\begin{figure}[H]
    \centering
    \includegraphics[width=\linewidth]{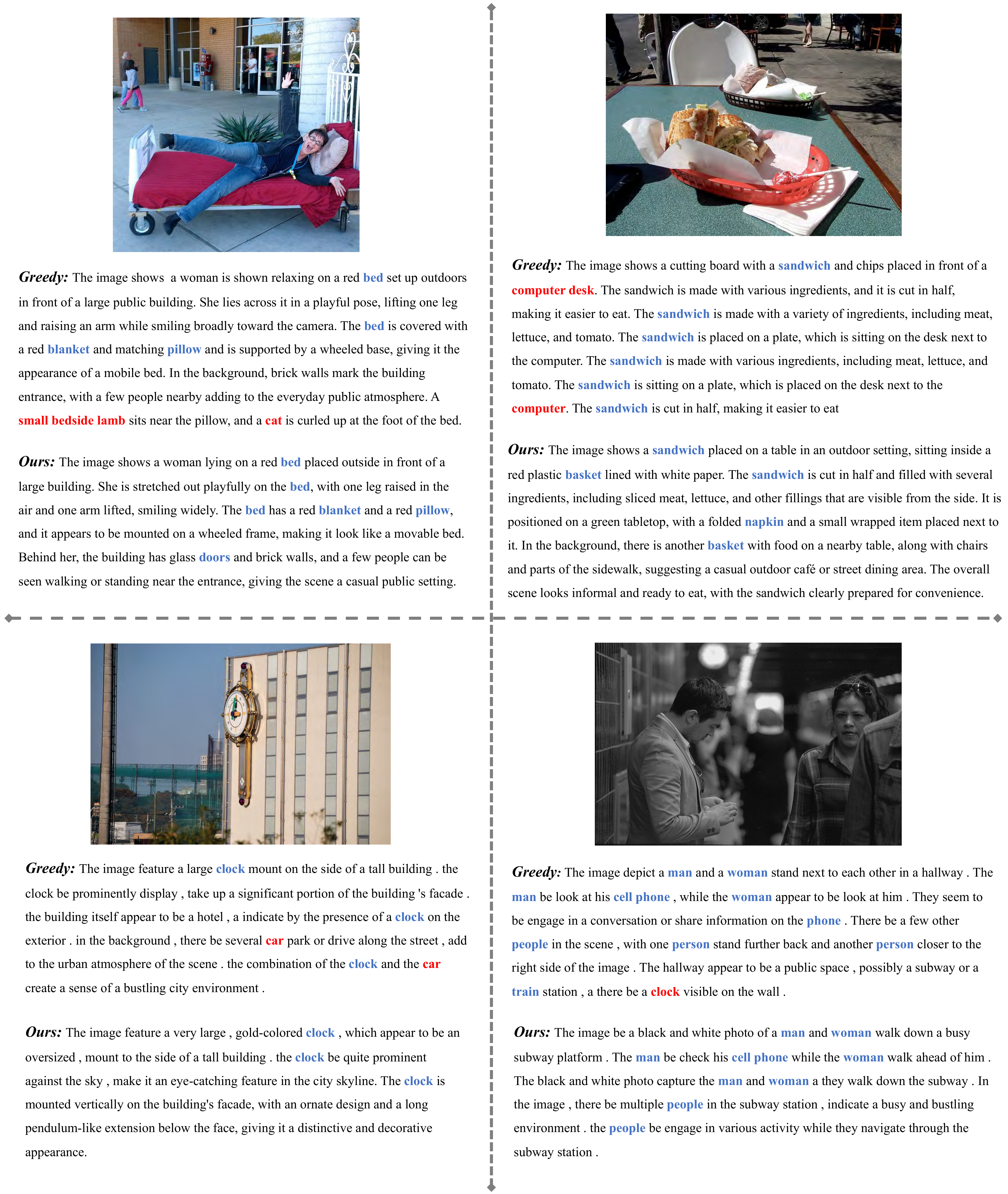}
    \caption{Qualitative results of hallucination mitigation on Shikra}
    \label{fig:shikra}
\end{figure}

\begin{figure}[H]
    \centering
    \includegraphics[width=\linewidth]{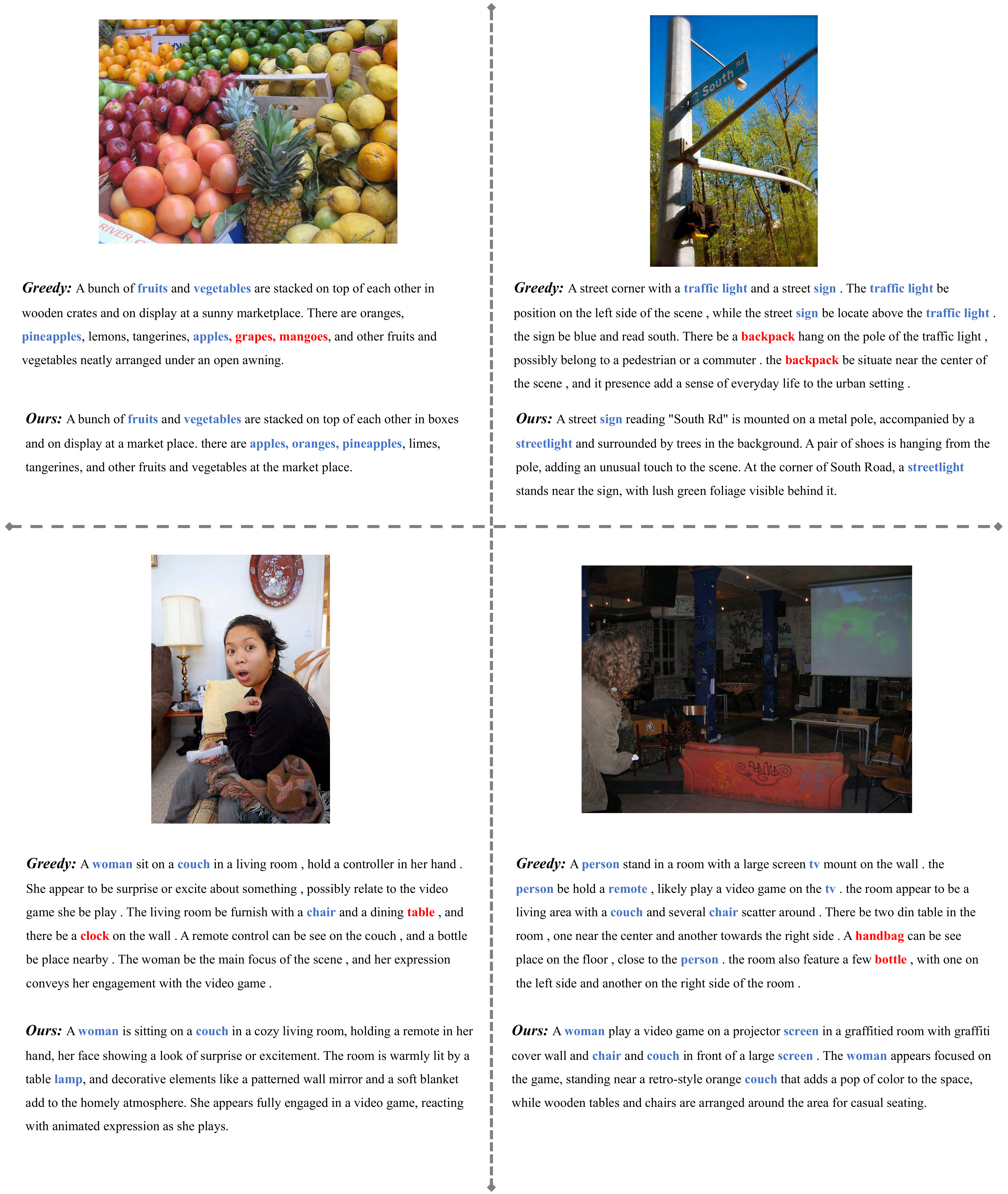}
    \caption{Qualitative results of hallucination mitigation on Qwen-VL}
    \label{fig:qwen}
\end{figure}

\end{document}